%% file: preprint.tex
\documentclass[numbers]{assets/matterlab}

\input{assets/shared-preamble.tex}
\input{editing_commands}

\usepackage{array}

\newcolumntype{P}[1]{>{\raggedright\arraybackslash}p{#1}}
\newcommand{\pipe}{\,\textrightarrow\,}

\usepackage{cinzel}
\usepackage{quantikz}
\usepackage{adjustbox}
\usepackage{wrapfig}
\usepackage{booktabs}
\usepackage{tabularx}
\usepackage{xcolor}

\usepackage[most]{tcolorbox}
\usepackage{braket}
\usepackage[nolist]{acronym}
\usepackage{needspace}
\usepackage{titletoc}
\usepackage{lineno}

\definecolor{LightGray1}{rgb}{0.97,0.97,0.97}
\usepackage{listings}

\lstdefinelanguage{SPARQL}{
  basicstyle=\footnotesize\ttfamily,
  backgroundcolor=\color{LightGray1},
  columns=fullflexible,
  breaklines=true,
  sensitive=true,
  frame=bt,
  aboveskip=1em,
  belowskip=1em,
  xleftmargin=.5em,
  xrightmargin=.5em,
  framexleftmargin=.5em,
  framextopmargin=.5em,
  framexbottommargin=.5em,
  framexrightmargin=.5em,
  tabsize = 2,
  showstringspaces=false,
  morecomment=[l][\color{gray}]{\#},       
  morecomment=[n][\color{blue}]{<http}{>}, 
  morestring=[b][\color{OliveGreen}]{\"},  
  keywordsprefix=?,
  classoffset=0,
  keywordstyle=\color{Sepia},
  morekeywords={},
  classoffset=1,
  keywordstyle=\color{Purple},
  morekeywords={purl,rdf,rdfs,skos,xsd,time,prov,om,owl,saref,yago,dbo,dbr,OntoCAPE_Behavior,OntoCAPE_Material,OntoCAPE_Phase_System,OntoCAPE_Reaction_Mechanism,OntoCAPE_Substance,OntoCAPE_System,OntoSpecies,OntoKin,OntoReaction,OntoDoE,OntoLab,OntoVapourtec,OntoHPLC,OntoDerivation,OntoGoal,OntoAgent,dc,mofs,grafico},
  classoffset=2,
  keywordstyle=\color{MidnightBlue},
  morekeywords={
    SELECT,CONSTRUCT,DESCRIBE,ASK,WHERE,FROM,NAMED,PREFIX,BASE,OPTIONAL,VALUES,DISTINCT,
    FILTER,GRAPH,LIMIT,OFFSET,SERVICE,UNION,EXISTS,NOT,BINDINGS,MINUS,a
  }
}

\tcbset{textmarker/.style={%
    enhanced,
    breakable,
    before=\par\Needspace{12\baselineskip}, 
    frame hidden,
    parbox=false,
    boxrule=0mm,
    boxsep=0mm,
    arc=3mm,          
    outer arc=3mm,    
    left=3mm,right=3mm,
    top=7pt,bottom=7pt,
    toptitle=1mm,bottomtitle=1mm,
    left skip=0.2cm,
    fontupper=\footnotesize
}}

\tcbset{textmarker2/.style={%
    enhanced,
    breakable,
    before=\par\Needspace{12\baselineskip}, 
    frame hidden,
    parbox=false,
    boxrule=0mm,
    boxsep=0mm,
    arc=3mm,          
    outer arc=3mm,    
    left=3mm,right=3mm,
    top=7pt,bottom=7pt,
    toptitle=1mm,bottomtitle=1mm,
    left skip=0.8cm,
    fontupper=\footnotesize
}}

\newtcolorbox{hintBox}{textmarker,
    colback=blue!10!white,
    fontupper=\footnotesize
}

\newtcolorbox{graficoLongBox}{textmarker,
    parbox=true,
    colback=blue!10!white,
    fontupper=\footnotesize,
    title={\faRobot\ \underline{\textbf{Grafico}}},
    title after break={\faRobot\ \underline{\textbf{Grafico (continued)}}},
    colbacktitle=blue!10!white,
    coltitle=black,
    fonttitle=\footnotesize
}

\newtcolorbox{hintBox2}{textmarker2,
    colback=blue!5!white,
    fontupper=\footnotesize
}

\newtcolorbox{importantBox}{textmarker,
    colback=red!10!white,
    fontupper=\footnotesize
}

\newtcolorbox{noteBox}{textmarker,
    colback=matterbg,
    fontupper=\footnotesize
}

\newtcolorbox{noteBoxSmall}{%
    textmarker,
    colback=matterbg,
    fontupper=\footnotesize
}

\newcommand{\prompt}[1]{\begin{noteBoxSmall} \underline{\textbf{Prompt}} #1 \end{noteBoxSmall}}

\newcommand{\graficoreslong}[1]{\begin{graficoLongBox} #1 \end{graficoLongBox}}

\newcommand{\elagenteQ}{{\cinzel El Agente Q}}
\newcommand{\elagenteG}{{\cinzel El Agente Gr\'afico}}
\newcommand{\grafico}{{\cinzel Gr\'afico}}

\title{El Agente Gráfico: A Semantic Execution Runtime for Scientific Agents}

\author[1,2,\dagger,\orcidlink{0000-0002-1246-1993}]{Jiaru~Bai}
\author[1,2,\dagger,\orcidlink{0000-0003-0700-7996}]{Abdulrahman~Aldossary}
\author[3,4,\orcidlink{0009-0009-1341-5831}]{Thomas~Swanick}
\author[1,2,\orcidlink{0000-0002-5398-4192}]{Marcel~M{\"u}ller}
\author[1,2,5,\orcidlink{0000-0001-5191-5735}]{Yeonghun~Kang}
\author[1,2,\orcidlink{0009-0003-1903-2184}]{Changhyeok~Choi}
\author[6,\orcidlink{0000-0003-1546-8709}]{Naruki~Yoshikawa}
\author[2,3,\orcidlink{0000-0002-0581-9168}]{Zijian~Zhang}
\author[7,\orcidlink{0009-0006-8665-011X}]{Jin~Won~Lee}
\author[1,2,\orcidlink{0000-0002-0802-9559}]{Tsz~Wai~Ko}
\author[3,\orcidlink{0009-0004-6626-6714}]{Aiwei~Yin}
\author[1,2,3,\orcidlink{0000-0002-2927-4975}]{Mohammad~Ghazi~Vakili}
\author[3,\orcidlink{0009-0004-5929-109X}]{Chris~Crebolder}
\author[1,3,11,*,\orcidlink{0000-0002-8446-7956}]{Varinia~Bernales}
\author[1,2,3,8,9,10,11,12,13,*,\orcidlink{0000-0002-8277-4434}]{Al\'an~Aspuru-Guzik}

\affiliation[1]{\addressCHEM}
\affiliation[2]{\addressVECTOR}
\affiliation[3]{\addressCS}
\affiliation[4]{\addressMATH}
\affiliation[5]{Department of Chemistry, Sung Kyun Kwan University, 2066 Seobu-ro, 16419, Suwon-si, Gyeonggi-do, Republic of Korea}
\affiliation[6]{National Institute for Materials Science, 1-2-1 Sengen, Tsukuba, Ibaraki 305-0047, Japan}
\affiliation[7]{School of Computer Science, McGill University, 3480 University St., Montréal, QC H3A 0E9, Canada}
\affiliation[8]{\addressMSE}
\affiliation[9]{\addressCHEMENG}
\affiliation[10]{\addressMEDICALSCI}
\affiliation[11]{\addressAC}
\affiliation[12]{\addressCIFAR}
\affiliation[13]{\addressNVIDIA}

\contribution[\dagger]{Contributed equally to this work.}

\abstract{
\input{includes/include-abstract}
}

\date{\today}
\correspondence{
\email{varinia@bernales.org} and \email{alan@aspuru.com}}

\begin{document}

\acrodef{dft}[DFT]{density functional theory}
\acrodef{dftb}[DFTB]{density functional tight-binding}
\acrodef{td-dft}[TD-DFT]{time-dependent density functional theory}
\acrodef{mlip}[MLIP]{machine-learned interatomic potential}
\acrodef{wft}[WFT]{wave function theory}
\acrodef{hf}[HF]{Hartree-Fock}
\acrodef{crest}[CREST]{conformer-rotamer ensemble sampling tool}
\acrodef{qcg}[QCG]{quantum cluster growth}
\acrodef{td}[TD]{time-dependent}
\acrodef{tddft}[TDDFT]{time-dependent density functional theory}
\acrodef{md}[MD]{molecular dynamics}
\acrodef{rmsd}[RMSD]{root mean square deviation}
\acrodef{qc}[QC]{quantum chemistry}
\acrodef{mof}[MOF]{metal-organic framework}
\acrodef{csd}[CSD]{Cambridge Structural Database}
\acrodef{bet}[BET]{Brunauer–Emmett–Teller}
\acrodef{kg}[KG]{knowledge graph}
\acrodef{ogm}[OGM]{object graph mapper}
\acrodef{iri}[IRI]{internationalized resource identifier}
\acrodef{crdt}[CRDT]{conflict-free replicated data type}

\acrodef{llm}[LLM]{large language model}
\acrodef{coala}[CoALA]{cognitive architectures for language agents}
\acrodef{ai}[AI]{artificial intelligence}
\acrodef{ttl}[TTL]{time-to-live}
\acrodef{mcp}[MCP]{model context protocol}
\acrodef{api}[API]{application programming interface}

\maketitle


\input{includes/include-body}



\section*{Acknowledgments}
\input{includes/include-acknowledgement}



{
\small
\bibliography{references}
\bibliographystyle{assets/plainnat}
}


\clearpage

\appendix

{\Huge \textbf{Supporting Information}}
\startcontents[appendices] 
\section*{Contents}
\printcontents[appendices]{l}{1}{\setcounter{tocdepth}{2}}

\input{includes/include-appendix}

\end{document}

%% file: assets/shared-preamble.tex
\usepackage{microtype}
\usepackage{graphicx}
\usepackage{booktabs}
\usepackage{float}
\usepackage{xurl}
\usepackage{hyperref}
\usepackage{titlesec}
\usepackage{tabularx}
\usepackage{chemformula}
\usepackage{siunitx}

\usepackage{algorithm}
\usepackage{algpseudocode}

\usepackage{amsmath}
\usepackage{amssymb}
\usepackage{mathtools}
\usepackage{amsthm}
\usepackage{bm}

\usepackage[noabbrev,nameinlink]{cleveref}

\input{assets/math_commands}

\usepackage{orcidlink}
\usepackage{comment}
\usepackage[version=4]{mhchem}
\usepackage{longtable}
\usepackage{array}
\renewcommand{\arraystretch}{1.7}
\usepackage{multirow}           
\usepackage{amsfonts}           
\usepackage{nicefrac}
\usepackage{duckuments}         
\usepackage{thmtools,thm-restate}
\usepackage{enumitem}
\usepackage[usenames,dvipsnames]{xcolor}
\usepackage{colortbl}
\usepackage{tikz}
\usepackage{tikz-cd}
\usepackage{caption}
\usepackage{subcaption}
\usepackage{listings}
\usepackage{gensymb}
\usepackage{textcomp} 
\usepackage{newunicodechar} 
\newunicodechar{Δ}{$\Delta$}
\newunicodechar{∆}{$\Delta$}
\newunicodechar{−}{\textminus}
\newunicodechar{₀}{$_{0}$}
\newunicodechar{₁}{$_{1}$}
\newunicodechar{₂}{$_{2}$}
\newunicodechar{₃}{$_{3}$}
\newunicodechar{₄}{$_{4}$}
\newunicodechar{₅}{$_{5}$}
\newunicodechar{₆}{$_{6}$}
\newunicodechar{₇}{$_{7}$}
\newunicodechar{₈}{$_{8}$}
\newunicodechar{₉}{$_{9}$}

\newunicodechar{ₐ}{$_{a}$}
\newunicodechar{ₑ}{$_{e}$}
\newunicodechar{ₒ}{$_{o}$}
\newunicodechar{ₓ}{$_{x}$}
\newunicodechar{ₕ}{$_{h}$}
\newunicodechar{ₖ}{$_{k}$}
\newunicodechar{ₗ}{$_{l}$}
\newunicodechar{ₘ}{$_{m}$}
\newunicodechar{ₙ}{$_{n}$}
\newunicodechar{ₚ}{$_{p}$}
\newunicodechar{ₛ}{$_{s}$}
\newunicodechar{ₜ}{$_{t}$}
\newunicodechar{₋}{$_{-}$}
\newunicodechar{₍}{$_{(}$}
\newunicodechar{₎}{$_{)}$}
\newunicodechar{⁻}{$^{-}$}
\newunicodechar{α}{$\alpha$}
\newunicodechar{β}{$\beta$}
\newunicodechar{ω}{$\omega$}
\newunicodechar{σ}{$\sigma$}
\newunicodechar{≤}{$\leq$}
\newunicodechar{≥}{$\geq$}
\newunicodechar{λ}{$\lambda$}
\newunicodechar{≈}{$\approx$}
\newunicodechar{≡}{$\equiv$}

\newunicodechar{ }{\,}
\newunicodechar{─}{\textemdash}
\newunicodechar{π}{$\pi$}
\usepackage[inkscapelatex=false]{svg}
\usepackage{minted}
\usepackage{fontawesome5}
\usepackage[toc,page,header]{appendix}
\usepackage{minitoc}

\newcommand{\addressCHEM}{Department of Chemistry, University of Toronto,  80 St. George St., Toronto, ON M5S 3H6, Canada}
\newcommand{\addressAC}{Acceleration Consortium, 700 University Ave., Toronto, ON M7A 2S4, Canada}
\newcommand{\addressCS}{Department of Computer Science, University of Toronto, 40 St George St., Toronto, ON M5S 2E4, Canada}
\newcommand{\addressVECTOR}{Vector Institute for Artificial Intelligence, W1140-108 College St., Schwartz Reisman Innovation Campus, Toronto, ON M5G 0C6, Canada}
\newcommand{\addressMSE}{Department of Materials Science \& Engineering, University of Toronto, 184 College St., Toronto, ON M5S 3E4, Canada}
\newcommand{\addressCHEMENG}{Department of Chemical Engineering \& Applied Chemistry, University of Toronto, 200 College St., Toronto, ON M5S 3E5, Canada}
\newcommand{\addressMEDICALSCI}{Institute of Medical Science, 1 King's College Circle, Medical Sciences Building, Room 2374, Toronto, ON M5S 1A8, Canada}
\newcommand{\addressCIFAR}{Canadian Institute for Advanced Research (CIFAR), 661 University Ave., Toronto,
ON M5G 1M1, Canada}
\newcommand{\addressNVIDIA}{NVIDIA, 431 King St W \#6th, Toronto, ON M5V 1K4, Canada}

\newcommand{\addressMATH}{Department of Mathematics, University of Toronto, 40 St George St., Toronto, ON M5S 2E4, Canada}

\newcommand{\acknowAC}{This research is part of the University of Toronto’s Acceleration Consortium, which receives funding from the CFREF-2022-00042 Canada First Research Excellence Fund. }
\newcommand{\acknowGEN}[1]{This research was enabled in part by support provided by #1 and the Digital Research Alliance of Canada (\url{https://www.alliancecan.ca}). }
\newcommand{\acknowSciNet}[1]{Computations were performed on the #1 supercomputer at the SciNet HPC Consortium. SciNet is funded by: the Canada Foundation for Innovation; the Government of Ontario; Ontario Research Fund - Research Excellence; and the University of Toronto.}

\newcommand{\acknowDARPA}{This work was supported by the Defense Advanced Research Projects Agency (DARPA) under Agreement No.~HR0011262E022.}

\providecommand{\doi}[1]{\href{https://doi.org/#1}{doi:#1}}


%% file: assets/math_commands.tex
\usepackage{amsmath,amsfonts,bm}

\def\eqref#1{equation~\ref{#1}}

\def\1{\bm{1}}

\DeclareMathAlphabet{\mathsfit}{\encodingdefault}{\sfdefault}{m}{sl}
\SetMathAlphabet{\mathsfit}{bold}{\encodingdefault}{\sfdefault}{bx}{n}



%% file: editing_commands.tex
\usepackage{soul, color}
\newcommand{\Note}[1]{}
\renewcommand{\Note}[1]{#1}  



%% file: includes/include-abstract.tex
Large language models (LLMs) can plan scientific workflows and generate code, but these capabilities do not specify how scientific state is validated, transferred and recorded across heterogeneous computational and experimental operations. Here we present \elagenteG{}, a semantic execution runtime for scientific agents that uses typed execution graphs to enforce admissible scientific state transitions, record provenance and confine model judgement to explicit decision points. Using the same top-level LLM and task-specific rubrics on six university-level quantum chemistry exercises, \elagenteG{} improved performance while reducing model cost by approximately 80\% and wall-clock time by more than fourfold relative to our previous multi-agent architecture. Across ensemble spectroscopy and metal-organic framework (MOF) design, the runtime supported typed-state transfer, parallel execution and cross-session persistence. Open-ended studies of a disulfide redox mechanism and the vibrational hypothesis of olfaction required the agent to define the chemical scope and computational pathway of each investigation. A graph-construction skill distilled from these studies then enabled coding agents to author typed execution graphs for MOF literature mining. Together, these results show how typed execution graphs turn reusable procedural knowledge into scientific protocols that can be inspected, revised and transferred across computational and physical systems.

%% file: includes/include-body.tex
\section{Introduction}


Scientific discovery depends on reproducible protocols that coordinate heterogeneous computational and experimental operations while preserving the state and evidence exchanged between them~\cite{Baker2016reproducibility,wilkinson2016fair}. In chemistry and materials science, a single research question may span literature retrieval, hypothesis generation, simulation and experimental validation across software packages and instruments with different interfaces and data models. \Ac{llm}-based agents provide a flexible interface to this ecosystem by translating natural-language objectives into tool calls, parameterizing calculations and interpreting their outputs~\cite{chemcrow,Boiko2023,googleAIcoscientist,kosmos,Zou2025ElAgente,Ding2025SciToolAgent,Kang2024chatmof,cao2025automating,Pham2026ChemGraph,Wang2025dreams,yaghi2025agentMOF,aitomia,Navarro2025speak,Nduma2025Crystalyse}.

Moving beyond individual tool calls~\cite{chemcrow,Boiko2023}, architectures for scientific agents have evolved to coordinate longer procedures across specialized components. Multi-agent systems distribute tasks and context among specialized agents~\cite{Wei2025AgentSurvey}. In our previous work, we introduced \elagenteQ{}~\cite{Zou2025ElAgente}, a computational chemistry agent built on a multi-agent architecture. The same design has since been extended to automated quantum simulation~\cite{ElAgenteCuantico}, quantum-chemistry research~\cite{elAgenteQuntur}, molecular editing~\cite{elAgenteEstructural} and solid-state simulation~\cite{elAgenteSolido}. Multi-agent decomposition has often compensated for limited \ac{llm} capabilities and context windows by reducing the burden on each agent. More capable models allow a single coding agent to coordinate a workflow while expressing decomposition and orchestration as executable code~\cite{ElAgenteForjador,Huang2025CASCADE}, reducing the communication and verification overhead of multi-agent coordination~\cite{Cemri2025why,Khatua2026CooperBench,scienceOfScalingAgent}. Agent skills that supply reusable instructions and scripts~\cite{gao2025tooluniverse,k_dense2026scientific_agent_skills,deepmindskills,deng2026atomisticskills} further broaden the procedures that agents can construct. These developments expand what agents can generate, but do not define how those procedures should act on scientific state.

Graphs~\cite{Besta2024graph,Zhuge2024gptswarm,graphs_ai_agents}, standardized interfaces~\cite{qiu2026blueprint,Yoshikawa2026nimo} and digital twins~\cite{Darvish2026matterix} provide complementary building blocks for scientific agents. Graphs encode control flow, interfaces connect software and instruments, and digital twins support virtual protocol development and transfer to physical systems. What remains missing is a common execution model that governs state transformations and preserves their evidence. A useful precedent comes from chemical process engineering, where a process is represented as connected operations that exchange material streams with defined properties, and where the network structure constrains which chemical transformations can occur and in what order~\cite{Kondili1993,morbach2007ontocape}. Building on this principle, our central abstraction is a \emph{semantic execution runtime} that extends typed execution graphs from control-flow specification to the governance of scientific state. Node schemas govern state transitions, directed edges delimit admissible actions, and the runtime preserves object identity and native state while recording failures and provenance. Agent-generated procedures thereby become executable protocols with explicit types, decision boundaries and recovery paths. Conversational traces record model reasoning, while the protocol and its evolving scientific state remain the primary computational artifacts.

\begin{figure}[!ht]
    \centering
    \includegraphics[width=\linewidth]{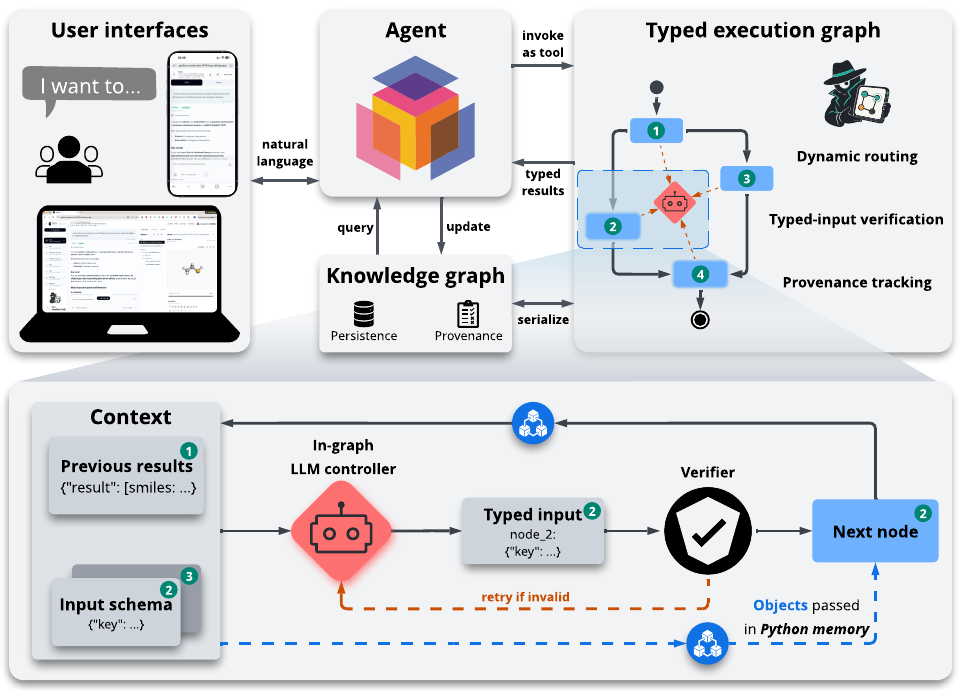}
    \caption{\textbf{Runtime architecture of \elagenteG{}.} Users interact with a top-level agent through a natural-language interface using multi-device access with shared context (Supplementary Figs.~\ref{fig:si-graphchat-interface} and \ref{fig:si-graphchat-interface-phone}). The agent invokes typed execution-graph as tools in which selected states are persisted to an external \ac{kg} after exit. The agent can also query or update the \ac{kg} as its long-term memory. Within each graph, directed edges restrict admissible transitions, and a focused \ac{llm} controller resolves graph-defined routing decisions. The lower panel expands one such decision: the result from node 1 and the input schemas of admissible successors 2 and 3 form the controller context. The controller selects node 2 and returns typed input, which the runtime validates before execution; validation errors are returned for correction. Large scientific objects bypass the controller and pass between nodes in Python memory, while node outputs update the context for subsequent decisions. The GraphChat interface is shown in Supplementary Sec.~\ref{si:graphchat} and Supplementary Video \url{https://www.youtube.com/playlist?list=PLaUD8plXw_ecR7A1EwVAKL3pzIZqjvuVU}.}
    \label{fig:overview}
\end{figure}

Here we present \elagenteG{}, an implementation of this semantic execution runtime. Each workflow is exposed to the agent as a graph tool. Large intermediate objects remain as native Python objects in runtime memory instead of being serialized through the \ac{llm} context, while selected scientific state and provenance are persisted through an \ac{ogm} into an external \ac{kg} for reuse across tools and sessions.
We evaluate the system at three levels. First, we compare task performance, model cost and wall-clock time with our previous multi-agent architecture~\cite{Zou2025ElAgente}. Second, workflows in ensemble spectroscopy, \ac{mof} design and open-ended chemistry investigation examine typed-state transfer, parallel execution and cross-session persistence. Finally, we distill graph-design principles into a reusable skill and test whether coding agents can compile natural-language requests into typed execution graphs. We examine this capability in \ac{mof} literature mining. \autoref{fig:overview} summarizes the runtime architecture.

\FloatBarrier

\section{Results}
We first establish how the runtime separates model-mediated scientific decisions from validated transformations of typed scientific state, then quantify task performance and efficiency on six university-level quantum chemistry exercises. Workflows in different chemistry use cases examine complementary properties of the runtime. Finally, we test whether the same execution contract can guide coding agents to generate new computational protocols.

\subsection{Typed execution graphs govern semantic decisions over scientific state}

\begin{figure}[ht!]
    \centering
    \includegraphics[width=\linewidth]{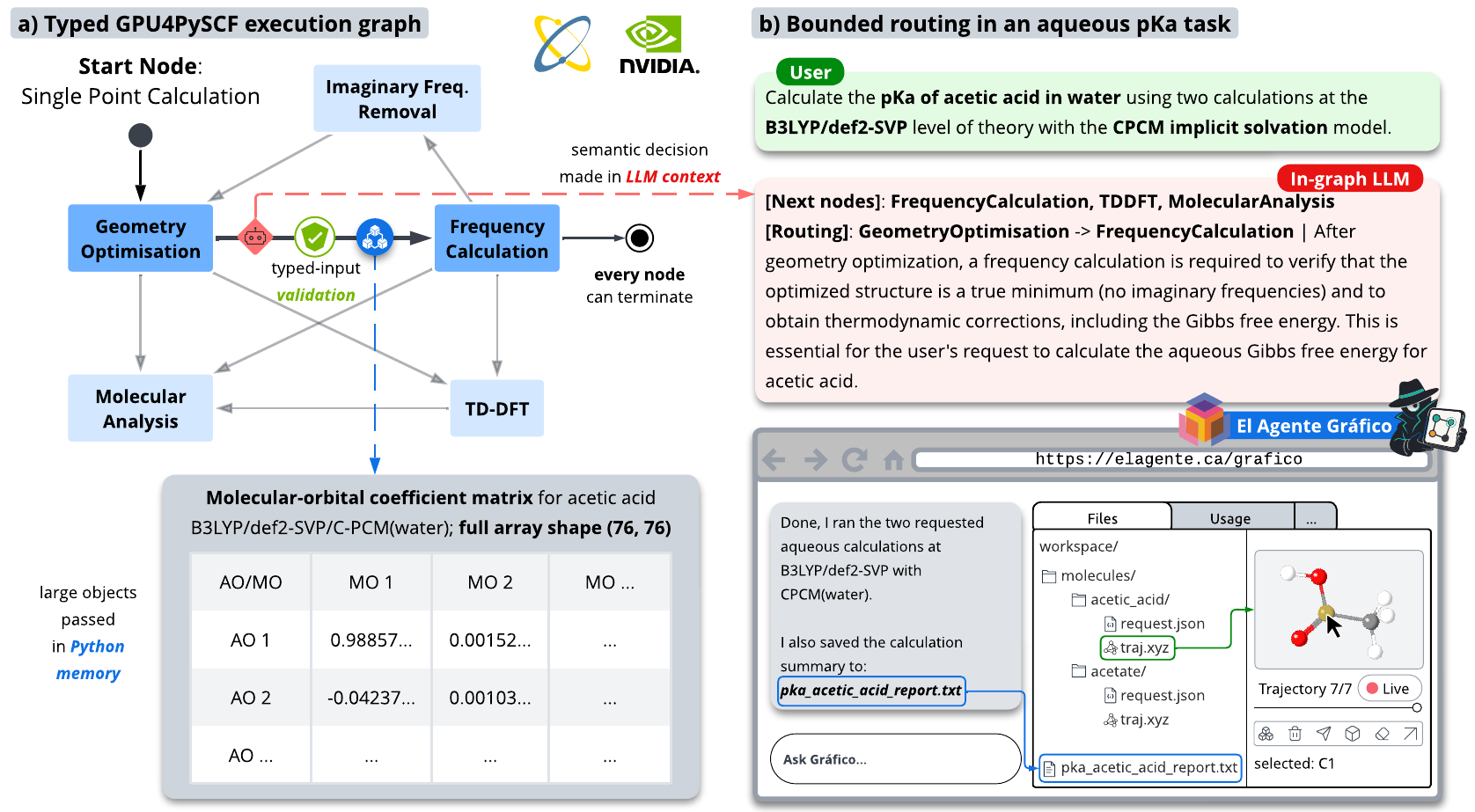}
    \caption{\textbf{Instantiation of the semantic execution runtime for GPU-accelerated quantum chemistry.} (a) The GPU4PySCF~\cite{Wu2025gpu4pyscf,Li2025gpu4pyscf} execution graph represents geometry optimization, frequency calculation, imaginary-frequency removal, molecular analysis and \ac{tddft} as nodes connected by admissible transitions. Large numerical objects, illustrated by the molecular-orbital coefficient matrix, or any other wavefunction data, pass between nodes as native Python objects without entering the \ac{llm} context; the controller may terminate the graph after any node once it determines that the user request has been satisfied. (b) For an aqueous pK$_\mathrm{a}$ benchmark task, the in-graph \ac{llm} controller uses the user request and preceding optimization result to select the permitted transition from geometry optimization to frequency calculation. The runtime validates the typed input before executing the selected node. The GraphChat interface reports the executed calculations and exposes the resulting files and molecular trajectory to the user.}
    \label{fig:architecture}
\end{figure}

\autoref{fig:architecture} illustrates an implementation of typed GPU4PySCF execution graph. The graph defines the available operations, their input state types and admissible transitions, while allowing the execution route to respond to the user's request and intermediate results. The workflow represents single-point calculation, geometry optimization, frequency calculation, imaginary-frequency removal, molecular analysis and \ac{tddft} as distinct nodes (\autoref{fig:architecture}a). Conditional and cyclic edges encode dependencies and corrective cycles, and every node may terminate the run once the request has been satisfied. Atomistic state is represented through an \ac{ogm} class \texttt{ConceptualAtoms}, which validates physicality including charge and multiplicity. Package-specific input and output models retain references to native Python objects, allowing downstream nodes to reuse large intermediate results, such as mean-field state and molecular-orbital coefficient arrays, without reconstructing them or serializing them through the \ac{llm} context.

The aqueous pK$_\mathrm{a}$ task in \autoref{fig:architecture}b illustrates a model-mediated routing decision within this graph. After geometry optimization, the controller selected a frequency calculation because the requested Gibbs free energies required verification of a true minimum and thermochemical corrections. At this decision point, the controller received a summary of the request, the preceding result and the input schemas of the successor nodes permitted by the graph. It supplied typed inputs for the selected node, which the runtime validated before execution. The controller could thus adapt the route to the scientific objective and intermediate result, but only within the successor set defined by the graph.

For reuse beyond the active graph run, the \ac{ogm} serializes selected results as \ac{kg} entries, assigns each entry a unique \ac{iri} and restores the corresponding Python object on retrieval (\autoref{fig:overview}). Relations and provenance retained in the \ac{kg} allow previous results to be queried and reused across tools and sessions without repeating the originating calculation. This persistence path remains separate from the native in-memory transfers used during an active run.

For the molecular tasks reported here, the top-level agent instantiated an independent graph for each molecule, and graph instances could execute concurrently in isolated GPU contexts. Their outputs could be passed to another graph or tool, persisted to the \ac{kg}, or used in the final response. The same runtime is deployed through the cloud-hosted GraphChat interface, where users can invoke the available graph tools and monitor execution events and molecular structures without installing the underlying workflow stack (Supplementary Sec.~\ref{si:graphchat}). We next compare this execution model with \elagenteQ{}~\cite{Zou2025ElAgente} on the same six university-level quantum chemistry exercises.

\FloatBarrier

\subsection{Externalized execution improves task performance and efficiency}
\label{sec:benchmark}

We compared the system-level performance of \elagenteG{} with that of our earlier multi-agent architecture, \elagenteQ{}~\cite{Zou2025ElAgente}, using the same six university-level quantum chemistry exercises and task-specific scoring rubrics. The exercises covered molecular analysis, reaction energetics, ring strain, aqueous p$K_\mathrm{a}$ prediction and excited-state calculations at two difficulty levels. Each exercise was repeated ten times, giving 120 runs per evaluated \ac{llm}. Within \grafico{}, all eight model drivers had access to the same GPU4PySCF workflow, code-execution and unit-conversion tools, and the routing controller was fixed as \texttt{gpt-4o-mini}.

Each \grafico{} run was assessed by two complementary evaluators. A deterministic numerical evaluator checked structured scientific outputs, such as energies and geometries. An independent \texttt{gpt-4o} judge applied the task-specific rubrics previously used by human evaluators for \elagenteQ{} to the full execution trace, assessing task completion, reasoning and reporting. We additionally recorded model requests, trace tokens, monetary cost and wall-clock time. Prompts, rubrics, evaluator implementation, model configurations and diagnostic analyses of context use, error recovery and caching are provided in \cref{subsec:grafico_infrastructure} and Supplementary Secs.~\ref{si:benchmark_questions}, \ref{si:rubric}, \ref{si:llm_model_configs} and~\ref{si:statistics_plots}.

\begin{table}[ht!]
\centering
\caption{Task performance and execution efficiency of \ac{llm} drivers in \grafico{}, averaged over 120 runs per model (six exercises, two difficulty levels, and ten repeats). The best performance is shown in bold, and the second-best performance is underlined. Human evaluators assigned the rubric-based score for \elagenteQ{}~\cite{Zou2025ElAgente} (powered by \texttt{sonnet-3.7}); an independent \texttt{gpt-4o} judge assigned those for \grafico{} using the same task-specific rubrics. The deterministic numerical evaluator was applied only to \grafico{}. Token costs exclude provider-side cache discounts. Context-window saturation, error-recovery cost and carryover tokens quantify trace-level context use; full definitions and model configurations are provided in Supplementary Secs.~\ref{si:statistics_plots} and~\ref{si:llm_model_configs}. All calculations used a compute node with four H100 GPUs, with at most three calculations per GPU and four concurrent agents.}
\label{tab:evals}
\resizebox{\textwidth}{!}{%
\begin{tabular}{lccccccccc}
\toprule
\textbf{\begin{tabular}[c]{@{}c@{}}\ac{llm}\\model\end{tabular}} & \textbf{\begin{tabular}[c]{@{}c@{}}Numerical\\eval. (\%)\end{tabular}} & \textbf{\begin{tabular}[c]{@{}c@{}}\ac{llm} Judge\\eval. (\%)\end{tabular}} & \textbf{\begin{tabular}[c]{@{}c@{}}Trace\\tokens\end{tabular}} & \textbf{\begin{tabular}[c]{@{}c@{}}\ac{llm} \ac{api}\\requests\end{tabular}} & \textbf{\begin{tabular}[c]{@{}c@{}}Token cost\\(USD)\end{tabular}} & \textbf{\begin{tabular}[c]{@{}c@{}}Task duration\\(s)\end{tabular}} & \textbf{\begin{tabular}[c]{@{}c@{}}Context window\\saturation (\%)\end{tabular}} & \textbf{\begin{tabular}[c]{@{}c@{}}Error recovery\\cost (\%)\end{tabular}} & \textbf{\begin{tabular}[c]{@{}c@{}}Carryover\\tokens (\%)\end{tabular}} \\
\midrule
El Agente Q~\cite{Zou2025ElAgente} & \multicolumn{2}{c}{\textit{88.25 (human evaluation)}} & \multicolumn{1}{c}{1,649,616} & 168.43 & \multicolumn{1}{c}{4.67} & \multicolumn{1}{c}{1,827} & n/a & n/a & n/a \\
\hline
sonnet-3.7 & 93.69 & 90.94 & 284,036 & 9.14 & 0.92 & 404 & 19.46 & \textbf{5.07} & 86.30 \\
sonnet-4.5 & 96.07 & 95.67 & 320,397 & 6.58 & 1.09 & 273 & 26.79 & 9.02 & 83.28 \\
gpt-4.1 & 93.71 & 96.52 & 113,175 & 5.11 & 0.25 & \underline{208} & \textbf{3.59} & \underline{8.17} & 66.79 \\
gpt-5 & \textbf{98.88} & \textbf{98.50} & \textbf{83,613} & \textbf{3.32} & \underline{0.17} & 228 & 9.78 & 9.59 & \textbf{53.21} \\
gpt-5.1 & 98.37 & \underline{96.61} & 99,775 & \underline{3.65} & 0.17 & 211 & 10.43 & 8.69 & \underline{58.19} \\
gpt-5.2 & \underline{98.69} & 96.19 & \underline{95,153} & 4.40 & 0.22 & \textbf{180} & \underline{8.42} & 8.27 & 64.59 \\
minimax-m2 & 89.66 & 87.61 & 210,018 & 6.40 & \textbf{0.05} & 315 & 21.93 & 16.38 & 78.60 \\
qwen3-max & 93.54 & 90.33 & 178,671 & 3.84 & 0.24 & 296 & 26.27 & 11.82 & 61.45 \\
\bottomrule
\end{tabular}}
\end{table}

The comparison using \texttt{sonnet-3.7} in \autoref{tab:evals} provides the direct system-level contrast. With the common task-specific rubrics, the score increased from 88.25\% for \elagenteQ{} (human evaluators) to 90.94\% for \grafico{} (\ac{llm} judge); the independent numerical evaluator assigned \grafico{} a score of 93.69\%. Over the same tasks, \grafico{} reduced the mean number of model requests from 168.4 to 9.1 and trace-token consumption from 1,649,616 to 284,036 per task. These changes correspond to 94.6\% fewer requests, 82.8\% fewer tokens, an 80.3\% reduction in model cost and a decrease in mean wall-clock time from 1,827 to 404~s (4.5-fold).

Holding the runtime and tool set fixed exposed differences in how model drivers allocated reasoning and interacted with scientific operations. In the more difficult ring-strain exercise, \texttt{gpt-4.1} frequently generated cycloalkenes instead of the requested cycloalkanes; this error was absent from the \texttt{gpt-5.x} runs (Supplementary Figs.~\ref{fig:si_radar_numerical_evaluator} and~\ref{fig:si_radar_llm_judge_evaluator}). Under the same low reasoning-effort setting, \texttt{gpt-5} used more reasoning tokens than \texttt{gpt-5.1} and \texttt{gpt-5.2} across the exercises, while the relative token use of the two later models varied by task (Supplementary Fig.~\ref{fig:si_reasoning_tokens}). Thus, similar aggregate task scores can conceal substantial differences in both task-specific errors and reasoning-token allocation~\cite{Wu2025whenMoreIsLess,Du2025ockbench}.

Tool traces showed a distinct interaction pattern. GPT drivers generally batched several scientific tool calls within one model turn, whereas the Claude drivers more often interleaved smaller call batches with additional model turns. Within the tested GPT sequence, PySCF use remained similar while calls to the general-purpose \texttt{run\_python\_code} tool increased from \texttt{gpt-4.1} through \texttt{gpt-5.2} (Supplementary Figs.~\ref{fig:si_tool_call_agg} and~\ref{fig:si_tool_call_details}). This increased use of executable code coincided with the 2025 introduction of coding agents such as Codex and the explicit optimization of GPT-5 for coding and agentic tool use~\cite{OpenAI2025Codex,OpenAI2025GPT5Developers}. This trend motivated the graph-authoring experiments below, in which coding agents compile procedures into typed execution graphs. Interaction style also affected repeated context processing: carryover tokens accounted for 53.21--86.30\% of trace-token consumption across the model drivers, and the reported costs exclude provider-specific cache discounts (Supplementary Sec.~\ref{si:statistics_plots}).

To summarize repeated-run outcomes, we defined a successful run as one with a numerical score of 1.00 and a rubric-based score greater than 0.90. We then calculated both pass@k~\cite{chen2021evaluating}, the probability that at least one of (k) runs is successful, and $\mathrm{pass}^{k}$~\cite{demystifying_evals_for_AI_agents}, the probability that all of the (k) runs are successful. For \texttt{gpt-5}, pass@3 was 0.99 and $\mathrm{pass}^{3}$ was 0.54. Full repeated-run results are provided in Supplementary Sec.~\ref{si:pass_k_results}.

\subsection{Runtime properties across scientific workflows}

We next examined how the runtime handled scientific state across heterogeneous tools, concurrent branches and separate sessions. Three application classes examined complementary properties: ensemble spectroscopy tested typed state transfer and parallel fan-out, \ac{mof} design tested persistent relational state, and two open-ended chemical investigations tested model-directed choices within graph-constrained execution.

\subsubsection{Typed state transfer enables parallel ensemble spectroscopy}

\begin{figure}[!ht]
    \centering
    \includegraphics[width=\linewidth]{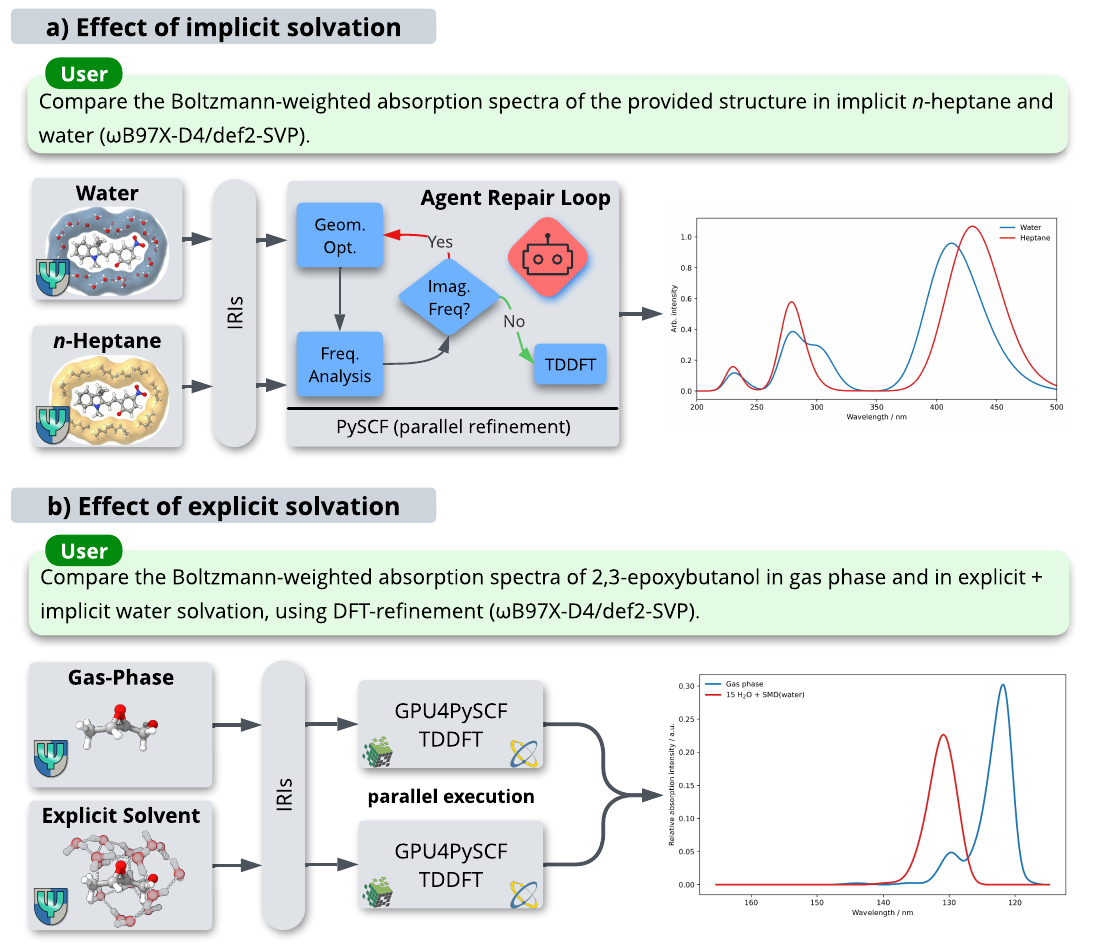}
    \caption{\textbf{Typed state transfer and parallel execution for ensemble spectroscopy.} \grafico{} generated gas-phase and solvated conformer ensembles with \ac{crest} and \ac{qcg}, persisted the resulting \texttt{ConceptualAtoms} objects and passed their \acp{iri} as lightweight handles to parallel GPU4PySCF refinement and \ac{tddft} workflows. The conformer-level outputs were joined for Boltzmann reweighting and absorption-spectrum construction. The resulting plots compare merocyanine under implicit solvation and 2,3-epoxybutanol in the gas phase and an explicit water environment. Spectra were generated by \grafico{}; plot titles and background grids were removed manually. Prompts are summarized; complete prompts and transcripts are provided in Supplementary Sec.~\ref{si:crest_pyscf}.}
    \label{fig:crest}
\end{figure}

Spectroscopic observables of flexible molecules depend on conformer ensembles beyond a single optimized structure~\cite{desouzaGOATGlobalOptimization2025}. Their prediction requires stochastic conformer sampling, transfer to higher-level electronic-structure methods, validation of each optimized structure and population-weighted analysis~\cite{desouzaGOATGlobalOptimization2025,prachtAutomatedExplorationLowenergy2020,prachtCRESTProgramExploration2024,rdkit_2025_03}. We used implicit- and explicit-solvation studies to examine whether typed state could move across these stages while calculations fanned out over solvents and conformers (\autoref{fig:crest}). \ac{crest} and \ac{qcg} outputs were persisted as \texttt{ConceptualAtoms} objects and passed between graph tools by \ac{iri}, keeping coordinate payloads outside the model context. Within each graph run, native Python objects remained in runtime memory. The agent selected the workflows and comparisons, dispatched solvent- and conformer-level calculations in parallel, and wrote Python code to deduplicate conformers, recompute Boltzmann weights and aggregate the \ac{tddft} outputs into spectra. Typed graph nodes performed the underlying calculations and, in the implicit-solvent study, enforced frequency validation and repair. A detailed account of the agent's tool orchestration, code generation and analysis is provided in Supplementary Sec.~\ref{si:crest_pyscf}.

\paragraph{\textbf{Implicit-solvent ensembles}}
For a merocyanine compound, \grafico{} launched concurrent \ac{crest} searches at the GFN2-xTB/ALPB level~\cite{bannwarthGFN2xTBAccurateBroadly2019,ehlertRobustEfficientImplicit2021} in water and \textit{n}-heptane and dispatched five conformers from each solvent to parallel $\omega$B97X-D4~(SMD)/def2-SVP refinement and \ac{tddft} workflows~\cite{mardirossianOb97XV10parameterRangeseparated2014,caldeweyherGenerallyApplicableAtomiccharge2019,weigendBalancedBasisSets2005,marenichUniversalSolvationModel2009}. Frequency analysis verified the retained minima; one \textit{n}-heptane conformer entered the graph's imaginary-mode removal and reoptimization path before excited-state calculation. After deduplication and Boltzmann reweighting, the ensemble absorption maximum shifted from 412.8~nm in water to 429.9~nm in \textit{n}-heptane. The corresponding dominant-conformer maxima were 412.6 and 429.6~nm, indicating that the calculated 17~nm shift arose primarily from the solvent model and was only weakly altered by conformational averaging.

\paragraph{\textbf{Explicit-solvent ensembles}}
For 2,3-epoxybutanol~\cite{beckEvaluatingImportanceConformers2024}, a gas-phase \ac{crest} search ran concurrently with construction of a 15-water solvation shell by \ac{qcg}~\cite{Spicher2022QCG} and subsequent non-covalent conformer sampling. Four gas-phase and eight microsolvated conformers covered more than 95\% of their xTB-determined Boltzmann populations and were refined before \ac{tddft} calculation. Explicit water together with implicit solvation model SMD shifted the lowest bright absorption from 9.47 to 9.21~eV and the ensemble absorption maximum from 121.8 to 130.9~nm, consistent with perturbation of the excited-state electronic structure by direct solute--solvent interactions.

Across both studies, lightweight persistent identifiers connected heterogeneous tools, native objects carried computational state within each graph, and parallel branches returned typed outputs for agent-authored ensemble analysis. The runtime preserved a traceable object lifecycle from conformer sampling to an ensemble observable without reconstructing molecular coordinates from text.

\subsubsection{Persistent relational state supports iterative MOF design}

\begin{figure}[!ht]
    \centering
    \includegraphics[width=\linewidth]{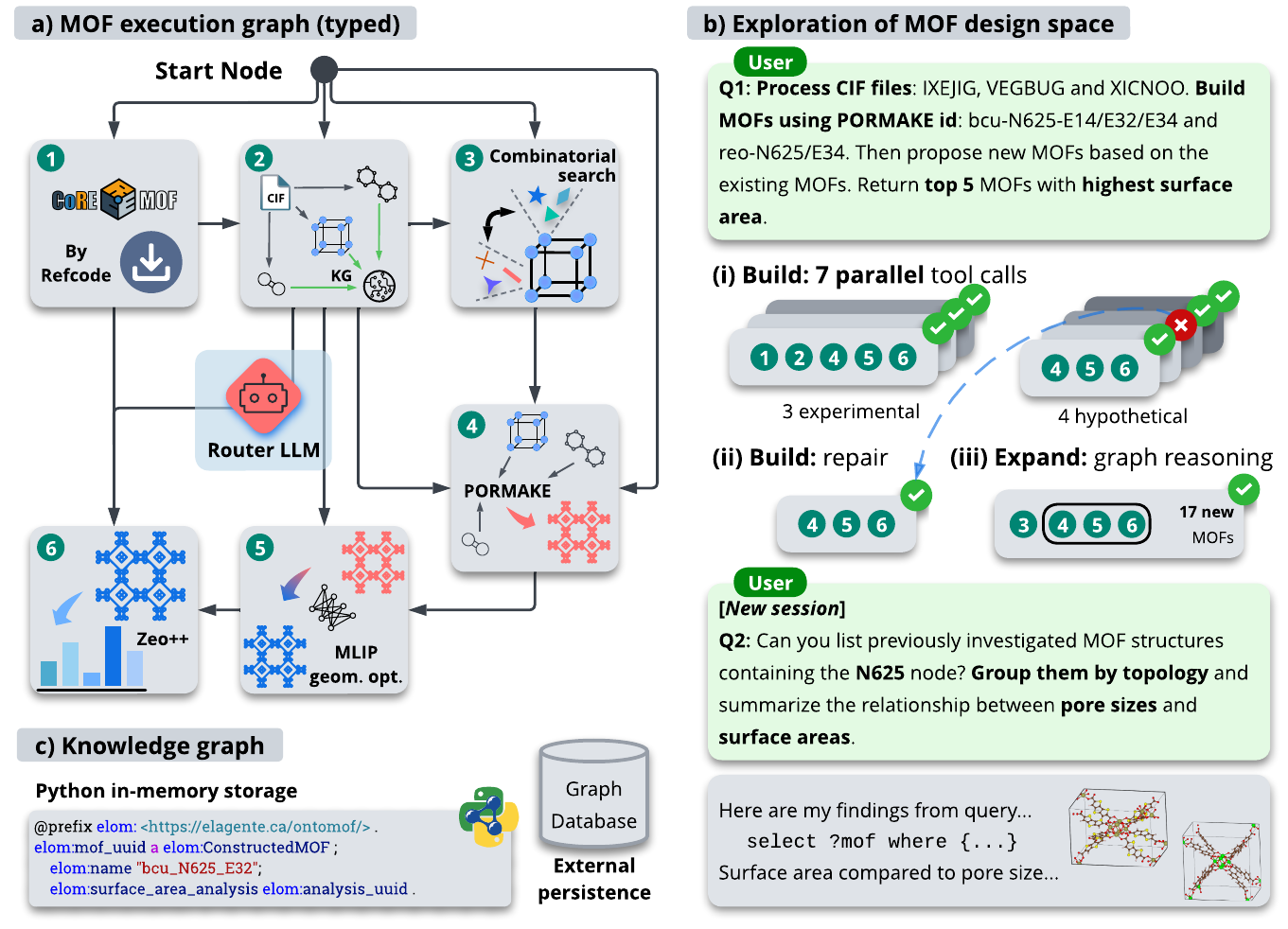}
    \caption{\textbf{Persistent relational state for iterative MOF design.} (a) The workflow acquired structures from the CoRE-MOF database by CCDC refcode~\cite{Moghadam2017CCDC,Zhao2025coremof}, decomposed them into related topology, metal-node and linker entities, proposed new combinations, constructed structures with PORMAKE~\cite{Lee2021pormake}, optimized them with the GPU-accelerated MACE-MOF \ac{mlip}~\cite{batatia2022mace,mofsimbench} and calculated porosity with Zeo++~\cite{Willems2012zeopp}. The object graph mapper preserved entity identity and relations during concurrent construction. (b) An initial session built and analysed experimental and hypothetical structures (Q1); a new session queried the accumulated state by metal node and topology without access to the preceding conversation (Q2). Prompts are summarized; complete prompts and transcripts are provided in Supplementary Sec.~\ref{si:mof_exploration}.}
    \label{fig:mof}
\end{figure}

Reticular materials are defined by relations among topologies, metal nodes, organic linkers and the structures assembled from them~\cite{Boyd2016generalized,kondinski2022automated,Kim2024updown}. Exploring their combinatorial design space requires structure generation~\cite{Bucior2019mofid1,Yao2021vaemof,Lee2021pormake}, geometry optimization with MOF-adapted force fields~\cite{Addicoat2014uff} or machine-learned interatomic potentials~\cite{batatia2022mace,mofsimbench}, and porosity analysis~\cite{Willems2012zeopp}, while preserving object identity and provenance as components are reused across workflows. We examined this property by asking \grafico{} to construct and explore a \ac{kg} containing experimentally reported, hypothetical and graph-inferred \acp{mof} (\autoref{fig:mof}; Supplementary Sec.~\ref{si:mof_exploration}).

\paragraph{\textbf{Concurrent construction and relational reuse}}
In the initial session, \grafico{} concurrently dispatched seven workflows to process three deposited structures and construct four initial hypothetical \acp{mof}. A retrieve-or-create operation in the \ac{ogm} canonicalized topology and building-block objects in memory and in the \ac{kg}, so repeated entities retained unique and consistent \acp{iri} across branches. The agent then queried the accumulated relations to enumerate feasible topology--metal-node--linker combinations, after which deterministic workflows constructed and analysed 17 additional hypothetical \acp{mof}. Comparison of the resulting structures indicated that changing topology while holding the components fixed could decouple pore size from accessible surface area, providing a design variable distinct from linker substitution.

\paragraph{\textbf{Cross-session persistence}}
In a new session without the original conversation, we asked \grafico{} to retrieve all previously investigated structures containing a specified metal node, group them by topology and compare pore size with accessible surface area. The agent inspected the ontology, composed relation-aware graph queries and wrote Python code to calculate per-topology statistics from the retrieved porosity descriptors. It found that pore-limiting diameter and accessible surface area tended to increase together within a topology, whereas such relation did not persist across topologies because structures with different topologies showed distinct pore-size and surface-area distributions. 
The new analysis reused the identities, relations and calculated properties established by the earlier workflows. Persisted objects therefore remained both addressable and relationally interpretable after the originating graph runs and conversation had ended.

\FloatBarrier

\subsubsection{Bounded execution supports open-ended chemical investigations}

Open-ended scientific questions place more responsibility on the model to decide what should be computed, but the resulting calculations can still operate within an explicit execution contract. We used two prompts as illustrative investigations. In each case, the agent selected the chemical hypothesis, species, charge and spin states, and comparisons at semantic decision points. The GPU4PySCF graph then validated molecular inputs, constrained the calculation sequence, checked optimized structures by frequency analysis and recorded the resulting objects and outputs.

\paragraph{\textbf{Redox mechanism investigation}}
We asked \grafico{} to rationalize the separation between the cathodic and anodic peaks of diphenyl disulfide in dimethylformamide (Supplementary Fig.~\ref{si_fig:redox} and Sec.~\ref{subsubsec:redox}). The agent proposed a minimal network containing \(\mathrm{PhSSPh}\), \(\mathrm{[PhSSPh]^{\bullet-}}\), \(\mathrm{PhS^-}\) and \(\mathrm{PhS^{\bullet}}\), assigned their charge and spin states, and dispatched an independent geometry-optimization and frequency workflow for each species. B3LYP/def2-SVP calculations with SMD(DMF) showed that reduction lengthened the S--S bond from 2.12 to 2.91~\AA. The thiyl/thiolate couple was calculated to lie 0.91~V more positive than reduction of the parent disulfide. These results favour cleavage of the radical anion followed by oxidation of thiolate-derived species and radical recombination, instead of direct reoxidation of \(\mathrm{[PhSSPh]^{\bullet-}}\), and are consistent with an E--C--E/E--C mechanism for the peak separation~\cite{liu1989electrochemical,antonello2003formation,Zhu2023}.

\paragraph{\textbf{Vibrational comparison for an olfaction hypothesis}}
We next asked \grafico{} to use vibrational analysis to explain why a thiol does not smell like ethanol (Supplementary Sec.~\ref{subsubsec:lucaturin}). The agent interpreted the requested thiol analogue as ethanethiol, assigned ethanol and ethanethiol as neutral singlets, and dispatched their calculations independently. Geometry optimization and frequency analysis at the $\omega$B97X-D4/def2-TZVP level placed both structures at local minima. The principal distinction was a shift from the O--H stretch of ethanol at 3879~$\mathrm{cm^{-1}}$ to the S--H stretch of ethanethiol at 2764~$\mathrm{cm^{-1}}$, together with lower-frequency C--S than C--O modes. These differences follow from the altered reduced masses and force constants of the sulfur-containing bonds and define vibrational features that could be compared within Turin's hypothesis~\cite{Turin1996,Brookes2007}. The calculation provides a mechanistic comparison of the two vibrational spectra; it does not validate a vibrational mechanism of olfaction.

In both investigations, model discretion determined scientific scope and interpretation, whereas the graph constrained and recorded the selected operations.

\subsection{From workflow execution to skill-guided protocol authoring}

The preceding studies executed graphs authored by human developers. We next investigated whether a coding agent could author a scientific workflow while preserving the similar contract over typed state, admissible transitions, failures and provenance. In two formative single-run traces conducted during system development, an unconstrained coding agent powered by \texttt{gpt-5} assembled PySCF procedures but omitted required scientific checks and consumed more tokens than the corresponding \grafico{} runs (Supplementary Sec.~\ref{si:bare_agent}). With the subsequent availability of more capable coding agents, including \texttt{gpt-5.6-sol} evaluated below, we investigated workflow authoring under a structured execution contract. We distilled the graph-design principles used throughout the preceding studies into a graph-construction skill specifying typed dataflow, the separation of model-mediated and deterministic operations, explicit fan-out and joins, terminal failure carriers, tests and file-backed provenance. Coding agents loaded this skill while authoring workflows for mining and validating \ac{mof} porosity data.

\subsubsection{Skill-guided agents author inspectable text-mining workflows}

Comparing reported and simulated \ac{mof} properties requires associating each experimental measurement with the correct deposited crystal structure, curating that structure and applying a consistent geometric calculation~\cite{Nandy2021using,Glasby2023DigiMOF}. Variation in terminology, reporting format and structure identifiers makes this association difficult to encode in a fixed script that generalizes across papers~\cite{zheng2023chatgptMOF,Kang2025L2M3,Pruyn2025mofChemUnity}. We therefore supplied \grafico{} (\texttt{gpt-5.6-sol}) and Claude Code (\texttt{claude-fable-5}, with fallback to \texttt{claude-opus-4.8}) with natural language task specifications and two skills. The graph-construction skill supplied the execution contract, whereas a CoRE-MOF skill derived from Ref.~\cite{Zhao2025coremof} supplied structure-curation guidance. The task specified to develop the graph by iterating over five open-access papers~\cite{Evans2022,Xia2024,Li2023,Gu2023,Booth2025}, followed by testing the resulting graph on three papers reserved from development~\cite{Wang2018,Lenzen2019,Cho2020}.

\begin{figure}[!ht]
    \centering
    \includegraphics[width=0.75\linewidth]{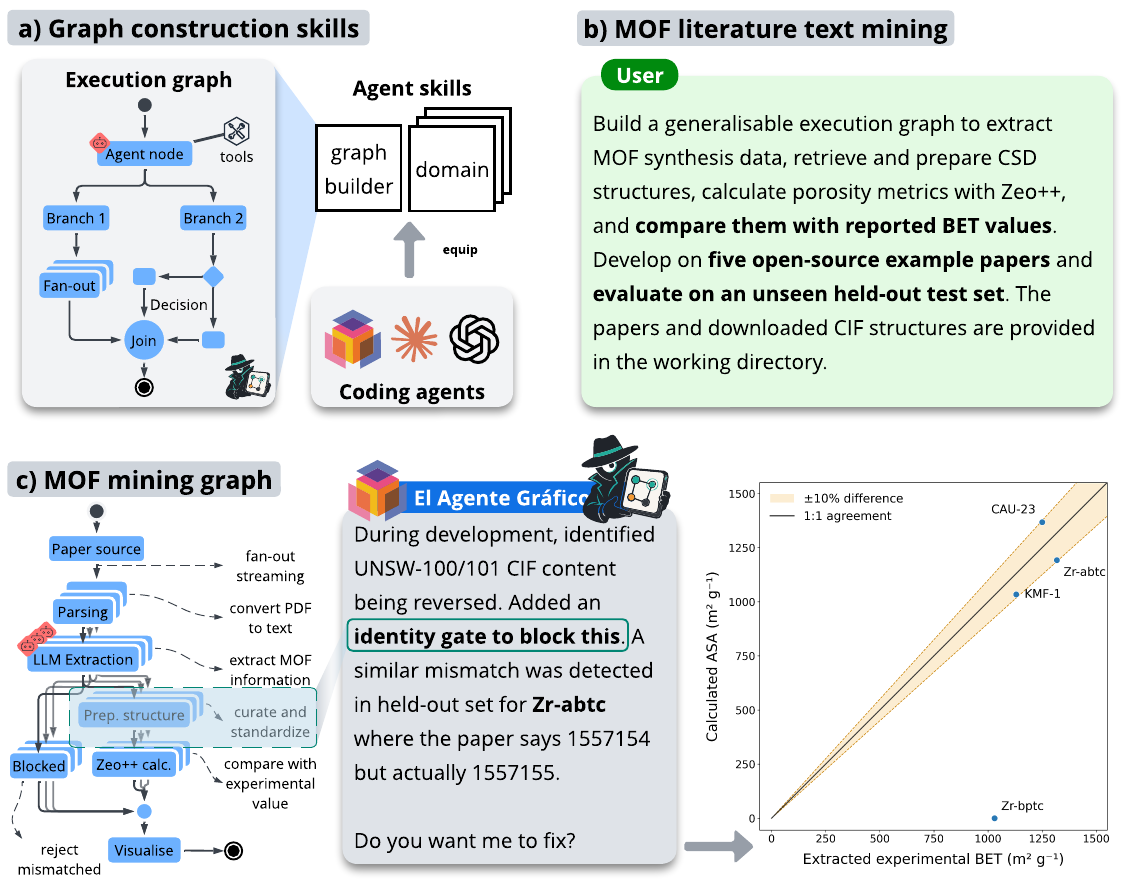}
    \caption{\textbf{Skill-guided authoring, held-out execution and diagnostic revision of an MOF-mining workflow.} (a) Coding agents were equipped with a graph-construction skill and a domain curation skill. (b) The (illustrative) natural-language request to agents to develop the workflow on five papers and evaluate it unchanged on three reserved papers. (c) The \grafico{}-generated graph separates LLM-mediated extraction from deterministic identity checking, structure curation, and Zeo++ calculation for comparison with reported Brunauer--Emmett--Teller (BET) values. An identity gate introduced after detecting a development-set CIF mismatch subsequently blocked an analogous held-out mismatch. Following user authorization, the revision enabled the rerun summarized in the scatter plot. Complete prompts and revision histories are reported in Supplementary Sec.~\ref{si:mof_mining}.
    }
    \label{fig:mof_mining}
\end{figure}

Both agents independently generated typed execution graph workflows that confined model use to information extraction while leaving structure preparation, Zeo++ calculations~\cite{Willems2012zeopp} and comparison with reported values deterministic (\autoref{fig:mof_mining}). During development, \grafico{} encoded a general paper--CIF identity gate after detecting swapped CIF contents; when applied unchanged to the reserved papers, the gate identified an analogous cross-assignment that Claude Code missed. After the held-out outcomes were observed, the user asked \grafico{} to explain the blocked identities and then authorized the proposed repair. Complete execution outcomes and subsequent revision histories are reported in Supplementary Sec.~\ref{si:mof_mining}.

For records whose structure assignments aligned, the post-diagnostic calculations were close across the two generated workflows. For CAU-23, \grafico{} and Claude Code calculated 1368 and 1397~\(\mathrm{m^2\,g^{-1}}\), respectively, compared with a reported \ac{bet} surface area of 1250~\(\mathrm{m^2\,g^{-1}}\). For KMF-1 deposition 1984701, the corresponding values were 1034 and 1041~\(\mathrm{m^2\,g^{-1}}\), compared with 1130~\(\mathrm{m^2\,g^{-1}}\). These calculations differed from the reported values by 7.9--11.8\%. As geometric accessible surface area and an experimental \ac{bet} fit are not the same observable, the comparison evaluates the complete association, curation and calculation chain (Supplementary Sec.~\ref{si:mof_mining}).

The larger differences exposed where supplied procedural knowledge entered the generated code. The CoRE-MOF skill instructed agents to select the least aggressive chemically valid candidate and not to use an all-solvent-removed structure without checking coordination and connectivity. \grafico{} preferentially selected the free-solvent-removed candidate, whereas Claude Code selected the all-solvent-removed candidate whenever available.

More consequentially, the identity gate distinguished computational success from scientific validity. A syntactically valid CIF can pass structure preparation and yield a Zeo++ result while representing a different material from the one described in the article. By encoding paper--CIF consistency as a pre-calculation condition, the generated graph from \grafico{} made this distinction and its downstream consequences inspectable.

At the reported trace scopes and public API prices, the \grafico{} trajectory, including initial authoring, explanation of the identity conflicts and the user-authorized repair, had an API-equivalent model cost of US\$3.13, compared with US\$28.10 for the Claude Code trajectory through workflow completion (Supplementary Table~\ref{tab:si_mof_trace_cost}).

\FloatBarrier

\section{Discussion}

The case studies distinguish three levels of validity in workflows executed by scientific agents. Structural validity concerns whether the workflow state conforms to declared schemas; procedural verifiability concerns whether execution routes and outcomes can be checked against an executable protocol; and scientific validity concerns whether the evidence and interpretations adequately address the research question. Schema validation enforces structural validity, while runtime traces support procedural verification. Scientific validity must instead be assessed with reference to domain knowledge and empirical evidence. In self-driving laboratories, the same distinction calls for a safety harness that translates model intent into executable experiments, monitor actions and retain the resulting evidence~\cite{leong2025steering,openai2026sdl,Chen2026autonomyHarness}.

Externalizing execution made the first two levels explicit and reduced the need for conversational coordination. In the six-exercise benchmark, moving validation and state management into the runtime demonstrated lower model cost and shorter wall-clock time. Native objects carried active computational state, whereas the \ac{kg} retained selected objects, relations and provenance for later use. This division resembles object-oriented agent designs that keep typed references and deterministic operations outside the model loop~\cite{Furgale2026OOAgents}. Execution-graph topology further constrains the protocol by specifying which state transitions may follow one another.

Agent-authored protocols introduce a second object of verification: the generated graph itself. In the authoring studies, a shared skill constrained procedural design without prescribing a unique implementation. Both coding agents implemented the requested stages, but they adopted different curation choices and validation checks. Their graphs, tests and provenance records made these differences inspectable. Although the skill provided reusable design rules, the generated artifact ultimately determined the outcome of execution. Revisions to reusable protocols should therefore pass interface checks, domain invariants, regression tests and held-out reference cases. Versioning and rollback can preserve previously validated states.

Going forward towards globally connected laboratories in which \ac{llm} agents translate scientific requests into remotely executed protocols~\cite{Gaidimas2026Reimagining}. To build on existing demonstrations of cloud-coordinated multi-laboratory operation~\cite{StriethKalthoff2023delocalized,bai2024dynamic}, \ac{llm}-mediated control introduces additional requirements for operation scheduling and rate limiting, together with scoped access to instrument state and scientific context. Further requirements for safe, distributed and verifiable laboratory coordination are outlined in Supplementary Sec.~\ref{si:runtime_roadmap}.

\section{Conclusions}
Reliable scientific autonomy requires explicit contracts governing how probabilistic model decisions become computational or physical actions. Compared with approaches based on multi-agent coordination or unconstrained iterative code generation, \elagenteG{} represents scientific state and admissible transitions explicitly in typed execution graphs, grounding model-generated scientific decisions in validated execution. Within each graph, the runtime manages heavyweight objects and deterministic validation, while the model is only invoked at the semantic decision points. This division of labour reduces the context burden and leaves a traceable record of the workflow. Across six university-level quantum chemistry exercises, this design improved task performance while reducing model cost by 80\% and wall-clock time by a factor of 4.5 relative to our previous multi-agent architecture.
Redox and olfaction studies further showed that offloading execution allows the agent to focus on defining the chemical scope and computational pathway of open-ended investigations.
We then encoded graph construction as a reusable skill for coding agents and demonstrated a dynamic workflow of literature mining and computational calculations. By shifting model inference from orchestration towards scientific reasoning, \grafico{} provides a concrete path for scaling scientific agents to more complex and adaptive research systems.

\section{Methods}

\subsection{\grafico{} infrastructure}
\label{subsec:grafico_infrastructure}

The object graph mapper (OGM) in this work is a customized version of the World Avatar~\cite{krdzavac2019ontology,Pascazio2023ontospecies,Zhou2026ontology} Python package \texttt{twa}~\cite{Bai2025twa}, refactored to provide a fully Python-native interface and to align more closely with Pydantic data structures and usage patterns. This OGM provides a bidirectional serialization layer in which Python classes are mapped to classes and relationships in ontologies, with strict type enforcement. Key custom features include a lazy-loading mechanism that efficiently retrieves subgraph data on demand when processing large-scale ontologies, and specialized support for serializing NumPy arrays using \texttt{numpydantic}. By integrating with \texttt{rdflib} and \texttt{pydantic-ai}, the OGM ensures that complex chemical data structures are represented with formal semantic validation while maintaining idiomatic Python interoperability for downstream integration with \ac{llm}-based agents. For the graph-authoring experiments, coding agents were instantiated using \texttt{pydantic-deep} (v0.3.22), a Pydantic AI-based agent harness. The framework provided reusable skill loading, tool orchestration and filesystem-based code development for constructing and testing execution graphs.

The computational chemistry agent was instantiated using the \texttt{pydantic-ai} framework, configured with a system prompt that injects the workflow constraints and domain-specific instructions. Tools are strictly typed using Pydantic models and registered via a tool decorator or through the tools list, enabling the \ac{llm} to invoke complex functions with validated schemas. The framework's dependency injection system is critical for our architecture, passing a \texttt{GraficoDeps} object to tools that provide access to shared resources like the frontend and \ac{kg} endpoints. We also leverage pydantic-ai's model context protocol (MCP) support to integrate external execution environments (\textit{e.g.}, a sandboxed Python runner) and its \texttt{ModelRetry} mechanism to handle tool execution failures by providing feedback loops back to the model. The \acp{llm} were configured with temperatures of 0 for non-reasoning models and 1 for models using explicit reasoning modes.

Custom GPU scheduling was developed for GPU4PySCF to prevent CUDA context initialization conflicts and illegal memory access patterns common in multi-threaded GPU4PySCF jobs. The concurrency levels were managed by a thread-safe token queue, defaulting to three execution slots per GPU to optimize device utilization while preventing memory exhaustion. Upon task submission, a dedicated scheduler assigns an available GPU identifier and execution slot, filtering visible devices through environment variables. Each workflow executes in a separate OS-level child process while sharing the same CUDA driver context. Inter-process communication is handled via pickled execution payloads. Integrated distributed tracing (via logfire) provides real-time observability into slot occupancy.

The evaluation benchmark was developed using \texttt{pydantic-evals}. The system assesses the agent's performance by extracting structured workflow results from tool outputs and comparing them against the ground truth. The evaluation methodology employs a multi-dimensional scoring metric that verifies correctness across several chemical properties: (i) parameter integrity, to ensure that the selected functional, basis set, charge, and multiplicity match the intended calculation; (ii) energetic accuracy, which validates total energies against reference values within a threshold of 0.01 Ha; (iii) structural fidelity, quantified by the RMSD of atomic positions relative to reference geometries within a threshold of 0.15~\AA, utilizing \texttt{spyrmsd} to account for symmetry and atom indexing permutations; and (iv) property analysis, which checks derived observables such as dipole moments, HOMO-LUMO gaps (threshold of 0.1 Ha), point group symmetry, and vibrational frequencies (ensuring no imaginary modes for stable minima). 
Scores are aggregated and normalized across these dimensions to yield a composite accuracy metric for each test case, with distributed tracing (via \texttt{logfire}) used to monitor agent reasoning and tool-invocation patterns during execution. Token counts and cost information were taken from the \texttt{logfire} records. For evaluations performed prior to the release of
\texttt{pydantic-evals==v1.23.0} (2025-11-25), token counts recorded by \texttt{logfire} were corrected
to account for a double-counting bug
(\url{https://github.com/pydantic/pydantic-ai/issues/3529}),
which was resolved in the v1.23.0 release.

\subsection{Computational tools}
\label{sec:computational_tools}

The project employed \texttt{uv} for a virtual environment using Python 3.12.12. All computational packages were installed in the same environment to enable single-agent operation using a shared Python runtime. To resolve dependency conflicts and enable custom functions, some packages were forked from their original repositories. Specific modifications are noted for each package. Where feasible and once the interfaces stabilize, we intend to upstream generalizable improvements to the original repositories via pull requests.

\begin{enumerate}

\item \textbf{pydantic and numpydantic}: The customized object graph mapper is implemented using \texttt{pydantic} (v2.13.4) for basic typed classes and \texttt{numpydantic} (v1.6.11) for serialisation of NumPy arrays.

\item \textbf{pydantic-ai family}: The agentic framework and workflow execution are implemented using \texttt{pydantic-ai} (v1.102.0), with the coding agents instantiated using \texttt{pydantic-deep} (v0.3.22). Structured workflow graphs are managed via the \texttt{pydantic-graph} component, with customized base node classes for dynamic routing over typed node definitions. The framework was iteratively updated during development to incorporate new features and was stabilized at version v1.102.0. Execution traces, tool invocations, and structured state transitions were instrumented using \texttt{logfire} (accessed via \texttt{pydantic-ai}) to enable fine-grained observability, debugging, and post-hoc analysis of agent behaviour. For web search, the agent utilised the built-in tool provided by the \ac{llm} providers whenever possible, with a fallback DuckDuckGo search tool provided by  \texttt{pydantic\_ai}.

\item \textbf{PySCF family}: Density functional theory calculations were performed using the \texttt{pyscf} (v2.10.0), with \texttt{gpu4pyscf-cuda12x} (v1.5.0) for GPU acceleration on CUDA 12.x hardware and \texttt{pyscf-dispersion} (v1.3.0) for semiclassical dispersion corrections \cite{grimmeConsistentAccurateInitio2010,grimmeEffectDampingFunction2011,caldeweyherExtensionD3Dispersion2017,caldeweyherGenerallyApplicableAtomiccharge2019,wittmannExtensionD3D42024}.

\item \textbf{PubChemPy}: The molecular generation from names or SMILES was implemented using \texttt{pubchempy} (v1.0.4)~\cite{kim2016pubchem,Swain_PubChemPy_2017}.

\item \textbf{RDKit}: The molecular generation from SMILES was implemented using \texttt{rdkit} (v2025.3.5)~\cite{rdkit_2025_03}.

\item \textbf{OpenBabel}: In cases where RDKit failed, \texttt{openbabel-wheel} (v3.1.1.22) (a prebuilt wheel through PyPI) was used~\cite{o2011open,openbabel_wheel}.

\item \textbf{QCElemental}: Elemental data, physical constants, and unit conversions were standardized using \texttt{qcelemental} (v0.29.0), ensuring consistent representation of molecular quantities across electronic-structure calculations and workflow components.

\item \textbf{QCG and CREST}: The conformer and rotamer ensemble sampling tool \ac{crest} generates ranked ensembles of low-energy conformers for a given three-dimensional molecular structure \cite{prachtAutomatedExplorationLowenergy2020,prachtCRESTProgramExploration2024}.
Using iterative metadynamics simulations \cite{grimmeExplorationChemicalCompound2019} in combination with conventional molecular dynamics, \ac{crest} systematically samples the potential energy surface and explores the accessible conformational space.
In this work, we implemented its GPU-accelerated variant described in Ref.~\cite{Steinbach2025crestGPU} (commit \texttt{75135dd}).
For the generation of explicitly solvated molecular structures, we use the \ac{qcg} algorithm \cite{Spicher2022QCG} implemented in \texttt{crest-3.0.2} together with \texttt{xtb-6.7.1}.

\item \textbf{PORMAKE}: PORMAKE~\cite{Lee2021pormake} enables topology-aware assembly of hypothetical \acp{mof} from predefined building blocks (metal nodes and organic linkers). Building blocks are assigned to topology connection sites using RMSD-based matching of local coordination environments (RMSD $\leq$ 0.3~\AA{}). PORMAKE's experimental decomposition routines are used to extract provisional building blocks from existing CIF files. We use a GitHub snapshot of PORMAKE pinned to commit \texttt{639caad} (\url{https://github.com/Sangwon91/PORMAKE}), ensuring reproducible behaviour.

\item \textbf{CoRE-MOF-Tools}: CoRE-MOF-Tools (package: \texttt{coremof-tools}, v0.3.1)~\cite{Zhao2025coremof} is used for preprocessing experimental CIF files, including solvent removal and structural normalization, and for interfacing with Zeo++ to compute porosity-related descriptors. We use a fork of CoRE-MOF-Tools (\url{https://github.com/jb2197/CoRE-MOF-Tools}) in which modifications are confined to \texttt{setup.py} to resolve dependency incompatibilities with the broader software environment. Specifically, strict version pins are relaxed to compatible lower bound constraints (\textit{e.g.}, \texttt{scikit-learn==1.3.2} to \texttt{scikit-learn>=1.3.2}). The fork also redirects the \texttt{MOFClassifier}~\cite{Zhao2025MOFClassifier} dependency to a compatible fork (\url{https://github.com/jb2197/MOFClassifier}), where analogous changes are applied in \texttt{setup.py}: strict version pins for \texttt{numpy}, \texttt{torch}, \texttt{pymatgen}, \texttt{scikit-learn}, \texttt{tqdm}, and \texttt{pandas} are relaxed to lower-bounded constraints, the \texttt{requests} package is added to \texttt{install\_requires}, and an unused \texttt{requests} import is removed from the setup script.

\item \textbf{MOFid}: MOFid (upstream v1.1.0; forked and released as v1.1.1 with mofid-v2 utilities) provides the \texttt{convert\_ase\_pymat} function for ASE-to-pymatgen conversion, used during building block deduplication. This conversion enables structural comparison via pymatgen's \texttt{StructureMatcher}, as part of a second-stage verification when PORMAKE's experimental hash-based decomposition yields multiple candidate metal nodes or organic linkers. We use a fork of MOFid (\url{https://github.com/swanickt/mofid}) that corrects the upstream package distribution to make the mofid-v2 analysis code programmatically accessible. Specifically, the fork updates \texttt{setup.py} to include the \texttt{mofid\_v2} modules in the installable package set, adds missing \texttt{\_\_init\_\_.py} files, refactors the node-grouping logic into an importable library module (\texttt{group\_node\_by\_structure\_library.py}), and updates dependency declarations to support reliable installation in Python environments. These changes remove script-style side effects present in the upstream implementation and enable direct programmatic use within Python workflows.

\item \textbf{pymatgen}: pymatgen (v2024.8.9)~\cite{ong2013python} provides structure-level equivalence checking throughout the workflow. For building block deduplication, \texttt{StructureMatcher} compares candidate structures converted from ASE via MOFid's \texttt{convert\_ase\_pymat} utility. For general structure validation and matching in the \texttt{ConceptualAtoms} data model, pymatgen's native \texttt{AseAtomsAdaptor} performs ASE-to-pymatgen conversion, with \texttt{StructureMatcher} used for periodic structures and \texttt{MoleculeMatcher} for non-periodic molecules. Together, these capabilities support deduplication, \ac{kg} queries for equivalent components, and structure validation.

\item \textbf{CrystalNets.jl}: CrystalNets.jl (v1.1.0)~\cite{CrystalNets}, accessed via \texttt{juliacall} (v0.9.29), is used for topological classification of \ac{mof} structures. The package determines the underlying net topology, dimensionality, and catenation from input CIF files. We employ the \texttt{SingleNodes} clustering mode, in which each identified building block is mapped to a single vertex.

\item \textbf{Zeo++}: Zeo++ (v0.3)~\cite{Willems2012zeopp} is used for porosity analysis, including largest cavity diameter (LCD), pore limiting diameter (PLD), largest free sphere path diameter (LFPD), accessible surface area, pore volume, and framework dimensionality. Zeo++ is accessed through the CoRE-MOF-Tools interface and is obtained from the \texttt{zeo++/0.3} module directly loaded on the trillium-gpu HPC compute node.

\item \textbf{MACE}: MACE-torch (v0.3.14)~\cite{batatia2022mace} provides machine-learned interatomic potentials for geometry optimization. We employ MACE-MP-MOF0 (\href{https://github.com/ddmms/data/blob/9bb729a1a32a3700ecc64f4780f743a7409840af/mace-mof-0/v2/mof-omat-0-v2.model}{mof-omat-0-v2}) to perform structure relaxation prior to porosity analysis. We use a fork (\url{https://github.com/jb2197/mace}) that relaxes the strict dependency constraint \texttt{e3nn==0.4.4} to an unpinned version (\texttt{e3nn}), and further in our environment, we pin \texttt{e3nn==0.5.0} to ensure compatibility with other packages (MatterSim).

\item \textbf{MatterSim}: MatterSim (v1.2.0)~\cite{yang2024mattersim} is supported as an alternative machine-learned force-field backend for geometry optimization within the workflow.

\item \textbf{Orb models}: The Orb models (v0.5.4)~\cite{rhodes2025orb} provide an additional machine-learned interatomic potential option for geometry relaxation.

\item \textbf{ASE}: The Atomic Simulation Environment (v3.26.0)~\cite{larsen2017atomic} serves as the common data structure for atomic configurations throughout the workflow. Building blocks from the PORMAKE decomposer, structures for \ac{mlip} optimization, and final \ac{mof} geometries are all represented as ASE \texttt{Atoms} objects.

\item \textbf{RDFLib and SPARQLWrapper}: RDFLib (v7.1.4) is used for RDF serialization and local graph operations, while SPARQLWrapper (v2.0.0) handles remote SPARQL queries against the \ac{kg}. Together, they enable combinatorial search algorithms, equivalence queries for component deduplication, and persistence of workflow results.

\item \textbf{Blazegraph}: Blazegraph (engine v2.1.6) serves as the triplestore backend, hosting the persistent \ac{kg} and exposing a SPARQL endpoint for querying \ac{mof} components, topologies, and compatibility relationships. The \ac{kg} is deployed using the Cambridge CARES Docker image \texttt{ghcr.io/cambridge-cares/blazegraph:1.2.0} (container registry: \url{https://github.com/orgs/cambridge-cares/packages/container/package/blazegraph}), which provides a standardized containerized runtime consistent with the World Avatar \ac{kg} infrastructure.

\item \textbf{mcp-run-python}: Sandboxed Python code execution \texttt{mcp-run-python} was used as a model context protocol server, where a specific forked version (\url{https://github.com/jb2197/mcp-run-python/releases/tag/0.0.22.2-file}) was used to enable output files.

\item \textbf{PythonREPL}: The Python code execution tool from \texttt{langchain-experimental} (v0.4.1) was used.

\end{enumerate}

\subsection{Graphical user interface}
\label{subsec:gui}
The Graphical User Interface (GUI) is implemented using the React framework. Communication between the frontend and backend is facilitated by the \texttt{GraphChat} SDK, a custom Python library that utilizes the WebSocket protocol for efficient data exchange. The key features of the chat interface include real-time streaming of agent responses, visibility into tool execution logs, and display of the agent's reasoning steps, all implemented using the Pydantic AI framework and rendered via \texttt{pycrdt} for synchronization and real-time updates. Furthermore, the interface dynamically renders a molecular viewer upon agent activation. When the agent generates molecules, the molecule viewer uses the corresponding conceptual atom \ac{iri} as a unique identifier. 
The molecule viewer supports both molecular and periodic systems, with interactive 3D rendering, optional atom selection, and the ability to send snapshots (captured images) to the \ac{llm}. 
The molecule viewer visualizes the molecular trajectory in real-time and allows users to download structures (and trajectories) in standard formats such as (extended) \texttt{xyz} and CIF. 
Additionally, plots generated by calls to the code execution tool are displayed directly in the chat interface for immediate interpretation. The frontend also supports dynamic \ac{llm} model switching, allowing users to experiment with different models on the fly without interrupting the ongoing session.


%% file: includes/include-acknowledgement.tex
The authors thank Yiqun~Bian for the helpful discussions in data analysis, Dr.~Pit~Steinbach for the helpful suggestions in setting up the GPU-accelerated implementation of CREST, and Samuel Chen for the feedback on the manuscript.
J.B. acknowledges funding from the Eric and Wendy Schmidt AI in Science Postdoctoral Fellowship Program, a program by Schmidt Futures.
A.A. acknowledges support from King Abdullah University of Science and Technology (KAUST), Kingdom of Saudi Arabia, for the KAUST Ibn Rushd Postdoctoral Fellowship.
M.M. was partially supported through a collaborative partnership with Merck KGaA, Darmstadt, Germany.
Y.K. acknowledges support from the CIFAR AI Safety Catalyst Award (Catalyst Fund Project \#CF26-AI-001).
N.Y. acknowledges the support from JSPS KAKENHI Grant Number JP25K21333.
T.W.K. acknowledges the support of the Vector Distinguished Postdoctoral Fellowship.
A.A.-G. thanks Anders~G.~Fr{\o}seth for his generous support. A.A.-G. also acknowledges the generous support of Natural Resources Canada and the Canada 150 Research Chairs program. This work was supported by the AI2050 program of Schmidt Sciences.
\acknowDARPA
\acknowAC
\acknowGEN{SciNet HPC Consortium (\url{https://scinethpc.ca/})}
\acknowSciNet{Trillium}

%% file: includes/include-appendix.tex
\setcounter{table}{0}
\renewcommand{\thetable}{S\arabic{table}}%
\setcounter{figure}{0}
\renewcommand{\thefigure}{S\arabic{figure}}%
\setcounter{lstlisting}{0}
\renewcommand{\thelstlisting}{S\arabic{lstlisting}}%
\setcounter{algorithm}{0}
\renewcommand{\thealgorithm}{S\arabic{algorithm}}%

\section{List of tools}
\input{includes/si_tex/si-agent-tools}

\newpage
\section{Typed execution graph abstraction}
\label{si:execution_graph}
\input{includes/si_tex/si-execution-graph}

\newpage
\section{GraphChat user interface}
\label{si:graphchat}
\input{includes/si_tex/si-graphchat}

\newpage
\section{University-level quantum chemistry benchmark}

\subsection{LLM model specifications}
\label{si:llm_model_configs}
\input{includes/si_tex/si-benchmark-llms}

\newpage
\subsection{User prompts}
\label{si:benchmark_questions}
\input{includes/si_tex/si-benchmark-questions}

\newpage
\subsection{Rubrics for LLM judge}
\label{si:rubric}
\input{includes/si_tex/si-benchmark-rubrics}

\newpage
\subsection{Pass k analysis}
\label{si:pass_k_results}
\input{includes/si_tex/si-benchmark-pass-k}

\newpage
\subsection{Statistics plots}
\label{si:statistics_plots}

\input{includes/si_tex/si-benchmark-plots}

\newpage
\subsection{Bare LLM agent}
\label{si:bare_agent}
\subsubsection{Summary of results}
\label{si:bare_agent_summary}
\input{includes/si_tex/si-bare-agent-summary.tex}

\newpage
\subsubsection{Bare agent transcript: inorganic compound: level 1}
\label{si:chat_bare_agent_inorganic}
Gráfico was configured with \texttt{gpt-5} (temperature = 1, reasoning effort set to \texttt{low}) for this run, with access to the tools \texttt{Python\_REPL} for code execution and \texttt{duckduckgo\_search} for web search.

For the complete chat transcript, please refer to GitHub repo: \url{https://github.com/jb2197/ElAgenteGrafico-ChatTranscript}.

\subsubsection{Bare agent transcript: pka prediction: level 2}
\label{si:chat_bare_agent_pka}
Gráfico was configured with \texttt{gpt-5} (temperature = 1, reasoning effort set to \texttt{low}) for this run, with access to the tools \texttt{Python\_REPL} for code execution and \texttt{duckduckgo\_search} for web search.

For the complete chat transcript, please refer to GitHub repo: \url{https://github.com/jb2197/ElAgenteGrafico-ChatTranscript}.

\newpage

\section{Use case extension 1: Boltzmann-weighted spectroscopic properties}
\label{si:crest_pyscf}

\input{includes/si_tex/si-crest-pyscf-summary.tex}

\subsection{Boltzmann-weighted spectroscopic properties in implicit solutions}
\label{subsubsec:implicit_solvation}

Gráfico was configured with \texttt{gpt-5.4} (temperature = 1, reasoning effort set to \texttt{medium}) for this run. The responses API endpoint was used here. The routing agent was configured with \texttt{gpt-4.1} (temperature = 0.1 using the chat-completion API endpoint).

For the complete chat transcript, please refer to GitHub repo: \url{https://github.com/jb2197/ElAgenteGrafico-ChatTranscript}.

\subsection{Boltzmann-weighted spectroscopic properties with explicit solvation}
\label{subsubsec:explicit_solvation}

Gráfico was configured with \texttt{gpt-5.4} (temperature of 1, `medium' reasoning effort, and `detailed' reasoning summary) for this run. The responses API endpoint was used here. The routing agent was configured with \texttt{gpt-4.1} (temperature = 0.1 using the chat-completion API endpoint).

For the complete chat transcript, please refer to GitHub repo: \url{https://github.com/jb2197/ElAgenteGrafico-ChatTranscript}.

\newpage
\section{Use case extension 2: Exploration of metal-organic frameworks design space}
\label{si:mof_exploration}

\subsection{{\ac{mof} execution graph node implementations}}
\input{includes/si_tex/si-mof-workflow}

\subsection{Chat transcripts}
\input{includes/si_tex/si-mof-summary}

\subsubsection{Three-stage build and explore}
\label{subsubsec:mof_single_prompt}
Gráfico was configured with \texttt{gpt-5.2} (temperature of 1, `medium' reasoning effort, and `detailed' reasoning summary) for this run. The responses API endpoint was used here. The routing agent was configured with \texttt{gpt-4.1} (temperature = 0.1 with the chat-completion API endpoint).

For the complete chat transcript, please refer to GitHub repo: \url{https://github.com/jb2197/ElAgenteGrafico-ChatTranscript}.

\subsubsection{SPARQL exploration}
\label{subsubsec:mof_sparql_query_explore}

Gráfico was configured with \texttt{gpt-5.2} (temperature of 1, `medium' reasoning effort, and `detailed' reasoning summary) for this run. The responses API endpoint was used here.

For the complete chat transcript, please refer to GitHub repo: \url{https://github.com/jb2197/ElAgenteGrafico-ChatTranscript}.

\subsubsection{Interactive exploration}
\label{si:mof_interactive_exploration}
During development, users interacted with \grafico{} through natural language over five messages to incrementally construct and explore a knowledge graph populated with existing CIF-based \acp{mof}, hypothetical structures assembled using PORMAKE~\cite{Lee2021pormake}, and newly inferred candidates generated via combinatorial searches over building blocks and topologies instantiated from the previous steps. Rather than executing static scripts, the agent acted as a scientific copilot that translated user intent into a minimal sequence of executable workflow stages, dynamically invoking domain-specific \ac{mof} workflows, ontology introspection tools, and knowledge-graph queries while preserving full provenance of intermediate and final results.

Across multiple interaction rounds, the agent consistently parsed user intent into a minimal sequence of executable workflow stages that interacted with the knowledge graph, while also demonstrating effective context management through in-memory Python objects. Depending on the user intent and intermediate results, the router agent dynamically determined the next workflow nodes, for example, by checking whether a \ac{mof} had already been constructed or analyzed. This enabled coordinated orchestration of the \ac{mof} execution graph and knowledge graph tools across different tasks, while incrementally building a knowledge graph of \ac{mof} instances. In this proof-of-concept, we constrain \ac{mof} construction to one metal node and one organic linker. Nevertheless, the modular graph node architecture and the execution graph design enable straightforward extension of the system to more complex reticular material design workflows.

Gráfico was configured with \texttt{gpt-5.2} (temperature of 1, `medium' reasoning effort, and `detailed' reasoning summary) for this run. The responses API endpoint was used here. The routing agent was configured with \texttt{gpt-4.1} (temperature = 0.1 with the chat-completion API endpoint).

For the complete chat transcript, please refer to GitHub repo: \url{https://github.com/jb2197/ElAgenteGrafico-ChatTranscript}.

\newpage
\section{Use case extension 3: Towards open-ended research style questions}
\label{si:redox}

\subsection{Redox mechanism investigation}
\label{subsubsec:redox}
Gráfico was configured with \texttt{openai-responses:gpt-5.4} (`medium' reasoning effort and `detailed' reasoning summary) for this run. The routing agent was configured with \texttt{gpt-4.1} (temperature = 0.1).

\input{includes/si_tex/si-redox-short}

For the complete chat transcript, please refer to GitHub repo: \url{https://github.com/jb2197/ElAgenteGrafico-ChatTranscript}.

\subsection{Luca-Turin smell theory}
\label{subsubsec:lucaturin}
Gráfico was configured with \texttt{openai-responses:gpt-5.4} (`medium' reasoning effort and `detailed' reasoning summary) for this run. The routing agent was configured with \texttt{gpt-4.1} (temperature = 0.1).

\input{includes/si_tex/si-luca-short}

For the complete chat transcript, please refer to GitHub repo: \url{https://github.com/jb2197/ElAgenteGrafico-ChatTranscript}.

\newpage
\section{Skill-guided execution graphs}
\subsection{Mining the MOF literature}
\label{si:mof_mining}
\input{includes/si_tex/si-skills-mof-mining}

\newpage
\section{Prospective runtime extensions}
\label{si:runtime_roadmap}

\begin{table}[!ht]
    \centering
    \small
    \caption{\textbf{Current evidence and open requirements for scientific execution runtimes.} Each row separates a capability examined in this study from an unresolved requirement and a future evaluation that could test it. The ordering does not imply a maturity sequence.}
    \label{tab:si_runtime_roadmap}
    \begin{tabularx}{\textwidth}{@{}>{\raggedright\arraybackslash}p{0.14\textwidth}>{\raggedright\arraybackslash}X>{\raggedright\arraybackslash}X>{\raggedright\arraybackslash}X@{}}
        \toprule
        Dimension & Evidence in this study & Open requirement & Future evaluation \\
        \midrule
        Protocol semantics & Typed state, admissible transitions and explicit failure carriers & Static and domain-specific conformance checks, with versioned schema migration & Held-out protocol suites and regression testing across workflow revisions \\
        \addlinespace
        Protocol authoring & Skill-guided graph generation by two coding agents & Evidence-gated revision of graphs and skills & Repeated cross-model authoring with independent scientific audit \\
        \addlinespace
        Execution substrate & Native-object transfer, persistent relational state and parallel graph lanes & Isolated, resource-aware and distributed execution that preserves identity and provenance & Cross-environment replay, recovery and resource accounting \\
        \addlinespace
        Physical grounding & Digital-twin-to-robot protocol transfer and conversion of laboratory feedback into a tested constraint & Typed multimodal feedback, calibration and uncertainty-aware twin updates & Repeated closed-loop experiments on held-out targets \\
        \bottomrule
    \end{tabularx}
\end{table}

\begin{figure}[!ht]
    \centering
    \includegraphics[width=\linewidth]{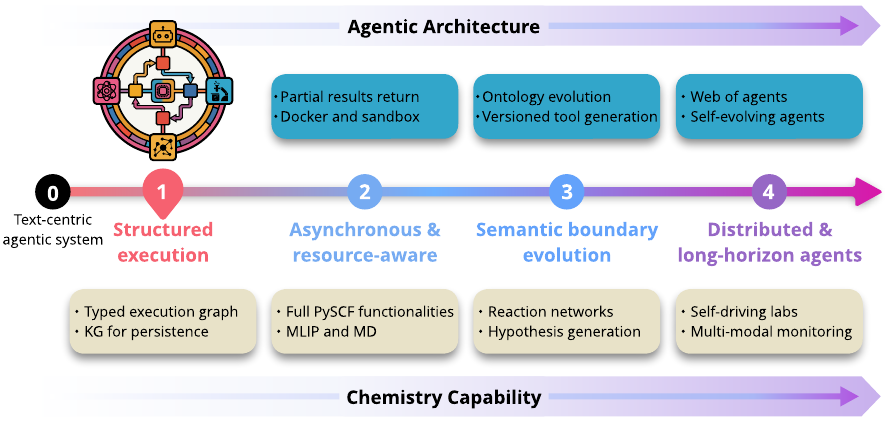}
    \caption{Current stage and roadmap of future work showing: (1) structured execution of typed execution graphs and tools (as implemented in the current \grafico{} system); (2) asynchronous and resource-aware execution to enable proactive agents; (3) semantic boundary evolution allowing rapid extension of tools and ontologies; (4) long-horizon agents towards a distributed network of AI scientists.}
    \label{fig:roadmap}
\end{figure}

%% file: includes/si_tex/si-agent-tools.tex

\noindent
\begin{tabular}{p{0.22\linewidth} p{0.72\linewidth}}
\hline
\textbf{Tool Name} & \textbf{Tool description received by agent/LLMs} \\
\hline
\texttt{run\_mof\_workflow} &
Run the MOF workflow graph for building, optimizing, and/or analyzing MOFs.

This workflow uses dynamic AI-based routing to chain together multiple stages
based on the user's intent. The workflow is NOT limited to a single operation---
it can automatically proceed through multiple stages (e.g., download -> decompose
-> build -> optimize -> analyze).

WORKFLOW STAGES (can be combined automatically):
\begin{itemize}[label=\tiny$\bullet$, noitemsep, topsep=0pt, leftmargin=*]
  \item Download: Fetch CIF from CSD/CoRE-MOF database
  \item Decompose: Extract topology and building blocks from CIF
  \item Search: Query knowledge graph for compatible MOF combinations
  \item Build: Construct MOF structure from components
  \item Optimize: Geometry relaxation via MLFF (MACE-MOF, etc.)
  \item Analyze: Zeo++ porosity analysis
\end{itemize}

SAMPLE WORKFLOWS:

Build directly from PORMAKE components

Required:
\begin{itemize}[label=\textendash, noitemsep, topsep=0pt, leftmargin=*]
  \item \texttt{topology\_pormake\_id} (e.g., "pcu", "dia)
  \item \texttt{node\_pormake\_id} (e.g., "N409", "N111")
  \item \texttt{linker\_pormake\_id} (e.g., "E1", "E42")
\end{itemize}
Optional:
\begin{itemize}[label=\textendash, noitemsep, topsep=0pt, leftmargin=*]
  \item \texttt{mof\_name}
\end{itemize}

Example user intents:
\begin{itemize}[label=\textendash, noitemsep, topsep=0pt, leftmargin=*]
  \item "Build a MOF with topology Z using metal node X and organic linker Y."
  \item "Construct this specific MOF (I know the building blocks)."
\end{itemize}

CSD/CoRE-MOF Refcode -> Download, then automatically routed based on user intent (summarised\_user\_query)

Required:
\begin{itemize}[label=\textendash, noitemsep, topsep=0pt, leftmargin=*]
  \item \texttt{csd\_refcode}
\end{itemize}
Optional:
\begin{itemize}[label=\textendash, noitemsep, topsep=0pt, leftmargin=*]
  \item \texttt{mof\_name}
\end{itemize}

Example user intents:
\begin{itemize}[label=\textendash, noitemsep, topsep=0pt, leftmargin=*]
  \item "What is the topology of MOF ABAVIJ?"
  \item "What is the surface area of this CSD structure?"
  \item "Optimize the MOF with refcode ABAVIJ."
\end{itemize}

Process a local CIF file

Required:
\begin{itemize}[label=\textendash, noitemsep, topsep=0pt, leftmargin=*]
  \item \texttt{input\_cif\_filename}
\end{itemize}
Optional:
\begin{itemize}[label=\textendash, noitemsep, topsep=0pt, leftmargin=*]
  \item \texttt{mof\_name}
\end{itemize}

Typical user intents:
\begin{itemize}[label=\textendash, noitemsep, topsep=0pt, leftmargin=*]
  \item "Here is a CIF. What's its topology?"
  \item "Optimize this MOF."
  \item "Run Zeo++ analysis on this structure."
\end{itemize}

Knowledge Graph Search -> Find compatible combinations, then optionally build

Optional:
\begin{itemize}[label=\textendash, noitemsep, topsep=0pt, leftmargin=*]
  \item \texttt{topology\_iris}: list of topology instance IRIs to restrict
  \item \texttt{metal\_node\_iris}: list of metal-node instance IRIs to restrict
  \item \texttt{organic\_linker\_iris}: list of linker instance IRIs to restrict
\end{itemize}
\\
\hline
\end{tabular}

\vspace{0.5em}

\noindent\textbf{Parameters schema}\par
\vspace{0.25em}

\begin{lstlisting}[
basicstyle=\ttfamily\normalsize,
frame=single,
breaklines=true,
columns=fullflexible,
xleftmargin=0.5em,
xrightmargin=0.5em
]
{"additionalProperties":false,"properties":{"summarised_user_query":{"type":"string"},"topology_pormake_id":{"default":null,"anyOf":[{"type":"string"},{"type":"null"}]},"node_pormake_id":{"default":null,"anyOf":[{"type":"string"},{"type":"null"}]},"linker_pormake_id":{"default":null,"anyOf":[{"type":"string"},{"type":"null"}]},"mof_name":{"default":null,"anyOf":[{"type":"string"},{"type":"null"}]},"csd_refcode":{"default":null,"anyOf":[{"type":"string"},{"type":"null"}]},"input_cif_filename":{"default":null,"anyOf":[{"type":"string"},{"type":"null"}]},"topology_iris":{"default":null,"anyOf":[{"items":{"type":"string"},"type":"array"},{"type":"null"}]},"metal_node_iris":{"default":null,"anyOf":[{"items":{"type":"string"},"type":"array"},{"type":"null"}]},"organic_linker_iris":{"default":null,"anyOf":[{"items":{"type":"string"},"type":"array"},{"type":"null"}]},"update_graph":{"default":false,"type":"boolean"}},"required":["summarised_user_query"],"type":"object"}
\end{lstlisting}

\vspace{0.75em}

\vspace{0.25em}

\noindent
\begin{tabular}{p{0.22\linewidth} p{0.72\linewidth}}
\hline
\textbf{Tool Name} & \textbf{Tool description received by agent/LLMs} \\
\hline
\texttt{duckduckgo\_web\_search} &
<summary>Searches DuckDuckGo for the given query and returns the results.</summary> <returns> <description>The search results.</description> </returns> \\
\hline
\end{tabular}

\vspace{0.5em}

\noindent\textbf{Parameters schema}\par
\vspace{0.25em}

\begin{lstlisting}[
basicstyle=\ttfamily\normalsize,
frame=single,
breaklines=true,
columns=fullflexible,
xleftmargin=0.5em,
xrightmargin=0.5em
]
{"additionalProperties":false,"properties":{"query":{"description":"The query to search for.","type":"string"}},"required":["query"],"type":"object"}
\end{lstlisting}

\vspace{0.75em}

\noindent
\begin{tabular}{p{0.22\linewidth} p{0.72\linewidth}}
\hline
\textbf{Tool Name} & \textbf{Tool description received by agent/LLMs} \\
\hline
\texttt{get\_conversion\_factor} &
<summary>Convert physical quantities from a list of UnitConvPair each containing a numerical value along with source and target units, and the function returns a list of converted values.</summary> <returns> <type>list[float]</type> <description>Converted values in the target unit. str: Error message if conversion is not supported or input is invalid.</description> </returns> \\
\hline
\end{tabular}

\vspace{0.5em}

\noindent\textbf{Parameters schema}\par
\vspace{0.25em}

\begin{lstlisting}[
basicstyle=\ttfamily\normalsize,
frame=single,
breaklines=true,
columns=fullflexible,
xleftmargin=0.5em,
xrightmargin=0.5em
]
{"additionalProperties":false,"properties":{"UnitConversionPairs":{"description":"List of UnitConvPair objects containing value, from_unit, and to_unit.","items":{"$ref":"#/$defs/UnitConvPair"},"type":"array"}},"required":["UnitConversionPairs"],"type":"object","$defs":{"UnitConvPair":{"properties":{"value":{"type":"number"},"from_unit":{"enum":["hartree","eV","kJ/mol","kcal/mol","cm^-1","Hz"],"type":"string"},"to_unit":{"enum":["hartree","eV","kJ/mol","kcal/mol","cm^-1","Hz"],"type":"string"}},"required":["value","from_unit","to_unit"],"type":"object","additionalProperties":false}}}
\end{lstlisting}

\vspace{0.75em}

\noindent
\begin{tabular}{p{0.22\linewidth} p{0.72\linewidth}}
\hline
\textbf{Tool Name} & \textbf{Tool description received by agent/LLMs} \\
\hline
\texttt{capture\_viewer\_screenshot} &
<summary>Capture a screenshot of the current Molecule Viewer and provide it to the model.</summary> <returns> <description>List of str with BinaryContent image included in the content so the LLM can analyze it.</description> </returns> \\
\hline
\end{tabular}

\vspace{0.5em}

\noindent\textbf{Parameters schema}\par
\vspace{0.25em}

\begin{lstlisting}[
basicstyle=\ttfamily\normalsize,
frame=single,
breaklines=true,
columns=fullflexible,
xleftmargin=0.5em,
xrightmargin=0.5em
]
{"additionalProperties":false,"properties":{"molecule_id":{"default":null,"description":"Optional molecule id to target; defaults to \"current\" if omitted.","anyOf":[{"type":"string"},{"type":"null"}]}},"type":"object"}
\end{lstlisting}

\vspace{0.75em}

\noindent
\begin{tabular}{p{0.22\linewidth} p{0.72\linewidth}}
\hline
\textbf{Tool Name} & \textbf{Tool description received by agent/LLMs} \\
\hline
\texttt{run\_crest\_conformer\_search} &
<summary>Run a CREST conformer search for a given molecule identifier.</summary> <returns> <description>A JSON-serializable dictionary describing the CREST ensemble.</description> </returns> \\
\hline
\end{tabular}

\vspace{0.5em}

\noindent\textbf{Parameters schema}\par
\vspace{0.25em}

\begin{lstlisting}[
basicstyle=\ttfamily\normalsize,
frame=single,
breaklines=true,
columns=fullflexible,
xleftmargin=0.5em,
xrightmargin=0.5em
]
{"additionalProperties":false,"properties":{"identifier":{"description":"The molecule identifier (SMILES string, name, or XYZ file path).","type":"string"},"identifier_type":{"description":"The type of the identifier ('smiles', 'name', or 'xyz').","enum":["smiles","name","xyz"],"type":"string"},"charge":{"default":0,"description":"The molecular charge (default: 0).","type":"number"},"spin_multiplicity":{"default":1,"description":"The spin multiplicity (default: 1).","type":"integer"},"implicit_solvent":{"default":null,"description":"The implicit solvent model to use (default: None), example: \"water\", \"acetonitrile\".","anyOf":[{"type":"string"},{"type":"null"}]},"calculation_level_method":{"default":"gfn2","description":"The method to use for the CREST calculation (default: 'gfn2').","enum":["tblite","gfn2","gfn1","gfnff","gfn0","gfn0*","xtb","gfn","gfn-xtb","orca","generic"],"type":"string"},"crest_runtype":{"default":"imtd-gc","description":"CREST sampling mode ('imtd-gc' for standard sampling, 'nci' for non-covalent complexes and aggregates sampling which should be used for explicitly solvated structures).","enum":["imtd-gc","nci"],"type":"string"},"run_on_gpu":{"default":false,"description":"Whether to run the calculation on GPU (default: False).","type":"boolean"}},"required":["identifier","identifier_type"],"type":"object"}
\end{lstlisting}

\vspace{0.75em}

\noindent
\begin{tabular}{p{0.22\linewidth} p{0.72\linewidth}}
\hline
\textbf{Tool Name} & \textbf{Tool description received by agent/LLMs} \\
\hline
\texttt{run\_qcg\_cluster} &
Run a QCG grow calculation and return the merged cluster as XYZ. \\
\hline
\end{tabular}

\vspace{0.5em}

\noindent\textbf{Parameters schema}\par
\vspace{0.25em}

\begin{lstlisting}[
basicstyle=\ttfamily\normalsize,
frame=single,
breaklines=true,
columns=fullflexible,
xleftmargin=0.5em,
xrightmargin=0.5em
]
{"additionalProperties":false,"properties":{"solute_identifier":{"description":"Solute identifier (SMILES, name, or XYZ string).","type":"string"},"solvent_identifier":{"description":"Solvent identifier (SMILES, name, or XYZ string).","type":"string"},"solute_identifier_type":{"description":"Type of solute identifier.","enum":["smiles","name","xyz"],"type":"string"},"solvent_identifier_type":{"description":"Type of solvent identifier.","enum":["smiles","name","xyz"],"type":"string"},"charge":{"default":0,"description":"Total molecular charge.","type":"number"},"spin_multiplicity":{"default":1,"description":"Spin multiplicity (e.g., 1 for singlet, 2 for doublet).","minimum":1,"type":"integer"},"nsolv":{"default":15,"description":"Number of solvent molecules to add.","exclusiveMinimum":0,"type":"integer"},"threads":{"default":12,"description":"Threads passed to CREST via --T.","exclusiveMinimum":0,"type":"integer"},"alpb_solvent":{"default":"water","description":"ALPB solvent name (case-insensitive). Available: acetone, acetonitrile, aniline, benzaldehyde, benzene, ch2cl2, chcl3, cs2, dioxane, dmf, dmso, ethanol, ether, ethylacetate, furane, hexadecane, hexane, methanol, nitromethane, octanol, octanol (wet), phenol, toluene, thf, water","anyOf":[{"type":"string"},{"type":"null"}]}},"required":["solute_identifier","solvent_identifier","solute_identifier_type","solvent_identifier_type"],"type":"object"}
\end{lstlisting}

\vspace{0.75em}

\noindent
\begin{tabular}{p{0.22\linewidth} p{0.72\linewidth}}
\hline
\textbf{Tool Name} & \textbf{Tool description received by agent/LLMs} \\
\hline
\texttt{show\_molecule\_in\_viewer} &
<summary>Instruct the Molecule Viewer to display a ConceptualAtoms by its IRI.</summary> <returns> <description>A confirmation message indicating the viewer has been updated.</description> </returns> \\
\hline
\end{tabular}

\vspace{0.5em}

\noindent\textbf{Parameters schema}\par
\vspace{0.25em}

\begin{lstlisting}[
basicstyle=\ttfamily\normalsize,
frame=single,
breaklines=true,
columns=fullflexible,
xleftmargin=0.5em,
xrightmargin=0.5em
]
{"additionalProperties":false,"properties":{"conceptual_atoms_iri":{"description":"The IRI of the ConceptualAtoms instance to display.","type":"string"}},"required":["conceptual_atoms_iri"],"type":"object"}
\end{lstlisting}

\vspace{0.75em}

\noindent
\begin{tabular}{p{0.22\linewidth} p{0.72\linewidth}}
\hline
\textbf{Tool Name} & \textbf{Tool description received by agent/LLMs} \\
\hline
\texttt{run\_pyscf\_workflow} &
<summary>Run PySCF calculations using the workflow graph with dynamic routing capabilities.</summary> <returns> <type>list[PyscfOutput]</type> <description>PySCF calculation outputs gathered from the workflow run.</description> </returns> \\
\hline
\end{tabular}

\vspace{0.5em}

\noindent\textbf{Parameters schema}\par
\vspace{0.25em}

\begin{lstlisting}[
basicstyle=\ttfamily\normalsize,
frame=single,
breaklines=true,
columns=fullflexible,
xleftmargin=0.5em,
xrightmargin=0.5em
]
{"additionalProperties":false,"properties":{"summerised_user_query":{"description":"Description of the calculations the user wants to run, this will be used by the workflow routing agent to determine which next nodes to execute.","type":"string"},"identifier_type":{"description":"Kind of identifier supplied to describe the molecule.","enum":["name","smiles","xyz","xyz_filename","conceptual_atoms_iri"],"type":"string"},"identifier":{"description":"Identifier content (name, SMILES, or XYZ string) describing the conceptual atom; for XYZ, provide the literal file text such as ``3\\nCOMMENT\\nO 0.000 0.000 0.000\\nH 0.000 0.757 0.586\\nH 0.000 -0.757 0.586``.","type":"string"},"charge":{"default":0,"description":"Overall molecular charge; defaults to 0.","type":"integer"},"spin_multiplicity":{"default":1,"description":"Spin multiplicity of the molecule; defaults to 1.","type":"integer"},"basis_set":{"default":null,"description":"Basis set name for the calculation; falls back to workflow defaults when omitted.","anyOf":[{"type":"string"},{"type":"null"}]},"restricted":{"default":true,"description":"Whether to use a restricted reference (RHF/ROHF/RDFT) instead of unrestricted; defaults to True.","type":"boolean"},"xc_functional":{"default":null,"description":"Exchange-correlation functional to activate DFT; when omitted, the workflow uses Hartree-Fock.","anyOf":[{"type":"string"},{"type":"null"}]},"solvation_model":{"default":null,"description":"Implicit solvation model label; uses vacuum when not provided.","anyOf":[{"enum":["CPCM","COSMO","IEF-PCM","SS(V)PE","SMD"],"type":"string"},{"type":"null"}]},"implicit_solvent":{"default":null,"description":"Solvent environment name for the chosen solvation model; defaults to water when paired with an implicit model.","anyOf":[{"enum":["Water","Dimethylsulfoxide","Nitromethane","Acetonitrile","Methanol","Ethanol","Acetone","Methylenechloride","Tetrahydrofurane","Aniline","Chlorobenzene","Chloroform","Toluene","Benzene","Cyclohexane","N-heptane"],"type":"string"},{"type":"null"}]},"exit_node":{"default":null,"description":"Identifier of a workflow node to terminate at, enabling partial runs.","anyOf":[{"type":"string"},{"type":"null"}]}},"required":["summerised_user_query","identifier_type","identifier"],"type":"object"}
\end{lstlisting}

\vspace{0.75em}

\noindent
\begin{tabular}{p{0.22\linewidth} p{0.72\linewidth}}
\hline
\textbf{Tool Name} & \textbf{Tool description received by agent/LLMs} \\
\hline
\texttt{run\_sparql\_query} &
<summary>Execute a read-only SPARQL query against the knowledge graph.
Supports SELECT, CONSTRUCT, ASK, and DESCRIBE queries.
UPDATE operations are rejected. Queries are subject to:

45 second timeout
Complexity limits (max 50 triple patterns, OPTIONAL depth 5, UNION branches 10)
Injection pattern detection and basic syntax validation</summary> <returns>
<description>Query results as a string. Results are truncated if too large.</description>
</returns> \\
\hline
\end{tabular}

\vspace{0.5em}

\noindent\textbf{Parameters schema}\par
\vspace{0.25em}

\begin{lstlisting}[
basicstyle=\ttfamily\normalsize,
frame=single,
breaklines=true,
columns=fullflexible,
xleftmargin=0.5em,
xrightmargin=0.5em
]
{"additionalProperties":false,"properties":{"sparql":{"description":"The SPARQL query string. Must be a valid read-only query.","type":"string"}},"required":["sparql"],"type":"object"}
\end{lstlisting}

\vspace{0.75em}

\noindent
\begin{tabular}{p{0.22\linewidth} p{0.72\linewidth}}
\hline
\textbf{Tool Name} & \textbf{Tool description received by agent/LLMs} \\
\hline
\texttt{get\_ontology\_snapshot} &
<summary>Retrieve complete schema snapshot for an ontology.
Returns all classes with their data properties and object properties
in a single call, giving context for reasoning about
the knowledge graph structure.</summary>
<returns>
<description>{
"ontology\_iri": str,
"ontology\_description": str,
"class\_count": int,
"classes": {
class\_iri: {
"py\_class": "module.ClassName",
"description": "Class docstring summary",
"data\_properties": {
property\_iri: {
"field": "field\_name",
"type": "python.type.Path",
"description": "Field description"
}
},
"object\_properties": { ... }
}
}
}</description>
</returns> \\
\hline
\end{tabular}

\vspace{0.5em}

\noindent\textbf{Parameters schema}\par
\vspace{0.25em}

\begin{lstlisting}[
basicstyle=\ttfamily\normalsize,
frame=single,
breaklines=true,
columns=fullflexible,
xleftmargin=0.5em,
xrightmargin=0.5em
]
{"additionalProperties":false,"properties":{"ontology_iri":{"description":"The IRI of the ontology to inspect.\nExample: \"https://elagente.ca/ontomof\"","type":"string"},"class_iris":{"default":null,"description":"Optional filter - only return these specific classes.\nIf None, returns all classes in the ontology.","anyOf":[{"items":{"type":"string"},"type":"array"},{"type":"null"}]}},"required":["ontology_iri"],"type":"object"}
\end{lstlisting}

\vspace{0.75em}

\noindent
\begin{tabular}{p{0.22\linewidth} p{0.72\linewidth}}
\hline
\textbf{Tool Name} & \textbf{Tool description received by agent/LLMs} \\
\hline
\texttt{get\_instance\_from\_knowledge\_graph} &
<summary>Retrieve a knowledge graph object by its IRI.
Large data fields (NDArrays, CIF text, etc.) are stripped and replaced
with metadata. Use get\_cif\_content to retrieve full CIF text.</summary>
<returns>
<description>{
"instance\_iri": str,
"class\_type": "ClassName",
"data": { ... fields with large data stripped ... },
"large\_fields\_available": ["cif\_text", "positions", ...]
}</description>
</returns> \\
\hline
\end{tabular}

\vspace{0.5em}

\noindent\textbf{Parameters schema}\par
\vspace{0.25em}

\begin{lstlisting}[
basicstyle=\ttfamily\normalsize,
frame=single,
breaklines=true,
columns=fullflexible,
xleftmargin=0.5em,
xrightmargin=0.5em
]
{"additionalProperties":false,"properties":{"instance_iri":{"description":"The IRI of the object to retrieve.\nExample: \"https://elagente.ca/ontomof/ConstructedMOF_abc123\"","type":"string"},"recursive_depth":{"default":3,"description":"How deep to recursively pull related objects.\nDefault is 3 (direct properties and some related context). Increase for more context. For pulling ConceptualAtoms or related structures, use -1.","type":"integer"}},"required":["instance_iri"],"type":"object"}
\end{lstlisting}

\vspace{0.75em}

\noindent
\begin{tabular}{p{0.22\linewidth} p{0.72\linewidth}}
\hline
\textbf{Tool Name} & \textbf{Tool description received by agent/LLMs} \\
\hline
\texttt{get\_cif\_content} &
<summary>Retrieve the CIF content for a ConstructedMOF.
CIF (Crystallographic Information File) contains the full atomic structure
for visualization or analysis.</summary>
<returns>
<description>{
"mof\_iri": str,
"mof\_name": str or None,
"cif\_text": "full CIF content...",
"n\_atoms": int
}</description>
</returns> \\
\hline
\end{tabular}

\vspace{0.5em}

\noindent\textbf{Parameters schema}\par
\vspace{0.25em}

\begin{lstlisting}[
basicstyle=\ttfamily\normalsize,
frame=single,
breaklines=true,
columns=fullflexible,
xleftmargin=0.5em,
xrightmargin=0.5em
]
{"additionalProperties":false,"properties":{"mof_iri":{"description":"The IRI of the ConstructedMOF instance.","type":"string"}},"required":["mof_iri"],"type":"object"}

\end{lstlisting}

\vspace{0.75em}

\noindent
\begin{tabular}{p{0.22\linewidth} p{0.72\linewidth}}
\hline
\textbf{Tool Name} & \textbf{Tool description received by agent/LLMs} \\
\hline
\texttt{run\_python\_code} (exposed via MCP server) &
Tool to execute Python code and return stdout, stderr, and return value.

Guidelines

The code may be async, and the value on the last line will be returned as the return value.

The code will be executed with Python 3.13 using pyodide - so adapt your code if needed.

You code must be executed within a timeout. You have 60 seconds before the run is canceled.

You have these python packages installed: numpy,pandas,matplotlib

To output files or images, save them in the "/output\_files" folder.\\
\hline
\end{tabular}

\vspace{0.5em}

\noindent\textbf{Parameters schema}\par
\vspace{0.25em}

\begin{lstlisting}[
basicstyle=\ttfamily\normalsize,
frame=single,
breaklines=true,
columns=fullflexible,
xleftmargin=0.5em,
xrightmargin=0.5em
]
{"type": "object", "properties": {"python_code": {"type": "string", "description": "Python code to run"}, "global_variables": {"type": "object", "additionalProperties": {}, "default": {}, "description": "Map of global variables in context when the code is executed", "properties": {}}}, "required": ["python_code"], "additionalProperties": false}
\end{lstlisting}

\vspace{0.75em}

%% file: includes/si_tex/si-execution-graph.tex
This section describes the reusable abstraction layer (Python module \texttt{grafico.abstraction}) from which all domain-specific execution graphs in this work are constructed, including the GPU4PySCF workflow discussed in the main text and the \ac{mof} workflow detailed in Supplementary Sec.~\ref{si:mof_exploration}. The layer is built on top of \texttt{pydantic-graph} (part of the \texttt{pydantic-ai} family; see the infrastructure description in the main text) and comprises three components: (i) a shared workflow state that serves as the working memory and audit trail of a graph run, (ii) typed routing configurations that couple each routing decision to validated inputs for the selected node, and (iii) node base classes that derive the set of admissible transitions from Python type hints and constrain the routing \ac{llm} to that set via structured outputs. Domain workflows are implemented by subclassing these components; the abstraction layer itself contains no chemistry-specific logic.

\subsection*{Workflow state and working memory}
\label{si:subsec_grafico_state}

Every execution graph operates over a subclass of \texttt{GraficoState}, a Python dataclass holding the fields shared by all workflows:

\begin{itemize}
    \item \textbf{\texttt{summarised\_user\_query}:} a concise restatement of the user's intent, supplied by the main agent when it invokes the graph tool. This summary, rather than the full conversation history, is what the routing controller conditions on.
    \item \textbf{\texttt{workflow\_summary}:} an append-only list of one-line step summaries (\textit{e.g.}, \texttt{"Single point energy: -230.198654 Hartree"}). This list acts as the working memory of the run: nodes append a summary after each completed step, and routing prompts are assembled from these lines. Each append is simultaneously emitted as a structured log record (via \texttt{logfire}), so the working memory doubles as a live execution trace.
    \item \textbf{\texttt{node\_config\_history}:} an audit trail of routing decisions. For every transition, the state records the source node, the selected successor, the router's natural-language reasoning, its self-reported confidence, and the validated configuration payload passed to the successor.
\end{itemize}

Heavyweight scientific objects (\textit{e.g.}, PySCF mean-field objects, \texttt{ConceptualAtoms} instances, conformer ensembles) are held in fields added by the domain-specific state subclasses and are passed between nodes as direct Python object references. Only the one-line textual summaries enter \ac{llm} prompts; the scientific state itself never transits the model context.

\subsection*{Typed routing configurations}

Routing decisions are represented as Pydantic models derived from \texttt{GraficoNodeBaseConfig}, which declares three base fields: \texttt{node\_name} (the identifier of the chosen successor), \texttt{reasoning} (a 1--3 sentence justification), and \texttt{confidence} (constrained to $[0, 1]$). Each node class in a workflow declares a corresponding config subclass that (i) narrows \texttt{node\_name} to an exact \texttt{Literal} matching the node's class name, and (ii) adds the node's input parameters as validated fields with defaults, bounds, and natural-language descriptions (\textit{e.g.}, \texttt{GeometryOptimisationConfig} adds \texttt{max\_steps} with \texttt{ge=1}; \texttt{TDDFTConfig} adds \texttt{nstates}). A synthetic \texttt{EndConfig} (\texttt{node\_name = Literal["End"]}) represents graph termination. A routing decision is therefore not merely a choice of successor: the same validated object carries the arguments with which the successor is instantiated, so schema violations (out-of-range values, missing required fields, unknown node names) are rejected by Pydantic validation before any computation runs.

\subsection*{Deriving admissible transitions from type hints}

The node base class \texttt{GraficoNodeBase} extends \texttt{pydantic-graph}'s \texttt{BaseNode} and derives the set of admissible successors of a node directly from the return type annotation of its \texttt{run()} method. For example, the frequency-calculation node in the GPU4PySCF workflow declares

\begin{center}
\texttt{async def run(...) -> TDDFT | MolecularAnalysis | ImaginaryFrequencyRemoval | PySCFGraphEnd}
\end{center}

\noindent and these four classes become the only transitions the routing controller can select from this node. The annotation is resolved with \texttt{typing.get\_type\_hints} and recursively flattened across \texttt{Annotated}, \texttt{Optional}, and union wrappers. The declared graph topology is validated eagerly at graph-definition time, not at runtime: a plain \texttt{BaseNode} in the annotation (which would leave the successor set unspecified), a successor that does not subclass \texttt{GraficoNodeBase}, a duplicate successor label, or a config model whose \texttt{node\_name} literal does not match its target all raise a \texttt{GraficoNodeDefinitionError} before the graph can be executed. The resolved successor sets and routing schemas are cached per node class.

From the successor set, the layer constructs the routing output schema. The config model of each candidate is introspected from the candidate's \texttt{config} field annotation (each node must declare exactly one config model, optionally unioned with \texttt{None}); terminal targets map to \texttt{EndConfig}. When multiple successors are admissible, their config models are combined into a discriminated union over the \texttt{node\_name} literal. This union is supplied as the structured output type of the routing \ac{llm} call, so the model is constrained by the provider-side response schema, and subsequently by Pydantic validation, to select exactly one declared successor and to populate its typed arguments. Listing~\ref{lst:si-node-definition} shows the resulting definition pattern for one node of the GPU4PySCF workflow.

\begin{lstlisting}[
language=Python,
basicstyle=\ttfamily\small,
frame=single,
breaklines=true,
columns=fullflexible,
xleftmargin=0.5em,
xrightmargin=0.5em,
caption={Definition pattern for a deterministic node in the GPU4PySCF execution graph (simplified). The config subclass defines the typed inputs the router may instantiate; the return annotation of \texttt{run()} defines the admissible transitions.},
label={lst:si-node-definition}
]
class GeometryOptimisationConfig(GraficoNodeBaseConfig):
    node_name: Literal["GeometryOptimisation"] = "GeometryOptimisation"
    max_steps: int = Field(
        default=100, ge=1,
        description="Maximum steps for geometry optimization.",
    )

@dataclass
class GeometryOptimisation(PySCFWorkflowNode):
    """Optimize molecular geometry, should be followed by
    frequency calculation to check stability"""

    config: Optional[GeometryOptimisationConfig] = None

    async def run(
        self, ctx: GraphRunContext[PySCFWorkflowState, GraficoDeps]
    ) -> FrequencyCalculation | TDDFT | MolecularAnalysis | PySCFGraphEnd:
        ...  # deterministic GPU4PySCF geometry optimization
        return await self.route_structured(ctx=ctx)
\end{lstlisting}

\subsection*{Schema-conditioned routing and fallback semantics}

The complete routing procedure, including the fallback semantics described below and the single-call variant used by \ac{llm} nodes (Supplementary Sec.~\ref{si:subsec_llm_nodes}), is summarized in Algorithm~\ref{alg:si-routing}.

\begin{algorithm}[H]
\caption{Schema-conditioned routing between typed execution graph nodes. \textsc{RouteStructured} is invoked by deterministic nodes after their computation completes; \textsc{RunLLMNode} is the single-call variant for \ac{llm} nodes, in which the task output and the routing decision are returned by one envelope. Both share the normalization procedure \textsc{Materialize}.}
\label{alg:si-routing}
\begin{algorithmic}[1]
\Procedure{RouteStructured}{node $v$, state $s$}
    \State $T \gets$ admissible successors of $v$, resolved from the return type hints of $v$.\texttt{run()}
    \If{$T = \emptyset$} \Return \textsc{End} \EndIf
    \State $\Sigma \gets$ discriminated union of the config models of all $t \in T$ \Comment{cached per node class}
    \State $p \gets$ prompt$($current node, summarised user query, working memory, result of $v$, state snapshot, docstring description of each $t \in T)$
    \State $c \gets$ \Call{RouterLLM}{$p$, \,output schema $\Sigma$} \Comment{structured output}
    \If{the routing call raises} $c \gets \bot$ with the recorded failure reason \EndIf
    \State \Return \Call{Materialize}{$v$, $s$, $T$, $c$}
\EndProcedure
\Statex
\Procedure{RunLLMNode}{node $v$, state $s$}
    \State $T \gets$ successors declared in $v$.\texttt{next\_nodes}; \ $\Sigma \gets$ discriminated union as above
    \State $(r, c) \gets$ \Call{NodeLLM}{task prompt, \,output schema $\langle\texttt{result},\, \texttt{next\_node}{:}\,\Sigma\rangle$}
    \State append the task output $r$ to the working memory of $s$
    \If{$T = \emptyset$} \Return \textsc{End}$(r)$ \EndIf
    \State \Return \Call{Materialize}{$v$, $s$, $T$, $c$}
\EndProcedure
\Statex
\Procedure{Materialize}{$v$, $s$, $T$, $c$}
    \If{hard-exit field of $s$ names $v$} \Return \textsc{End} \Comment{deterministic bound set by the main agent} \EndIf
    \If{$c$ is a valid config \textbf{and} $c$.\texttt{node\_name} $\in T$}
        \State $d \gets c$
    \ElsIf{$|T| > 1$} \Comment{never guess among multiple branches}
        \State $d \gets \texttt{EndConfig}(\text{failure reason},\ \text{confidence} = 0.0)$
    \Else \Comment{a single admissible successor may be defaulted}
        \State $d \gets$ default config of the sole $t \in T$ with the fallback reason, confidence $= 0.1$
    \EndIf
    \State record $d$ in the working memory, \texttt{node\_config\_history}, and structured logs
    \If{$d$ selects \textsc{End}} \Return \textsc{End} \EndIf
    \State \Return instance of $T[d.\texttt{node\_name}]$ constructed with config $d$ \Comment{typed inputs travel with the node}
\EndProcedure
\end{algorithmic}
\end{algorithm}

After a deterministic node completes its computation, it delegates the choice of the next node to \texttt{route\_structured()}, which assembles a routing prompt containing: the name of the current node, the summarized user request, the working-memory summary (all prior \texttt{workflow\_summary} entries), the result of the current node, an optional domain-specific state snapshot (a per-workflow hook, \textit{e.g.}, listing which calculations have already succeeded), and a description of each admissible successor generated automatically from the successor's class docstring. The prompt is sent to a dedicated router agent whose model and system prompt are class attributes overridable per workflow; in this work the router is a lightweight model (\texttt{gpt-4.1} by default) so that routing adds only a small cost per transition, and the GPU4PySCF workflow overrides the system prompt with domain-specific node-selection rules. The router returns an instance of the discriminated union described above.

Every accepted decision is recorded three ways: as a structured log record, as a working-memory entry of the form \texttt{[AI Routing]: <from> -> <to> | <reasoning>}, and as an entry in \texttt{node\_config\_history}. The selected node is then materialized by passing the validated config object to the successor's constructor, so the router-instantiated arguments travel with the node instance.

Failures are handled conservatively:
\begin{itemize}
    \item \textbf{Hard exit:} a workflow may name a state field (\texttt{structured\_hard\_exit\_field}) that, when set to the current node's name, forces immediate termination regardless of the router output. This gives the main agent a deterministic mechanism to bound graph execution (\textit{e.g.}, ``stop after geometry optimization'').
    \item \textbf{Single admissible successor:} if the routing call fails or returns an unavailable target and only one successor is declared, the workflow falls back to that successor, recording the fallback reason and a nominal confidence of 0.1.
    \item \textbf{Multiple admissible successors:} automatic fallback is disabled; guessing among several branches could silently execute the wrong computation. Instead, the workflow terminates with an \texttt{EndConfig} carrying the failure reason and zero confidence, returning control (and the recorded error) to the main agent.
\end{itemize}

\subsection*{\ac{llm} nodes with combined task--routing outputs}
\label{si:subsec_llm_nodes}

Some nodes perform semantic rather than numerical work (\textit{e.g.}, extracting structured data from text). For these, the subclass \texttt{LLMGraficoNodeBase} embeds a \texttt{pydantic-ai} agent inside the node; the node's instructions, model, structured output type, and tool list are declared as class attributes. Because such a node produces its result and its routing decision from the same model, the two are merged into a single \emph{envelope} output: a Pydantic model generated at class-definition time with two fields, \texttt{result} (the node's declared task output type) and \texttt{next\_node} (the routing union of the declared successors). One \ac{llm} \ac{api} call thus yields both the validated task output and the validated transition, keeping the two consistent and avoiding a second routing call per node. The node's user prompt is composed from the summarized user request, the previous node's summary, the routed inputs it received (its config, excluding the routing metadata fields), and the routing context (working memory and successor descriptions).

Two implementation details differ from deterministic nodes. First, successors are declared through an explicit \texttt{next\_nodes} class attribute (rather than the \texttt{run()} return annotation, which for these nodes is generic), and \texttt{get\_node\_def()} is overridden so that \texttt{pydantic-graph} registers the correct edges for graph introspection and diagram generation. Second, the embedded agent is executed in streaming mode, and a per-event hook forwards intermediate output to the GraphChat frontend (Supplementary Sec.~\ref{si:graphchat}) for real-time display. A node that declares no successors returns its task output directly as the graph result.

\subsection*{Terminal nodes and \ac{kg} persistence}

Graph termination is implemented by \texttt{GraficoGraphEndBase}, a template that fixes the exit sequence of every workflow: build a typed result object from the final state (\texttt{build\_result()}, implemented per workflow), optionally push the result and its constituent \ac{ogm} objects to the \ac{kg}, and return the result to the calling agent as the graph tool's output. \ac{kg} persistence is gated on a state flag (\texttt{update\_graph}), so provenance is written when the user or agent opts in, and the push step requires the shared dependency object providing the \ac{kg} endpoint. Pre- and post-hooks around both result construction and persistence allow workflows to attach domain-specific bookkeeping (\textit{e.g.}, frontend notifications) without modifying the template. In the GPU4PySCF workflow, the terminal node assembles a result model carrying the computed energies, geometries, and the \acp{iri} of the persisted \texttt{ConceptualAtoms} instances, which the main agent then uses to dispatch downstream graphs or to answer the user.

%% file: includes/si_tex/si-graphchat.tex
GraphChat is the web interface through which users interact with \grafico{}. Beyond a conventional chat window, it is designed around a single principle that follows from the runtime/harness philosophy of the main text: the scientific state produced by the agent should be \emph{co-observable} by the user and the agent in real time, rather than reconstructed from text after the fact. This section documents the interface features and the synchronization architecture that makes this co-observability possible.

\paragraph{\textbf{Shared-document architecture}}
The frontend (a React application) and the \grafico{} Python backend do not communicate through a bespoke request/response protocol. Instead, both parties are peers on a single shared document held as a \ac{crdt}, a data structure in which concurrent edits from multiple clients converge without conflicts. Each conversation (a ``room'') corresponds to one document synchronized over WebSocket: the browser uses \texttt{Yjs}, while the \grafico{} backend uses \texttt{pycrdt} through the GraphChat SDK described in the main text, and both connect to the same room on the Node.js server. The document exposes a small set of named shared collections that constitute the entire interface contract: an append-only \texttt{messages} array, a \texttt{config} map (selected model, session identifier, backend readiness and capabilities, token-usage totals), a live molecule map (\texttt{xyz}), an execution-graph array (\texttt{graph}), atom-selection maps, and a camera-state map. The agent writes incremental output (assistant text, reasoning, logs, tool events, molecular frames) directly into these structures, and every connected client re-renders reactively. Because the same room document is served to any device that joins it, all interface features described below are, by construction, live and identical across devices and sessions, \textit{e.g.}, phone and laptop.

\begin{figure}[!ht]
    \centering
    \includegraphics[width=\linewidth]{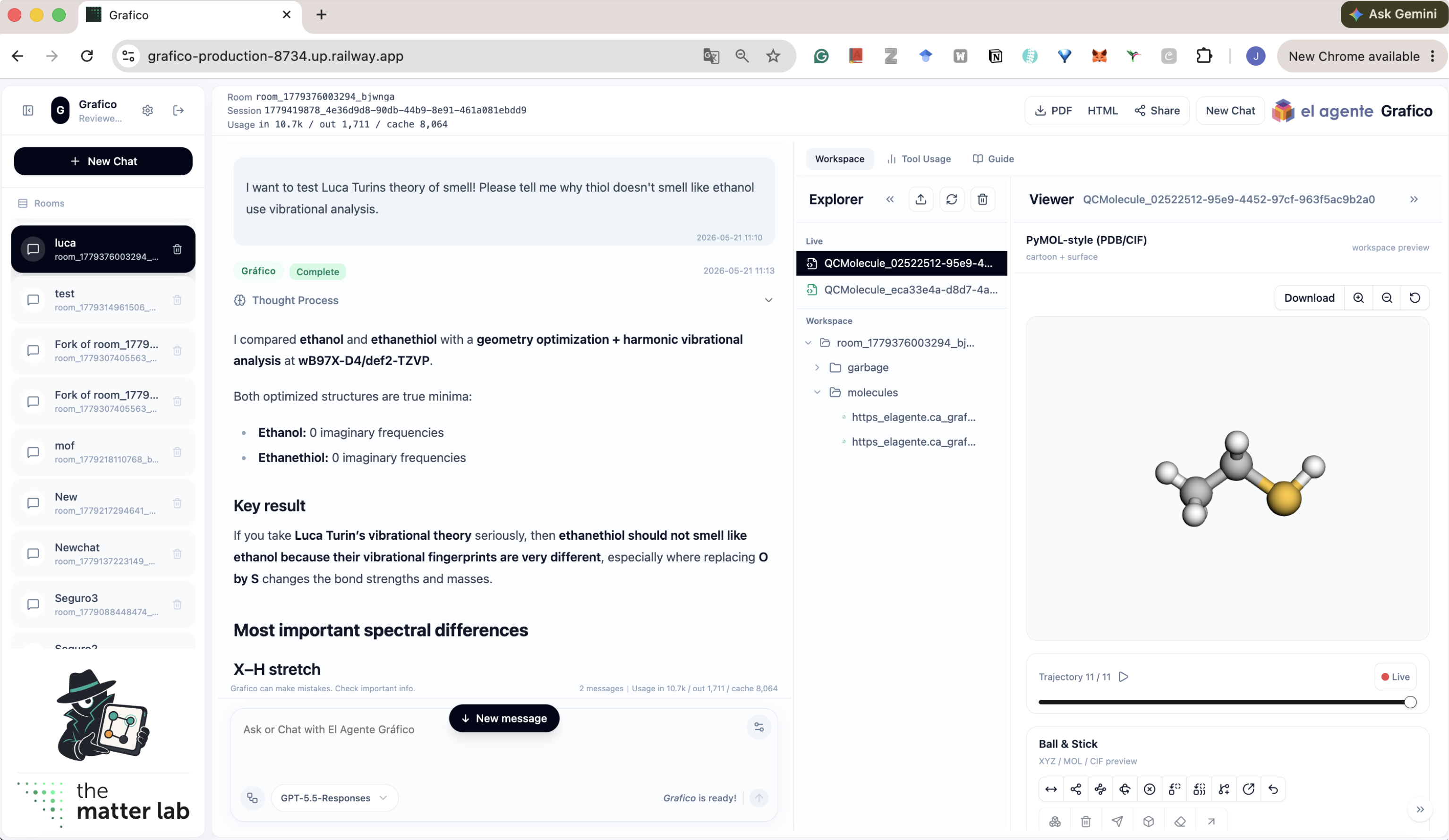}
    \caption{The default GraphChat interface. The chat panel (left) streams the agent's response, reasoning, and tool-call cards; the workspace panel (right) hosts the interactive molecule viewer and the room file explorer. A live token-usage readout and a model selector are shown in the header.}
    \label{fig:si-graphchat-interface}
\end{figure}

\paragraph{\textbf{Chat interface}}
Figure~\ref{fig:si-graphchat-interface} shows the default interface. Assistant responses stream token-by-token as the backend appends characters to the shared message text. Three response channels are rendered separately: the final answer (Markdown with syntax-highlighted code blocks and \LaTeX{} math rendered through KaTeX), a collapsible \emph{reasoning} panel that streams the model's thinking, and a collapsible \emph{logs} panel. Tool calls are shown as dedicated cards that display the invoked tool, its validated arguments, live \texttt{stdout}/\texttt{stderr}, and a status badge (running, completed, or failed); streamed tool output is merged into the matching card by call identifier. Per-message and room-level token usage (input, output, and cached tokens, normalized across provider conventions) are displayed continuously, and the model driving the agent can be switched from a menu at any point without interrupting the session. When a tool call is configured to require human approval, the interface surfaces an approval panel through which the user authorizes, denies, or supplies external results before the paused run resumes.

\begin{figure}[!ht]
    \centering
    \includegraphics[width=0.3\linewidth]{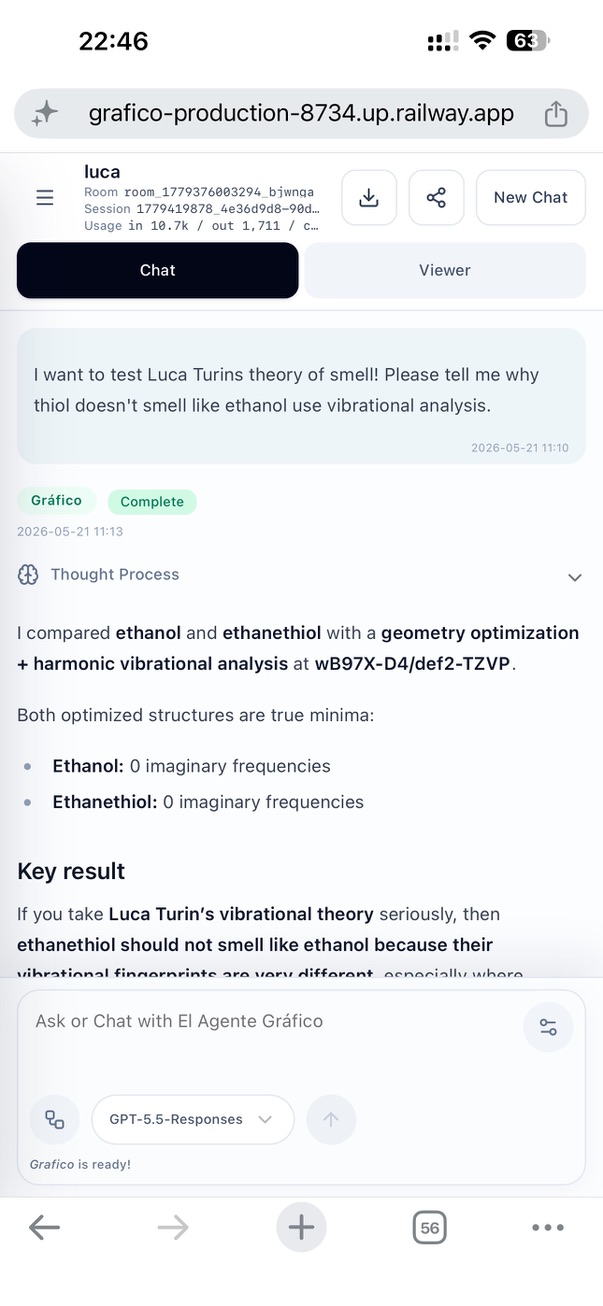}
    \caption{The default GraphChat interface accessed via phone.}
    \label{fig:si-graphchat-interface-phone}
\end{figure}

\paragraph{\textbf{Interactive molecule viewer}}
Molecular and periodic structures are rendered in an interactive WebGL viewer (built on 3Dmol.js) that opens whenever the agent produces a structure. Each structure is keyed by the \ac{iri} of its underlying \texttt{ConceptualAtoms} instance, so the object the user inspects is the same typed object that flows through the execution graph. The viewer supports both molecular systems and periodic systems, for which the crystallographic unit cell can be toggled and drawn with labeled lattice vectors. Users can rotate, zoom, and reset the camera; the camera state is itself written to the shared document, so the view can be observed by the agent and mirrored across clients. Two interaction modes feed information back to the agent. First, atoms can be selected by clicking (single molecule) or accumulated across several molecules (staged selection); the selected atom indices are written to the shared document and delivered to the agent as a structured message. Second, the viewer can capture a PNG snapshot and push it into the conversation as an image the \ac{llm} can see; the agent can also \emph{request} a snapshot programmatically, giving a two-way visual channel between model and rendering. 

\paragraph{\textbf{Trajectory streaming and download}}
When a calculation produces a sequence of geometries (\textit{e.g.}, a geometry-optimization or molecular-dynamics trajectory), each new frame is appended to the shared molecule entry and the viewer advances to it automatically, so the user watches the structure evolve in real time as the computation proceeds. A playback slider with play/pause controls allows scrubbing through the completed trajectory. Any single frame or the full multi-frame trajectory can be copied as text or downloaded in standard formats, including (extended) \texttt{xyz}, \texttt{mol}, and \texttt{CIF}.

\paragraph{\textbf{Workspace and file explorer}}
Each room is backed by a server-side workspace directory that the agent's tools execute within, exposed to the user as a file explorer alongside the viewer. Users can upload files (\textit{e.g.}, \texttt{xyz}/\texttt{CIF} structures or Python scripts for the agent to use), including \texttt{zip} archives that are extracted server-side, and the agent's own output artifacts appear in the same tree, refreshing automatically when the backend signals a change. The explorer previews content by type: molecular files open in the 3Dmol viewer, images and PDFs render inline, and source files (Python, JSON, YAML, and others) are shown with syntax highlighting. Individual files or entire folders (as \texttt{zip}) can be downloaded. Path handling is constrained to the room directory to prevent traversal outside the workspace.

\paragraph{\textbf{Multi-device access with shared context}}
Because the interface state lives entirely in the synchronized room document, GraphChat is fully responsive and usable from a mobile device, which joins the same room and therefore shares identical, live context with a desktop session. On small screens the layout collapses to a chat/viewer tab switcher with a drawer-based room list; virtual-keyboard and safe-area insets are handled explicitly. A user can, for example, launch a computation from a workstation and monitor its streaming trajectory from a phone, or vice versa. Figures~\ref{fig:si-graphchat-interface} and \ref{fig:si-graphchat-interface-phone} demonstrate the same session accessed via desktop and phone, respectively.

\paragraph{\textbf{Sharing, synchronization, and forking}}
A room owner can mint share links in three modes. A \emph{read} link grants a live, read-only view, enforced at the synchronization layer by discarding any edits from the viewer; a \emph{write} link lets an invitee join and collaborate in the same synchronized conversation; and a \emph{fork} link, when opened by an authenticated user, deep-copies the room's current state into a new independent room with a fresh session identifier and its own workspace, allowing a conversation to be branched and continued along a different path without affecting the original. Share tokens are opaque, server-generated identifiers.

\paragraph{\textbf{Concurrent conversations}}
Users can maintain multiple rooms and switch between them from a sidebar. Each room is an independent synchronized document with its own agent session, workspace, and persisted history, so several conversations (and their underlying agent runs) proceed concurrently and independently.

\paragraph{\textbf{Transcript export}}
A complete conversation can be exported to a self-contained HTML document or a paginated PDF, rendered directly from the room's message history. The export preserves Markdown formatting, reasoning and log panels, and tool-call/result blocks, while omitting hidden control messages; PDF export includes fallbacks to CJK-capable fonts.

\paragraph{\textbf{Backend, persistence, and deployment}}
The server is a single Node.js process (Express) that serves the REST endpoints (authentication, workspace files, transcript export, and sharing) and the WebSocket synchronization endpoint, and, in production, the compiled frontend. Every room's full document state is persisted to MongoDB and rehydrated on reconnection, so conversations survive server restarts; the agent's own message history is persisted separately per session to allow the reasoning context to resume. The agent backend connects as a privileged service peer via a service token, while browser clients authenticate against an external administration service. The interface is white-labelled through a shell-configuration layer (application and assistant names, branding assets, and disclaimers), and the deployment used in this work is configured with the \elagenteG{} profile. The shared-document schema is treated as a stable contract and documented in the repository, which also enables integration of alternative custom viewers.

%% file: includes/si_tex/si-benchmark-llms.tex
\begin{table}[H]
\centering
\caption{Pricing (per million tokens) and maximum context window size for benchmarked \acp{llm}. Data were sourced from official provider websites for proprietary models and OpenRouter for open models. Pricing for cached tokens was not considered since cashing policies vary across providers.}
\label{tab:llm_specs}
\begin{tabular}{lccc}
\toprule
\textbf{LLM model} & \textbf{Input tokens (\$/M)} & \textbf{Output tokens (\$/M)} & \textbf{Maximum context window} \\
\midrule
gpt-4.1 & 2 & 8 & 1,047,576 \\
gpt-5 & 1.25 & 10 & 400,000 \\
gpt-5.1 & 1.25 & 10 & 400,000 \\
gpt-5.2 & 1.75 & 14 & 400,000 \\
minimax-m2 & 0.2 & 1 & 205,000 \\
qwen3-max & 1.2 & 6 & 262,144 \\
claude-sonnet-3.7 & 3 & 15 & 200,000 \\
claude-sonnet-4.5 & 3 & 15 & 200,000 \\
\bottomrule
\end{tabular}
\end{table}

\begin{lstlisting}[
basicstyle=\ttfamily\normalsize,
% frame=single,
breaklines=true,
columns=fullflexible,
xleftmargin=0.5em,
xrightmargin=0.5em
]
"gpt-4.1-2025-04-14": {"temperature": 0.0, "provider_name": "openai", "provider_url": "https://api.openai.com/v1/"}

\end{lstlisting}

\begin{lstlisting}[
basicstyle=\ttfamily\normalsize,
% frame=single,
breaklines=true,
columns=fullflexible,
xleftmargin=0.5em,
xrightmargin=0.5em
]
"gpt-5-2025-08-07": {"temperature": 1.0, "openai_reasoning_effort": "low", "provider_name": "openai", "provider_url": "https://api.openai.com/v1/"}

\end{lstlisting}

\begin{lstlisting}[
basicstyle=\ttfamily\normalsize,
% frame=single,
breaklines=true,
columns=fullflexible,
xleftmargin=0.5em,
xrightmargin=0.5em
]
"gpt-5.1-2025-11-13": {"temperature": 1.0, "openai_reasoning_effort": "low", "provider_name": "openai", "provider_url": "https://api.openai.com/v1/"}

\end{lstlisting}

\begin{lstlisting}[
basicstyle=\ttfamily\normalsize,
% frame=single,
breaklines=true,
columns=fullflexible,
xleftmargin=0.5em,
xrightmargin=0.5em
]
"gpt-5.2-2025-12-11": {"temperature": 1.0, "openai_reasoning_effort": "low", "provider_name": "openai", "provider_url": "https://api.openai.com/v1/"}

\end{lstlisting}

\begin{lstlisting}[
basicstyle=\ttfamily\normalsize,
% frame=single,
breaklines=true,
columns=fullflexible,
xleftmargin=0.5em,
xrightmargin=0.5em
]
"openrouter:minimax/minimax-m2": {"temperature": 1.0, "provider_name": "openrouter", "provider_url": "https://openrouter.ai/api/v1"}

\end{lstlisting}

\begin{lstlisting}[
basicstyle=\ttfamily\normalsize,
% frame=single,
breaklines=true,
columns=fullflexible,
xleftmargin=0.5em,
xrightmargin=0.5em
]
"openrouter:qwen/qwen3-max": {"temperature": 0.0, "provider_name": "openrouter", "provider_url": "https://openrouter.ai/api/v1"}

\end{lstlisting}

\begin{lstlisting}[
basicstyle=\ttfamily\normalsize,
% frame=single,
breaklines=true,
columns=fullflexible,
xleftmargin=0.5em,
xrightmargin=0.5em
]
"claude-3-7-sonnet-20250219": {"temperature": 1.0, "anthropic_thinking": {"type": "enabled", "budget_tokens": 1024}, "extra_headers": {"anthropic-beta": "token-efficient-tools-2025-02-19", "provider_name": "anthropic", "provider_url": "https://api.anthropic.com"}}

\end{lstlisting}

\begin{lstlisting}[
basicstyle=\ttfamily\normalsize,
% frame=single,
breaklines=true,
columns=fullflexible,
xleftmargin=0.5em,
xrightmargin=0.5em
]
"claude-sonnet-4-5-20250929": {"temperature": 0.0, "kind": "response", "provider_name": "anthropic", "provider_url": "https://api.anthropic.com"}

\end{lstlisting}

\vspace{0.75em}

%% file: includes/si_tex/si-benchmark-questions.tex
Prompts were minimally revised to remove ORCA-specific data and ensure an unbiased comparison with the original El Agente~\cite{Zou2025ElAgente} benchmark questions; no additional instructions were given. Functionals and basis sets were adjusted specifically for PySCF compatibility.

\prompt{ \textbf{Organic compounds level 1}

Perform in parallel geometry optimization of the [compounds below] with the Hartree-Fock (HF) method and def2-SVP basis set in the gas phase. Once the calculations have been successfully completed, please generate individual reports for each of the molecules listed below, one at a time. Each report should include the final Cartesian coordinates (in Å), total energy (in Hartrees), point group symmetry, dipole moment (in Debye), molecular orbital analysis (including an MO energy table and the HOMO-LUMO gap), atomic charge analysis (Mulliken, Löwdin, and IAO).\\
Organic Compounds:\\
1. caffeine (SMILES: CN1C=NC2=C1C(=O)N(C(=O)N2C)C)\\
2. theobromine (SMILES: CN1C=NC2=C1C(=O)NC(=O)N2C)\\
3. acetylsalicylic\_acid (SMILES: CC(=O)OC1=CC=CC=C1C(=O)O)
}

\prompt{ \textbf{Organic compounds level 2}

Perform in parallel geometry optimization using the xyz files listed below from the default working directory with the Hartree-Fock (HF) method and def2-SVP basis set in the gas phase. Once the calculations have been successfully completed, please generate individual reports for each of the molecules listed below, one at a time. Each report should include the final Cartesian coordinates (in Å), total energy (in Hartrees), point group symmetry, dipole moment (in Debye), molecular orbital analysis (including an MO energy table and the HOMO-LUMO gap), atomic charge analysis (Mulliken, Löwdin, and Hirshfeld).\\
Organic Compounds:\\
1. caffeine\_openbabel.xyz (charge = 0; multiplicity = 1)\\
2. theobromine\_openbabel.xyz (charge = 0; multiplicity = 1)\\
3. aspirin\_openbabel.xyz (charge = 0; multiplicity = 1)\\
4. methyl\_salicylate\_openbabel.xyz (charge = 0; multiplicity = 1)\\
5. acetaminophen\_openbabel.xyz (charge = 0; multiplicity = 1)\\
6. triazaadamantane\_openbabel.xyz (charge = 0; multiplicity = 1)\\
7. limonene\_openbabel.xyz (charge = 0; multiplicity = 1)\\
8. D-glucose\_openbabel.xyz (charge = 0; multiplicity = 1)\\
9. creatinine\_amine\_tautomer\_openbabel.xyz (charge = 0; multiplicity = 1)\\
10. creatinine\_imine\_tautomer\_openbabel.xyz (charge = 0; multiplicity = 1)\\
11. L-phenylalanine\_zwitterion\_openbabel.xyz (charge = 0; multiplicity = 1)\\
12. 2-chloronitrobenzene\_openbabel.xyz (charge = 0; multiplicity = 1)\\
13. cis-1\_2-cyclohexanediol\_openbabel.xyz (charge = 0; multiplicity = 1)\\
14. L-histidine\_non\_zwitterion\_openbabel.xyz (charge = 0; multiplicity = 1)\\
15. 2\_2-biphenol\_openbabel.xyz (charge = 0; multiplicity = 1)\\
16. S-2-ethyl-2-fluoropentan-1-ol\_openbabel.xyz (charge = 0; multiplicity = 1)\\
17. R-3-hydroxycyclopentan-1-one\_openbabel.xyz (charge = 0; multiplicity = 1)\\
18. 3-methylbutanoate\_anion\_openbabel.xyz (charge = -1; multiplicity = 1)\\
19. diisopropylamide\_anion\_openbabel.xyz (charge = -1; multiplicity = 1)\\
20. diisopropylammonium\_cation\_openbabel.xyz (charge = +1; multiplicity = 1)
}

\prompt{ \textbf{Inorganic compounds level 1}

Plan and act directly. Do not ask for my confirmation this time. Complete the following request: Perform in parallel geometry optimization of the [compounds below] with the Hartree-Fock (HF) method and def2-SVP basis set in the gas phase. Once the calculations have been successfully completed, please generate individual reports for each of the molecules listed below, one at a time. Each report should include the final Cartesian coordinates (in Å), total energy (in Hartrees), point group symmetry, dipole moment (in Debye), molecular orbital analysis (including an MO energy table and the HOMO-LUMO gap), atomic charge analysis (Mulliken, Löwdin, and IAO).\\
Inorganic Compounds:\\
1. Chromium(0) hexacarbonyl (low spin) -- SMILES: [Cr](=C=O)(=C=O)(=C=O)(=C=O)(=C=O)(=C=O)\\
2. Chlorine trifluoride -- SMILES: FCl(F)F\\
3. Fluorophosphoric acid (singly deprotonated form) -- SMILES: [O-]P(F)(O)=O
}

\prompt{ \textbf{Inorganic compounds level 2}

Plan and act directly. Do not ask my confirmation this time. Complete the following request:\\
Perform in parallel geometry optimization using the xyz files listed below with the Hartree-Fock (HF) method and def2-SVP basis set in the gas phase. Once the calculations have been successfully completed, please generate individual reports for each of the molecules listed below, one at a time. Each report should include the final Cartesian coordinates (in Å), total energy (in Hartrees), point group symmetry, dipole moment (in Debye), molecular orbital analysis (including an MO energy table and the HOMO-LUMO gap), atomic charge analysis (Mulliken, Löwdin, and Hirshfeld).\\
List of Inorganic Compounds:\\
1. chromium\_hexacarbonyl.xyz (charge = 0; multiplicity = 1)\\
2. chlorine\_trifluoride.xyz (charge = 0; multiplicity = 1)\\
3. fluorophosphoric\_acid\_singly\_deprotonated\_form.xyz (charge = -1; multiplicity = 1)\\
4. trifluoromethane\_sulfonate.xyz (charge = -1; multiplicity = 1)\\
5. cyclohexyldimethylphosphine.xyz (charge = 0; multiplicity = 1)\\
6. t-butylisothiocyanate.xyz (charge = 0; multiplicity = 1)\\
7. chromic\_acid.xyz (charge = 0; multiplicity = 1)\\
8. permanganic\_acid.xyz (charge = 0; multiplicity = 1)\\
9. perchlorate.xyz (charge = -1; multiplicity = 1)\\
10. hexafluorophosphate.xyz (charge = -1; multiplicity = 1)\\
11. tetrafluoroborate.xyz (charge = -1; multiplicity = 1)\\
12. dicyanoaurate.xyz (charge = -1; multiplicity = 1)\\
13. nitrogen\_trifluoride.xyz (charge = 0; multiplicity = 1)\\
14. sulfur\_hexafluoride.xyz (charge = 0; multiplicity = 1)\\
15. sulfur\_tetrafluoride.xyz (charge = 0; multiplicity = 1)\\
16. xenon\_tetrafluoride.xyz (charge = 0; multiplicity = 1)
}

\prompt{ \textbf{Carbocations compounds level 1}

A carbocation formation reaction is given by R-H -> R+ + H-. Your task is to calculate the carbocation formation enthalpies and Gibbs free energies for R-H = methane, ethane, propane, 2-methylpropane, toluene, benzene, dimethyl ether, trimethylamine, and propene. In your working directory, you can find: carbo\_ch4.xyz, carbo\_C2H6.xyz, carbo\_C3H8.xyz, carbo\_2-methylpropane.xyz, carbo\_toluene.xyz, carbo\_benzene.xyz, carbo\_et2o.xyz, carbo\_et3n.xyz, carbo\_propene.xyz, and carbo\_h-.xyz. You will also find carbo\_ch3+.xyz, carbo\_C2H5+.xyz, carbo\_C3H7+.xyz, carbo\_2-methylpropyl+.xyz, carbo\_toluene+.xyz, carbo\_benzene+.xyz, carbo\_et2o+.xyz, carbo\_et3n+.xyz, and carbo\_propene+.xyz for the cations. Please optimize these structures (except the hydride) using DFT with the B3LYP functional and def2-SVP basis set, and from the outputs, extract the relevant information to calculate the carbocation formation enthalpies and Gibbs free energies of each R-H. Report the results (in kcal/mol) in a table. For charge and multiplicity, for molecules charge 0, multiplicity 1; carbocations charge 1, multiplicity 1; hydride charge -1, multiplicity 1.
}

\prompt{ \textbf{Carbocations compounds level 2}

A carbocation formation reaction is given by R-H -> R+ + H-. Your task is to calculate the carbocation formation enthalpies and Gibbs free energies for R-H = methane, ethane, propane, 2-methylpropane, toluene, benzene, dimethyl ether, trimethylamine, and propene. The SMILES of each R-H are as follows: C, CC, CCC, CC(C)C, Cc1ccccc1, c1ccccc1, COC, CN(C)C, C=CC. The SMILES of each R+ is given by [CH3+], [CH2+]C, C[CH+]C, C[C+](C)C, [CH2+]c1c(cccc1), c1[c+]cccc1, CO[CH2+], CN(C)[CH2+], [CH2+]C=C. Please use the SMILES strings in the table to generate the appropriate geometries, optimize them using DFT with the B3LYP functional and def2-SVP basis set, and from the outputs, extract the relevant information to calculate the carbocation formation enthalpies and Gibbs free energies of each R-H. Report the results (in kcal/mol) in a table.
}

\prompt{ \textbf{Ring Strain compounds level 1}

Compute the values of ∆H and ∆G for the following reactions:\\
cyclo(CnH2n) → cyclo(Cn-1H2n-3)-CH3\\
Perform these calculations using B3LYP/def2-svp for values of n from 4 to 8 and use them to approximate the relative ring strain energies of cycloalkanes of size 3 to 8.\\
Hint: The first reaction (n = 4) is cyclobutane (SMILES string C1CCC1) converting into methylcyclopropane (SMILES string CC1CC1). Each structure needs to be optimized and frequencies must be calculated to get the enthalpy and Gibbs free energies. You will need to pick a reference point to use as the ``zero ring strain'' point and compare the others relative to that. Report a table of ring size vs. ring strain enthalpy and free energy. To calculate the ring strain energy, start by assuming cyclooctane (n = 8) is the reference point, and that its ring strain is zero. Then, the ring strain of cycloheptane (n = 7) is determined by the enthalpy or Gibbs free energy of the reaction cyclooctane → methylcycloheptane (n = 8 to n = 7), and the ring strain energy of cyclooctane (n = 8). Obtain this for n = 8 to n = 3. Finally, use cyclohexane (n = 6) as the reference point of zero ring strain.
}

\prompt{ \textbf{Ring Strain compounds level 2}

Compute the values of ∆H and ∆G for the following reactions:\\
cyclo(CnH2n) → cyclo(Cn-1H2n-3)-CH3\\
Perform these calculations using B3LYP/def2-svp for values of n from 4 to 8 and use them to approximate the relative ring strain energies of cycloalkanes of size 3 to 8.\\
Hint: You will need to pick a reference point to use as the ``zero ring strain'' point and compare the others relative to that. Report a table of ring size vs. ring strain enthalpy and free energy. The ring strain of cyclo(CnH2n) is determined by the reaction energy of cyclo(CnH2n) → cyclo(Cn-1H2n-3)-CH3 and the ring strain energy of cyclo(CnH2n).
}

\prompt{ \textbf{pKa Prediction compounds level 1}

Plan and act directly. Do not ask for my confirmation this time. Complete the following request: Calculate the pKa of acetic acid in water using two calculations at the B3LYP/def2-SVP level of theory with the CPCM implicit solvation model.
}

\prompt{ \textbf{pKa Prediction compounds level 2}

Plan and act directly. Do not ask my confirmation this time. Complete the following request: Calculate the pKa of chlorofluoroacetic acid using B3LYP def2-SVP.\\
\\
To do so, first calibrate the free energy of solvation of the proton based on the known literature values of some related carboxylic acids:\\
\\
1. Acetic acid; pKa = 4.76\\
2. Fluoroacetic acid; pKa = 2.586\\
3. Chloroacetic acid; pKa = 2.86
}

\prompt{ \textbf{TDDFT compounds level 1}

Compute the energy level of S1, the energy difference between S1 and T1, and the oscillator strength to the S1 state for the following structures from the default working directory: tddft\_2.xyz, tddft\_3.xyz, tddft\_5.xyz. Perform a single-point TDDFT (after geometry optimization and checking for geometric stability) calculation with B3LYP/def2-SVP.
}

\prompt{ \textbf{TDDFT compounds level 2}

Compute the energy level of S1, the energy difference between S1 and T1, and the oscillator strength to the S1 state for the following structures from the default working directory: tddft\_2.xyz, tddft\_3.xyz, tddft\_5.xyz. Perform a single-point TDDFT calculation with B3LYP/def2-SVP.
}

%% file: includes/si_tex/si-benchmark-rubrics.tex
We reused the task-specific rubrics developed for the evaluation of \elagenteQ{}~\cite{Zou2025ElAgente}. In the present benchmark, \texttt{gpt-4o} served as an independent \ac{llm} judge and applied each rubric to the full \grafico{} execution trace. It returned a score from 0 to 1, with 1.0 indicating that all task-specific criteria were satisfied. The earlier \elagenteQ{} scores were assigned by human evaluators using the same criteria.
This dual evaluation was needed because repeated runs could perform comparable underlying calculations while differing in workflow structure and reporting. The deterministic evaluator assessed structured numerical outputs without depending on their textual presentation, whereas the rubric-based judge assessed task completeness, reasoning and reporting from the full trace.

\renewcommand{\arraystretch}{0.90}

\begin{longtable}{p{0.1\linewidth} p{0.05\linewidth} p{0.06\linewidth} p{0.62\linewidth}}
\caption{The rubric used by the LLM judge when grading the benchmark questions.} \label{tab:rubric} \\
\toprule
\textbf{Question} & \textbf{Level} & \textbf{Weight} & \textbf{Task} \\
\midrule
\endfirsthead

\multicolumn{4}{c}%
{{\bfseries \tablename\ \thetable{} -- continued from previous page}} \\
\toprule
\textbf{Question} & \textbf{Level} & \textbf{Weight} & \textbf{Task} \\
\midrule
\endhead

\midrule
\bottomrule
\endlastfoot


\multirow{10}{*}{ \raisebox{50pt}[\dimexpr\height+\depth\relax]{\rotatebox[origin=r]{90}{Organic compounds}}}
& \multirow{5}{*}{1} & 20\% & Correct input file, i.e., level of theory, required keywords, charge, and multiplicity; \\
& & 20\% & Convergence of calculation, i.e., SCF, geometry optimization, and absence of imaginary frequency; \\
& & 20\% & Report generation; \\
& & 20\% & Successful extraction and documentation of all report values; \\
& & 20\% & Successful generation of XYZ from SMILES. \\
\cmidrule{2-4}
& \multirow{5}{*}{2} & 20\% & Correct input file, i.e., level of theory, required keywords, charge, and multiplicity; \\
& & 20\% & Convergence of calculation, i.e., SCF, geometry optimization, and absence of imaginary frequency; \\
& & 20\% & Report generation; \\
& & 20\% & Successful extraction and documentation of all report values; \\
& & 20\% & Successful processing of all input XYZ. \\
\midrule

\multirow{10}{*}{ \raisebox{55pt}[\dimexpr\height+\depth\relax]{\rotatebox[origin=r]{90}{Inorganic compounds}}}
& \multirow{5}{*}{1} & 20\% & If correct input file, i.e., level of theory, required keywords, charge, and multiplicity; \\
& & 20\% & If convergence of calculation, i.e., SCF, geometry optimization, and absence of imaginary frequency; \\
& & 20\% & If generated a report; \\
& & 20\% & If successful extraction and documentation of all report values; \\
& & 20\% & If generated of XYZ from SMILES. \\
\cmidrule{2-4}
& \multirow{5}{*}{2} & 20\% & If correct input file, i.e., level of theory, required keywords, charge, and multiplicity; \\
& & 20\% & If convergence of calculation, i.e., SCF, geometry optimization, and absence of imaginary frequency; \\
& & 20\% & If generated a report; \\
& & 20\% & If successful extraction and documentation of all report values; \\
& & 20\% & If successful processing of all input XYZ. \\
\midrule

\multirow{6}{*}{ \raisebox{30pt}[\dimexpr\height+\depth\relax]{\rotatebox[origin=r]{90}{Carbocations}}}
& \multirow{3}{*}{1} & 40\% & Molecules: correct input file, output geometry, completed calculation, and data extraction; \\
& & 40\% & Carbocations: correct input file, output geometry, completed calculation, and data extraction; \\
& & 20\% & Results: correct $\Delta H$ and $\Delta G$ trends, and match chemical intuition. \\
\cmidrule{2-4}
& \multirow{3}{*}{2} & 40\% & Molecules: correct input file, output geometry, completed calculation, and data extraction; \\
& & 40\% & Carbocations: correct input file, output geometry, completed calculation, and data extraction; \\
& & 20\% & Results: correct $\Delta H$ and $\Delta G$ trends, and match chemical intuition. \\
\midrule

\newpage
\multirow{20}{*}{ \raisebox{100pt}[\dimexpr\height+\depth\relax]{\rotatebox[origin=r]{90}{Ring Strain}}}
& \multirow{10}{*}{1} & 10\% & Correct structures from the formula; \\*
& & 10\% & Reasonable energy scale (not Hartree); \\
& & 10\% & No imaginary frequencies; \\
& & 10\% & Performs just the right number of calculations (no strange single point); \\
& & 10\% & Consistent level of theory used (DFT, basis set, solvent model if any); \\
& & 10\% & Correctly extracted enthalpy and Gibbs free energies; \\
& & 10\% & Correct reference energy (cyclohexane); \\
& & 10\% & Correct ring strain magnitude (from extracted values); \\
& & 10\% & Correct sign for ring strain; \\
& & 10\% & All values reported (cyclopropane, Gibbs, and Enthalpy). \\
\cmidrule{2-4}
& \multirow{10}{*}{2} & 10\% & Correct structures from the formula; \\*
& & 10\% & Reasonable energy scale (not Hartree); \\
& & 10\% & No imaginary frequencies; \\
& & 10\% & Performs just the right number of calculations (no strange single point); \\
& & 10\% & Consistent level of theory used (DFT, basis set, solvent model if any); \\
& & 10\% & Correctly extracted enthalpy and Gibbs free energies; \\
& & 10\% & Correct reference energy (cyclohexane); \\
& & 10\% & Correct ring strain magnitude (from extracted values); \\
& & 10\% & Correct sign for ring strain; \\
& & 10\% & All values reported (cyclopropane, Gibbs, and Enthalpy). \\
\midrule

\multirow{7}{*}{ \raisebox{20pt}[\dimexpr\height+\depth\relax]{\rotatebox[origin=r]{90}{pKa Prediction}}}
& \multirow{3}{*}{1} & 33\% & Correct input files, i.e., level of theory, required keywords, charge, and multiplicity; \\
& & 33\% & Convergence of calculation, i.e., SCF, geometry optimization, and absence of imaginary frequency; \\
& & 33\% & Computed correct $pK_a \sim 22.05$, anything roughly above 19 or below 25 is correct. \\
\cmidrule{2-4}
& \multirow{4}{*}{2} & 25\% & Correct input files, i.e., level of theory, required keywords, charge, and multiplicity; \\
& & 25\% & Convergence of calculation, i.e., SCF, geometry optimization, and absence of imaginary frequency; \\
& & 25\% & Calibrated proton solvation energy using linear regression or averaging; \\
& & 25\% & Computed reasonable value for $pK_a$ (e.g., a value between -2.70 and 1.50). \\
\midrule

\multirow{6}{*}{ \raisebox{30pt}[\dimexpr\height+\depth\relax]{\rotatebox[origin=r]{90}{TDDFT}}}
& \multirow{3}{*}{1} & 40\% & Correct input file (level of theory as requested, tddft block); \\
& & 40\% & Calculation completed normally; \\
& & 20\% & Extract and report the values correctly, and the number matched. \\
\cmidrule{2-4}
& \multirow{3}{*}{2} & 40\% & Correct input file (level of theory as requested, tddft block); \\
& & 40\% & Calculation completed normally; \\
& & 20\% & Extract and report the values correctly, and the number matched. \\
\end{longtable}

%% file: includes/si_tex/si-benchmark-pass-k.tex
We report pass@k~\cite{chen2021evaluating}, the probability that at least one of $k$ runs meets a specified criterion, and pass\^{}k~\cite{demystifying_evals_for_AI_agents}, the probability that all $k$ runs meet it. A run passed when its numerical score was 1.00 and its rubric-based score exceeded 0.90. This joint criterion required both numerical validity and correct task-specific interpretation and reporting. For example, a p$K_\mathrm{a}$ trace had to report the value obtained from the regression of the computed energies, rather than merely execute the underlying calculations. Under this criterion, \texttt{gpt-5} achieved pass@3 of 0.99 and pass\^{}3 of 0.54. Results for all models and thresholds are given in Supplementary~\autoref{si_tab:pass_rates_numerical_1_llm_judge_09} and the tables below.

\begin{table}[h]
\centering
\caption{Pass@k and Pass\^{}k rates with criteria: $\tau_{\text{numerical}} \ge 0.90$ and $\tau_{\text{llm\_judge}} \ge 0.90$.}
\label{si_tab:pass_rates_numerical_09_llm_judge_09}
\begin{tabular}{lcccc|cccc}
\toprule
& \multicolumn{4}{c}{\textbf{pass@k}} & \multicolumn{4}{c}{\textbf{pass\^{}k}} \\
\cmidrule(lr){2-5} \cmidrule(lr){6-9}
\textbf{LLM Model} & \textbf{@1} & \textbf{@3} & \textbf{@5} & \textbf{@10} & \textbf{\^{}1} & \textbf{\^{}3} & \textbf{\^{}5} & \textbf{\^{}10} \\
\midrule
gpt-4.1 & 0.81 & 0.99 & 1.00 & 1.00 & 0.81 & 0.52 & 0.34 & 0.11 \\
gpt-5 & 0.94 & 1.00 & 1.00 & 1.00 & 0.94 & 0.83 & 0.74 & 0.53 \\
gpt-5.1 & 0.87 & 1.00 & 1.00 & 1.00 & 0.87 & 0.65 & 0.48 & 0.22 \\
gpt-5.2 & 0.84 & 1.00 & 1.00 & 1.00 & 0.84 & 0.59 & 0.42 & 0.17 \\
minimax-m2 & 0.53 & 0.90 & 0.98 & 1.00 & 0.53 & 0.15 & 0.04 & 0.00 \\
qwen3-max & 0.53 & 0.90 & 0.98 & 1.00 & 0.53 & 0.15 & 0.04 & 0.00 \\
sonnet-3.7 & 0.57 & 0.93 & 0.99 & 1.00 & 0.57 & 0.19 & 0.06 & 0.00 \\
sonnet-4.5 & 0.78 & 0.99 & 1.00 & 1.00 & 0.78 & 0.48 & 0.29 & 0.08 \\
\bottomrule
\end{tabular}
\end{table}

\begin{table}[h]
\centering
\caption{Pass@k and Pass\^{}k rates with criteria: $\tau_{\text{numerical}} = 1$.}
\label{si_tab:pass_rates_numerical_1}
\begin{tabular}{lcccc|cccc}
\toprule
& \multicolumn{4}{c}{\textbf{pass@k}} & \multicolumn{4}{c}{\textbf{pass\^{}k}} \\
\cmidrule(lr){2-5} \cmidrule(lr){6-9}
\textbf{LLM Model} & \textbf{@1} & \textbf{@3} & \textbf{@5} & \textbf{@10} & \textbf{\^{}1} & \textbf{\^{}3} & \textbf{\^{}5} & \textbf{\^{}10} \\
\midrule
gpt-4.1 & 0.68 & 0.97 & 1.00 & 1.00 & 0.68 & 0.30 & 0.13 & 0.02 \\
gpt-5 & 0.82 & 1.00 & 1.00 & 1.00 & 0.82 & 0.56 & 0.38 & 0.13 \\
gpt-5.1 & 0.76 & 0.99 & 1.00 & 1.00 & 0.76 & 0.43 & 0.24 & 0.06 \\
gpt-5.2 & 0.82 & 1.00 & 1.00 & 1.00 & 0.82 & 0.56 & 0.38 & 0.13 \\
minimax-m2 & 0.55 & 0.91 & 0.98 & 1.00 & 0.55 & 0.16 & 0.05 & 0.00 \\
qwen3-max & 0.55 & 0.91 & 0.98 & 1.00 & 0.55 & 0.16 & 0.05 & 0.00 \\
sonnet-3.7 & 0.54 & 0.91 & 0.98 & 1.00 & 0.54 & 0.16 & 0.04 & 0.00 \\
sonnet-4.5 & 0.68 & 0.97 & 1.00 & 1.00 & 0.68 & 0.30 & 0.13 & 0.02 \\
\bottomrule
\end{tabular}
\end{table}

\begin{table}[h]
\centering
\caption{Pass@k and Pass\^{}k rates with criteria: $\tau_{\text{numerical}} = 1$ and $\tau_{\text{llm\_judge}} \ge 0.90$.}
\label{si_tab:pass_rates_numerical_1_llm_judge_09}
\begin{tabular}{lcccc|cccc}
\toprule
& \multicolumn{4}{c}{\textbf{pass@k}} & \multicolumn{4}{c}{\textbf{pass\^{}k}} \\
\cmidrule(lr){2-5} \cmidrule(lr){6-9}
\textbf{LLM Model} & \textbf{@1} & \textbf{@3} & \textbf{@5} & \textbf{@10} & \textbf{\^{}1} & \textbf{\^{}3} & \textbf{\^{}5} & \textbf{\^{}10} \\
\midrule
gpt-4.1 & 0.63 & 0.95 & 0.99 & 1.00 & 0.63 & 0.25 & 0.10 & 0.01 \\
gpt-5 & 0.82 & 0.99 & 1.00 & 1.00 & 0.82 & 0.54 & 0.36 & 0.12 \\
gpt-5.1 & 0.70 & 0.97 & 1.00 & 1.00 & 0.70 & 0.34 & 0.16 & 0.02 \\
gpt-5.2 & 0.75 & 0.99 & 1.00 & 1.00 & 0.75 & 0.42 & 0.23 & 0.05 \\
minimax-m2 & 0.43 & 0.81 & 0.94 & 1.00 & 0.42 & 0.07 & 0.01 & 0.00 \\
qwen3-max & 0.39 & 0.78 & 0.92 & 0.99 & 0.39 & 0.06 & 0.01 & 0.00 \\
sonnet-3.7 & 0.43 & 0.82 & 0.95 & 1.00 & 0.43 & 0.08 & 0.01 & 0.00 \\
sonnet-4.5 & 0.62 & 0.95 & 0.99 & 1.00 & 0.62 & 0.24 & 0.09 & 0.01 \\
\bottomrule
\end{tabular}
\end{table}

%% file: includes/si_tex/si-benchmark-plots.tex
All benchmarks were executed on a high-performance computing node equipped with four NVIDIA H100 (80 GB) GPUs. We chose a maximum concurrency of four agent runs. GPU4PySCF jobs were parallelized with up to three concurrent jobs per GPU to saturate device utilization, allowing up to twelve molecules to be processed simultaneously.

The quantum-chemistry exercises were predominantly sequential: model decisions were interleaved with deterministic calculations whose outputs constrained subsequent steps. For these tasks, the same-model comparison reported in the main text favoured the single-agent execution-graph architecture, which retained high task scores while requiring fewer model requests, tokens and wall-clock time than the earlier multi-agent architecture. This direction is consistent with a controlled study of agent-system scaling, in which every tested multi-agent architecture reduced performance by 39--70\% relative to a single-agent baseline on the sequential PlanCraft benchmark~\cite{scienceOfScalingAgent}. That range was measured in PlanCraft and contextualizes the role of task structure.

To further understand the impact of varying \ac{llm} behaviours, we introduce additional metrics:
(i) \textit{context window saturation}, which measures the ratio of the total tokens in the final \ac{llm} \ac{api} request to its maximum context window allowed by the provider;
(ii) \textit{error recovery cost}, which is associated with the exceptions and tracebacks provided to the model during error handling;
and (iii) \textit{carryover tokens}, defined as the ratio of accumulated cacheable tokens to the total tokens consumed throughout the agentic trace. This metric highlights the importance of careful cache management for efficient agentic systems over long-horizon tasks. While many providers offer reduced rates for cached tokens, implementation varies across providers. For simplicity, our calculations rely on raw token counts and original pricing rather than cached rates. The monetary cost could be further reduced by employing additional caching strategies\footnote{\url{https://platform.claude.com/docs/en/build-with-claude/prompt-caching}}\textsuperscript{,}\footnote{\url{https://openrouter.ai/docs/guides/best-practices/prompt-caching}} or deploying local models.

\begin{figure}[!ht]
    \centering
    \includegraphics[width=0.75\linewidth]{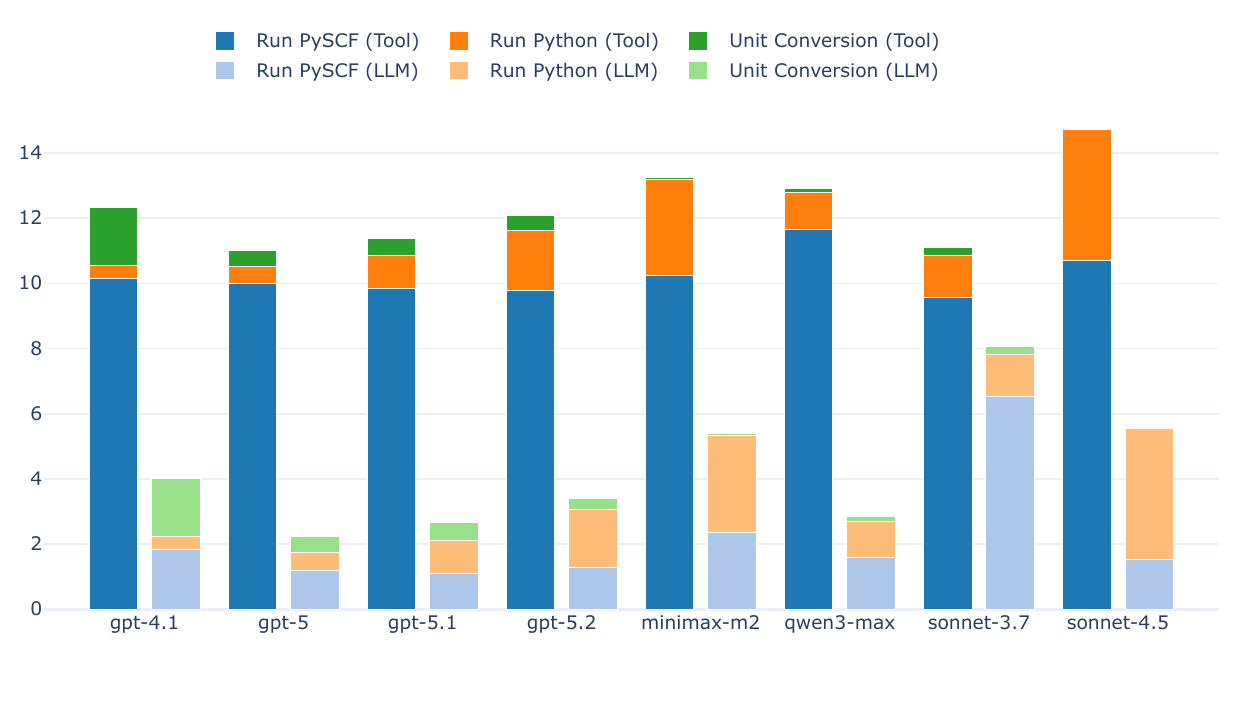}
    \caption{Average number of tool calls versus LLM API requests per model for the benchmark exercise. For each model, stacked bars compare three tool types (\texttt{run\_pyscf\_workflow}, \texttt{run\_python\_code}, \texttt{get\_conversion\_factor}), shown separately for actual tool calls and corresponding LLM requests (side-by-side groups).}
    \label{fig:si_tool_call_agg}
\end{figure}

\begin{figure}[!ht]
    \centering
    \includegraphics[width=\linewidth]{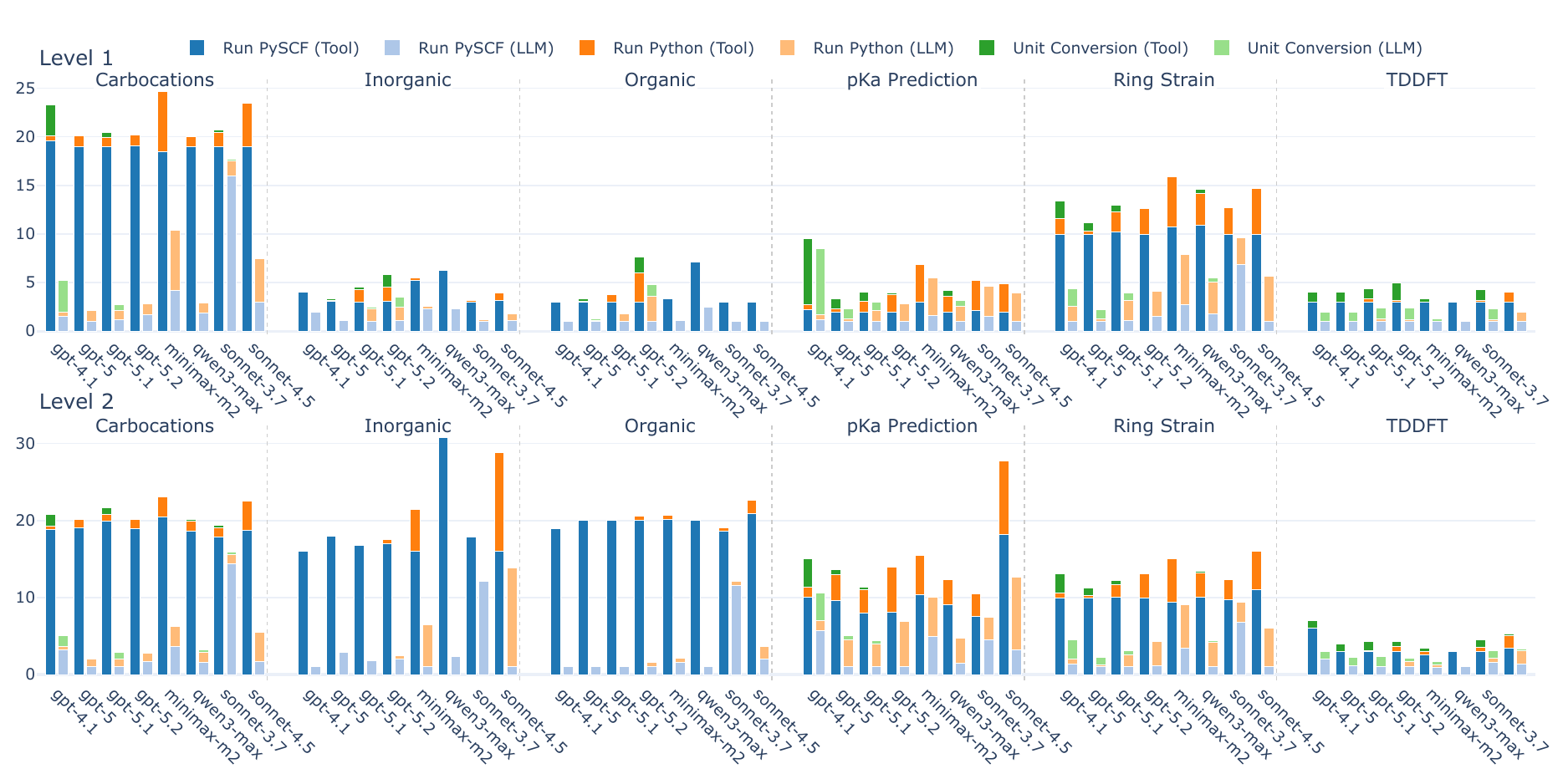}
    \caption{Tool usage versus LLM API requests by task and level for the benchmark exercise. For each task (grouped blocks) and model (ticks), stacked bars compare average tool calls (left bar) with corresponding LLM requests (right bar) for the \texttt{run\_pyscf\_workflow}, \texttt{run\_python\_code}, and \texttt{get\_conversion\_factor} tools; rows separate levels and colours encode tool type.}
    \label{fig:si_tool_call_details}
\end{figure}

We identify systematic differences in interaction patterns between the GPT and Claude model families that have direct implications for workflow efficiency. 
GPT models tend to batch tool invocations, dispatching multiple calls concurrently and returning to reason over aggregated results, whereas Claude models more frequently interleave tool calls with incremental reasoning. 
Within the Claude family, the elevated cost of \texttt{sonnet-3.7} is largely driven by its limited ability to parallelize tool calls, a known constraint that persists even under recommended system prompt configurations.\footnote{\url{https://platform.claude.com/docs/en/agents-and-tools/tool-use/implement-tool-use\#maximizing-parallel-tool-use}} \texttt{Sonnet-4.5} improves in its ability to invoke GPU4PySCF workflows in parallel but incurs high cost due to frequent calls to a general-purpose \texttt{run\_python\_code} tool for report generation (see Figs.~\ref{fig:si_tool_call_agg} and \ref{fig:si_tool_call_details}).
Notably, \texttt{sonnet-3.7} exhibits a recurring execution pattern in geometry optimization tasks, progressively reducing parallelism by issuing tool calls in batches of five, then three, two, and finally one, while \texttt{sonnet-4.5} repeatedly invokes excessive generic code execution calls for post-processing. 
These behaviours highlight that model capability alone does not determine system-level efficiency; rather, the alignment between a model's interaction style and the available tool abstractions plays a critical role. 
In such cases, encapsulating recurrent post-processing steps into dedicated tools could further mitigate unnecessary tool invocation overhead.

\begin{figure}
    \centering
    \includegraphics[width=\linewidth]{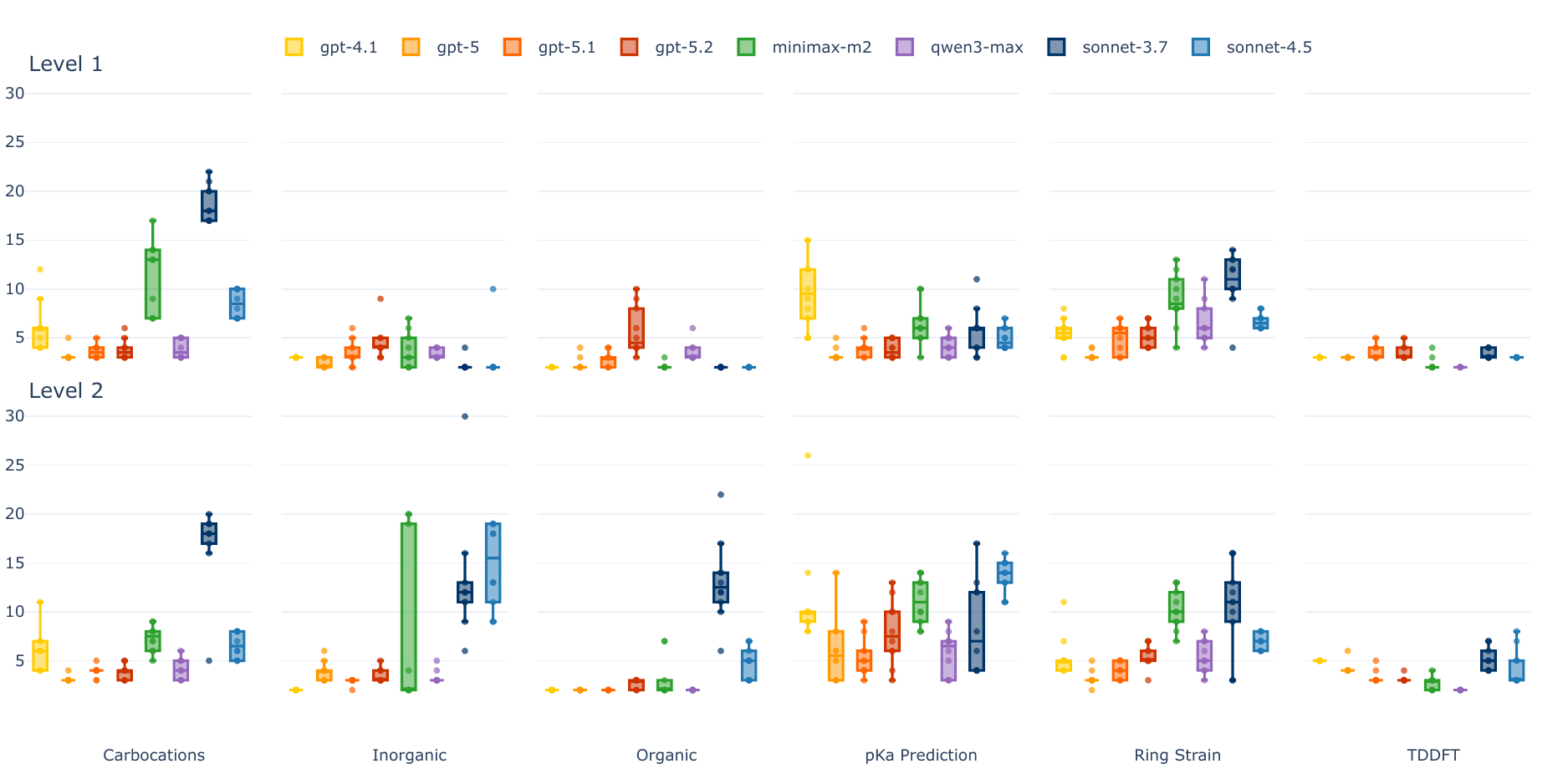}
    \caption{Distribution of LLM requests by model across tasks and levels. Boxes show per-model variability in LLM requests, with overlaid points for individual runs; facets separate tasks (columns) and levels (rows), and colours denote models.}
    \label{fig:si_number_of_llm_requests}
\end{figure}

\begin{figure}[!ht]
    \centering
    \includegraphics[width=\linewidth]{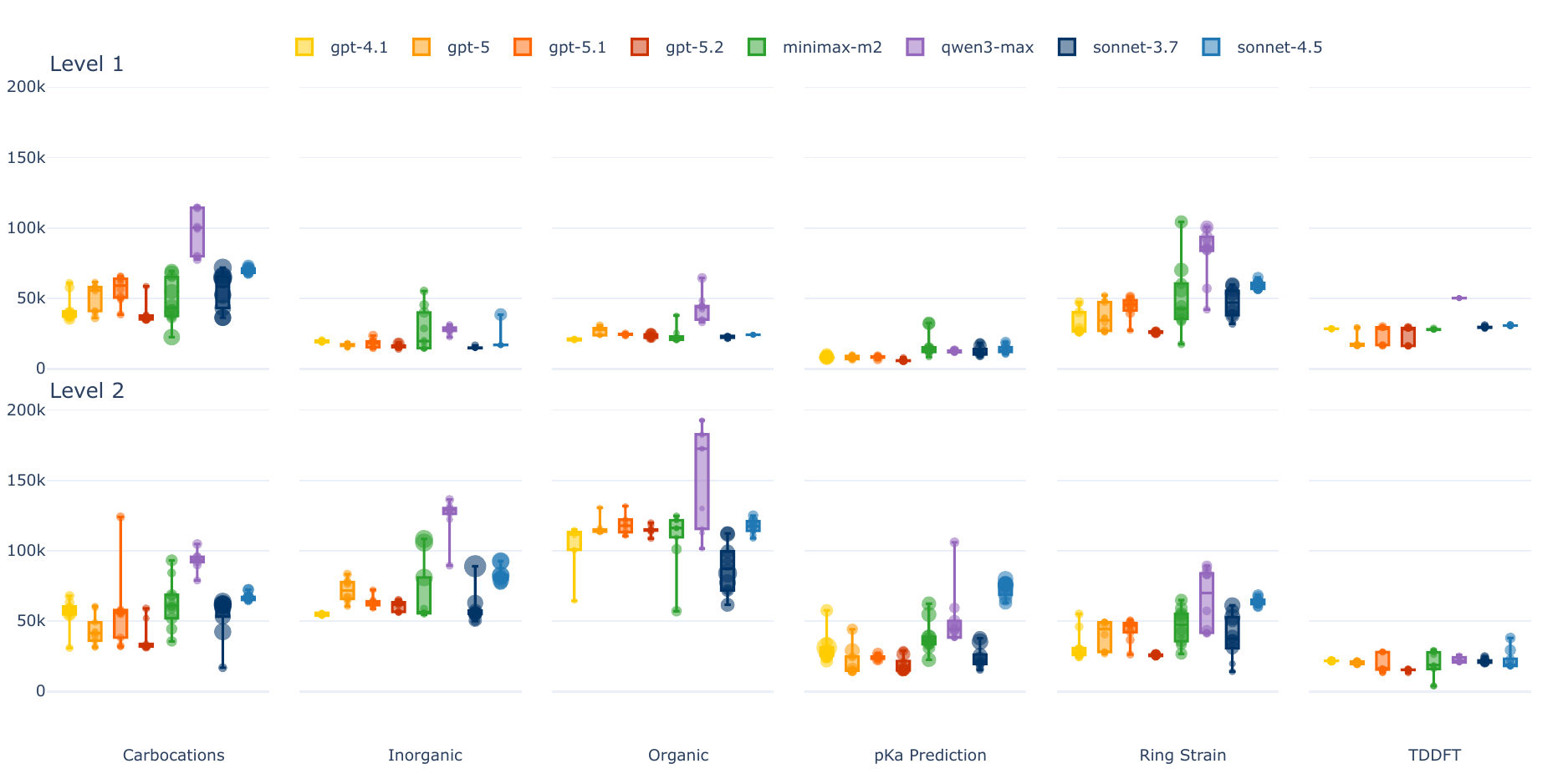}
    \caption{Distribution of final call tokens by model across tasks and levels. Boxes show per-model variability in final call total tokens, with overlaid points sized by number of LLM requests; facets separate tasks (columns) and levels (rows), and colours denote models.}
    \label{fig:si_last_call_tokens}
\end{figure}

\begin{figure}[!ht]
    \centering
    \includegraphics[width=\linewidth]{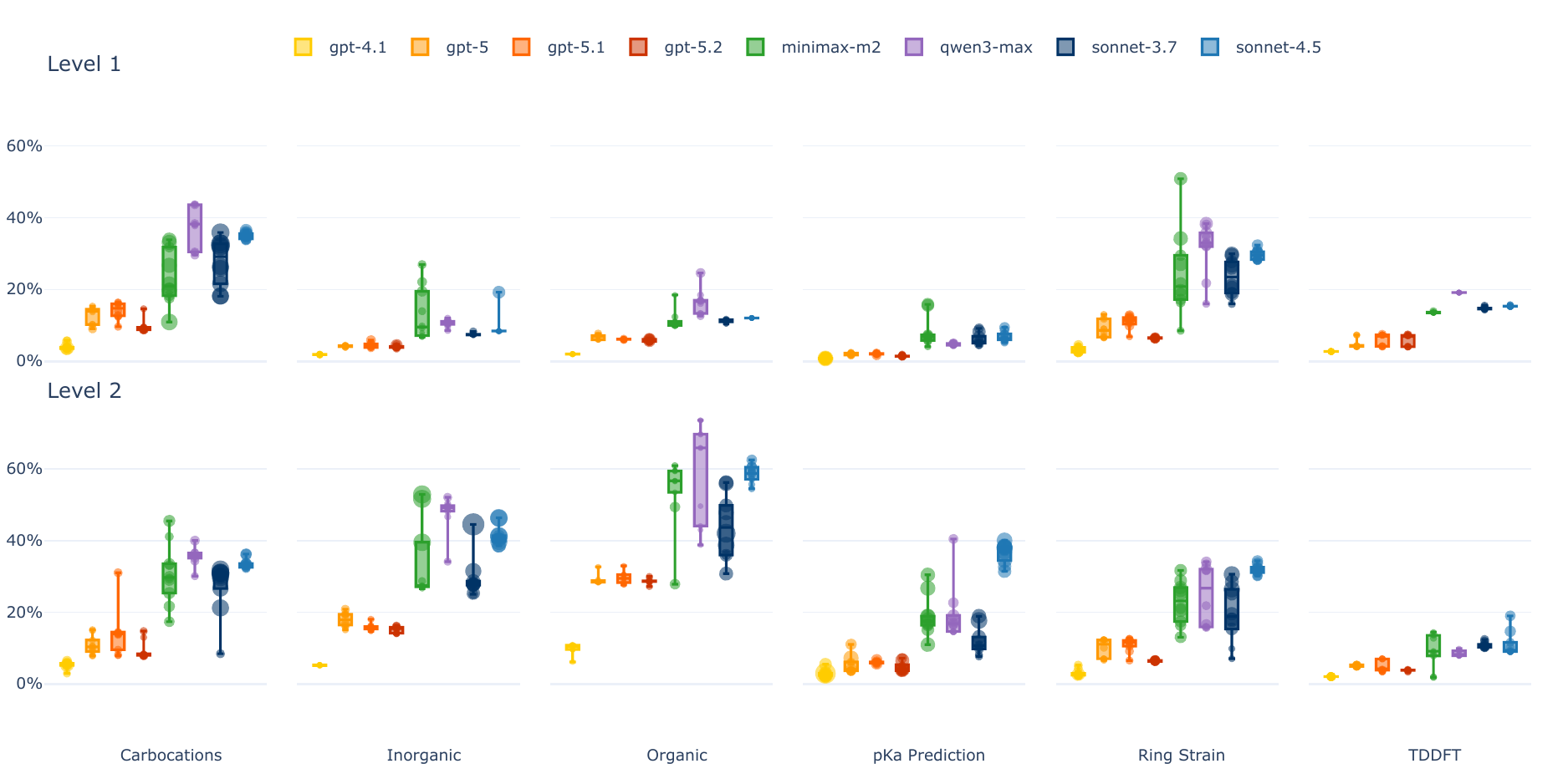}
    \caption{Distribution of final call context window pressure (final call tokens divided by the maximum context window) by model across tasks and levels. Boxes show per-model variability, with overlaid points representing individual runs scaled by the number of LLM requests; facets split tasks (columns) and levels (rows), with colours denoting models.}
    \label{fig:si_last_call_pressure}
\end{figure}

\begin{figure}[!ht]
    \centering
    \includegraphics[width=\linewidth]{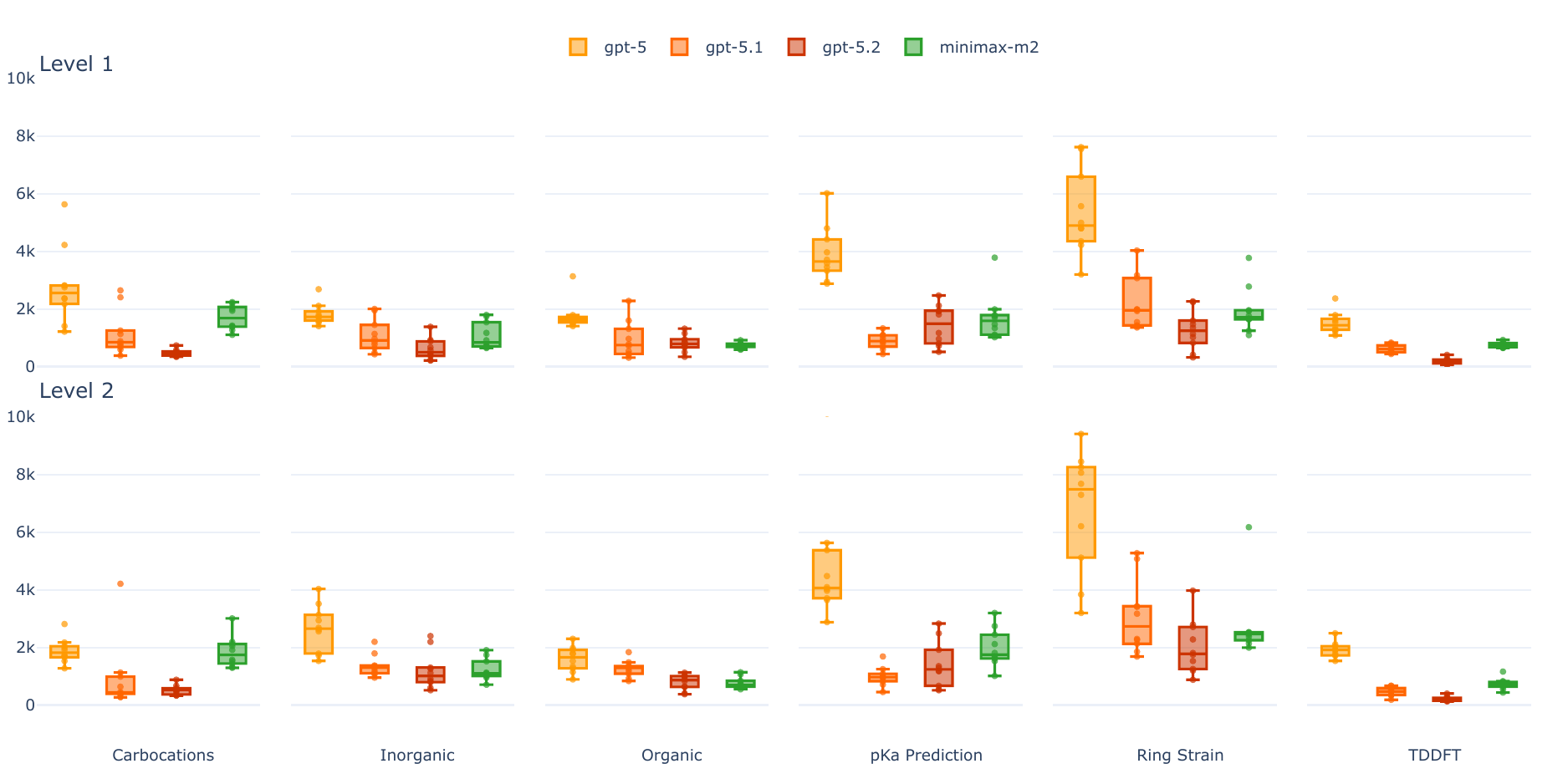}
    \caption{Distribution of reasoning token usage by model across tasks and levels. Boxes show per-model variability in reasoning tokens (filtered to LLM models with reasoning capabilities; \texttt{sonnet-3.7} has a consistent reasoning budget of 2024), with overlaid points for individual runs; facets split tasks (columns) and levels (rows), and colours denote models. One run from \texttt{minimax-m2} for Ring Strain (Level 2) consumed 102.91k reasoning tokens and is omitted from the plot due to the y-axis limit.}
    \label{fig:si_reasoning_tokens}
\end{figure}

For the box plots (Figs.~\ref{fig:si_number_of_llm_requests}, \ref{fig:si_last_call_tokens}, \ref{fig:si_last_call_pressure}, and \ref{fig:si_reasoning_tokens}), each box spans from quartile 1 (Q1) to quartile 3 (Q3). The second quartile (Q2) is marked by a line inside the box. By default, the whiskers extend to Q1 - 1.5$\times$IQR and Q3 + 1.5$\times$IQR, where IQR = Q3 - Q1.



\begin{figure}[!ht]
    \centering
    \includegraphics[width=\linewidth]{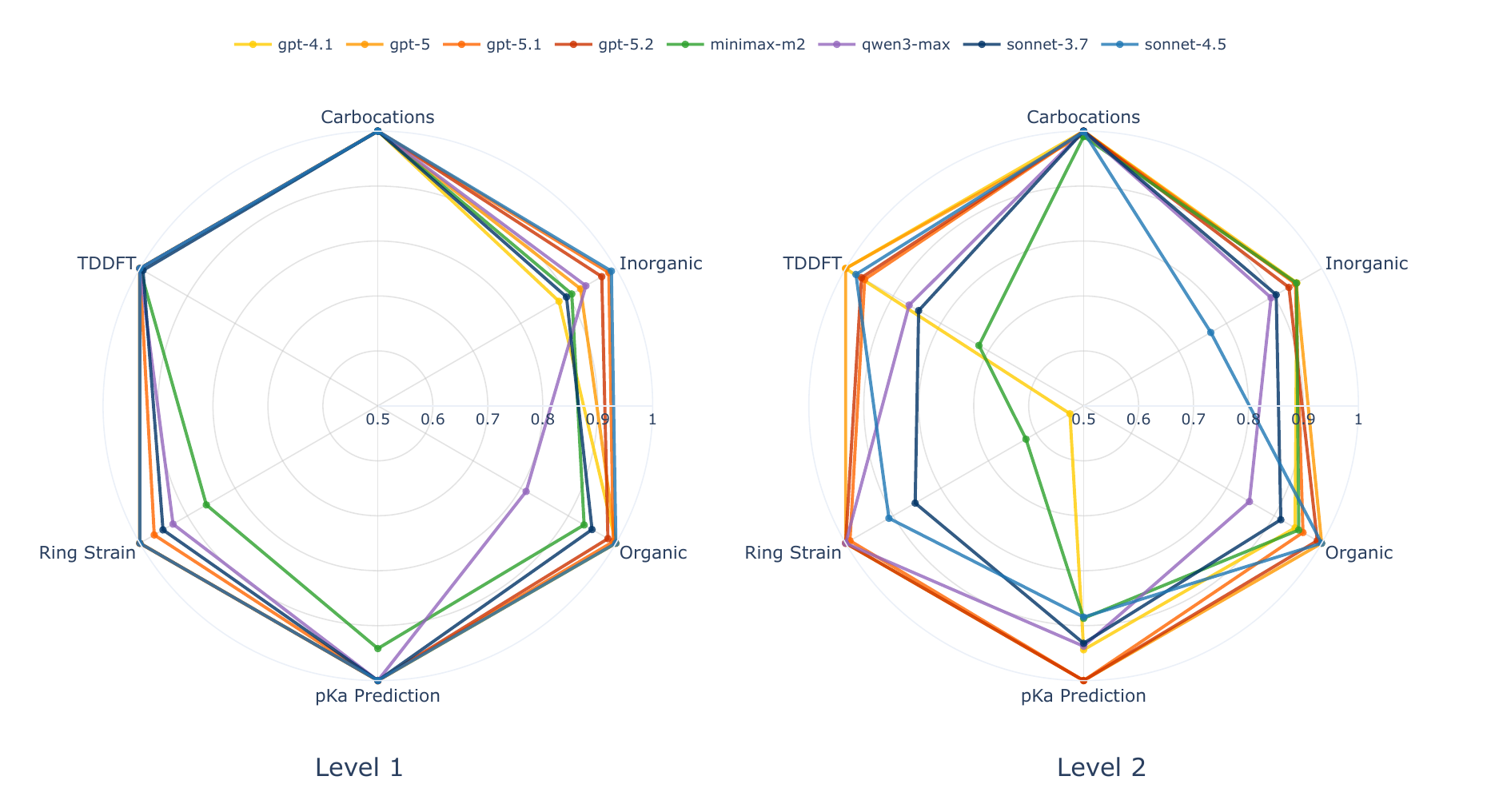}
    \caption{Radar plots of numerical-evaluator scores averaged over 10 runs by task and model for Level 1 and Level 2. Each polygon shows a model's mean score across tasks.}
    \label{fig:si_radar_numerical_evaluator}
\end{figure}

\begin{figure}[!ht]
    \centering
    \includegraphics[width=\linewidth]{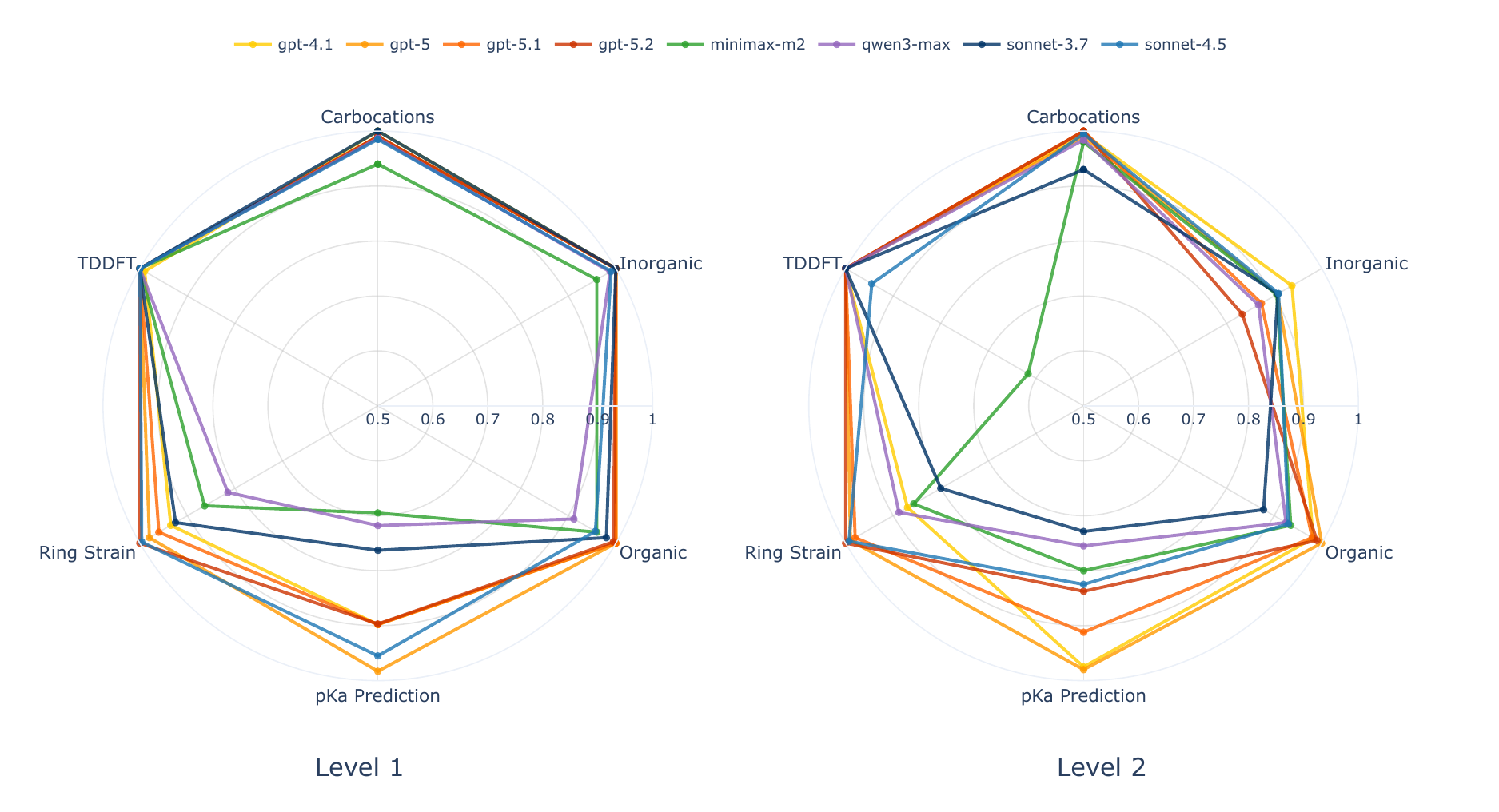}
    \caption{Radar plots of LLM-as-a-judge-evaluator scores averaged over 10 runs by task and model for Level 1 and Level 2. Each polygon shows a model's mean score across tasks.}
    \label{fig:si_radar_llm_judge_evaluator}
\end{figure}

\begin{figure}[!ht]
    \centering
    \includegraphics[width=\linewidth]{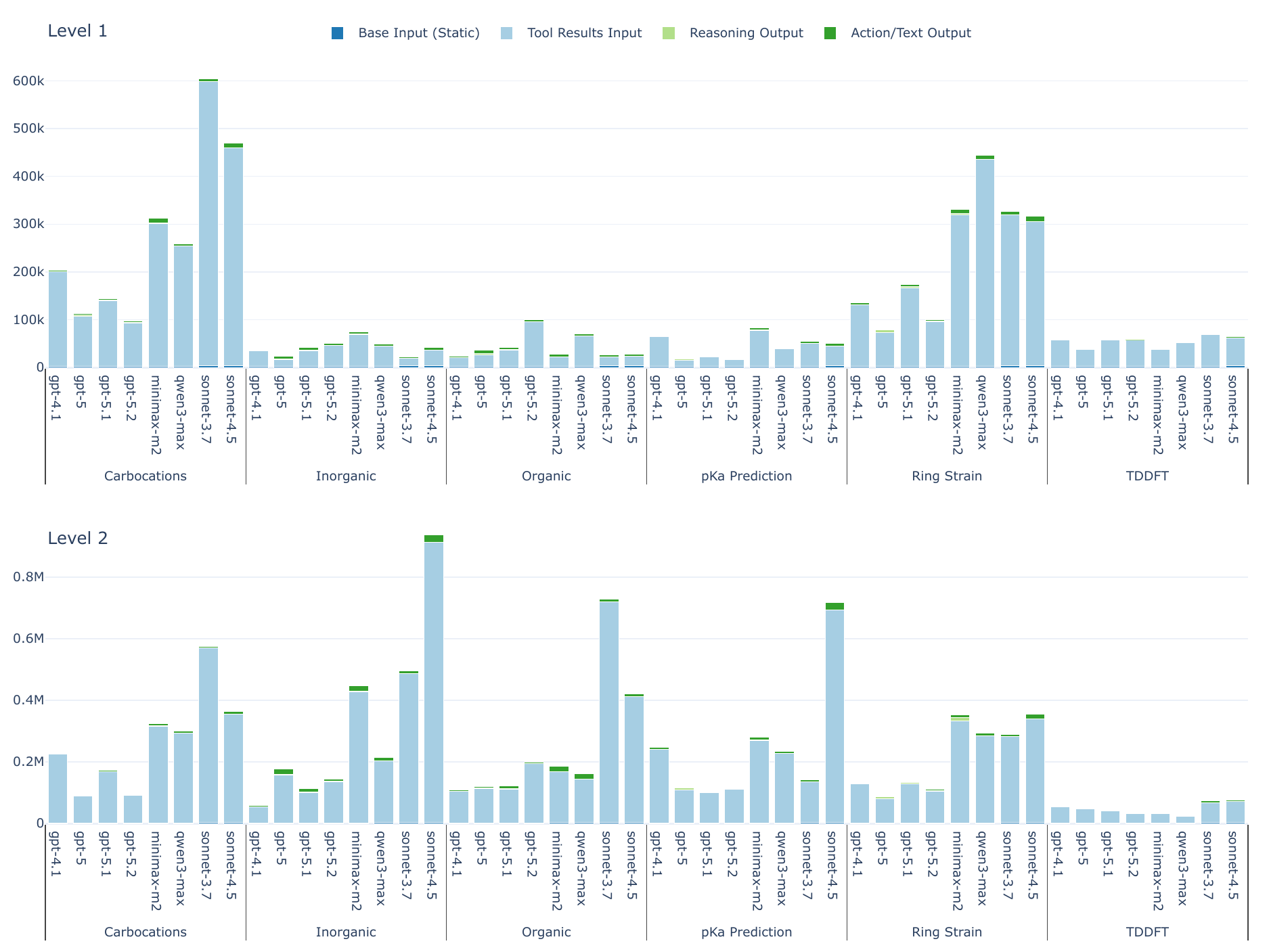}
    \caption{Mean token composition by generation source across tasks, models, and levels. Stacked bars categorize input tokens into base prompts (system prompts, tool schemas, and user task prompts) and tool-result inputs, and output tokens into reasoning tokens, action/tool-call outputs, and text outputs.}
    \label{fig:si_trace_input_output_token_composition}
\end{figure}

\begin{figure}[!ht]
    \centering
    \includegraphics[width=\linewidth]{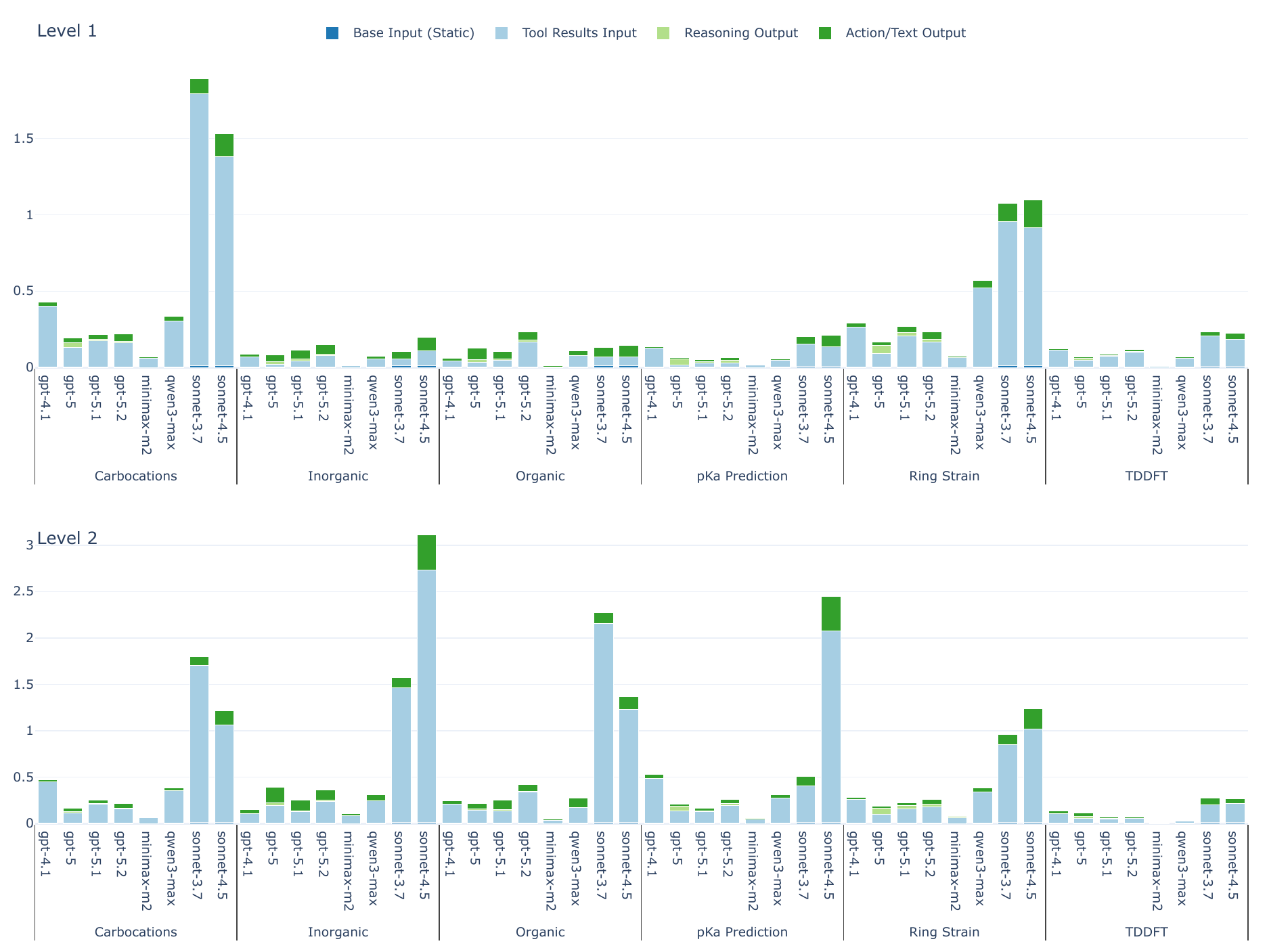}
    \caption{Same split as Fig.~\ref{fig:si_trace_input_output_token_composition}, but measured in USD cost calculated from pricing provided by LLM providers.}
    \label{fig:si_trace_input_output_usd_composition}
\end{figure}

\begin{figure}[!ht]
    \centering
    \includegraphics[width=\linewidth]{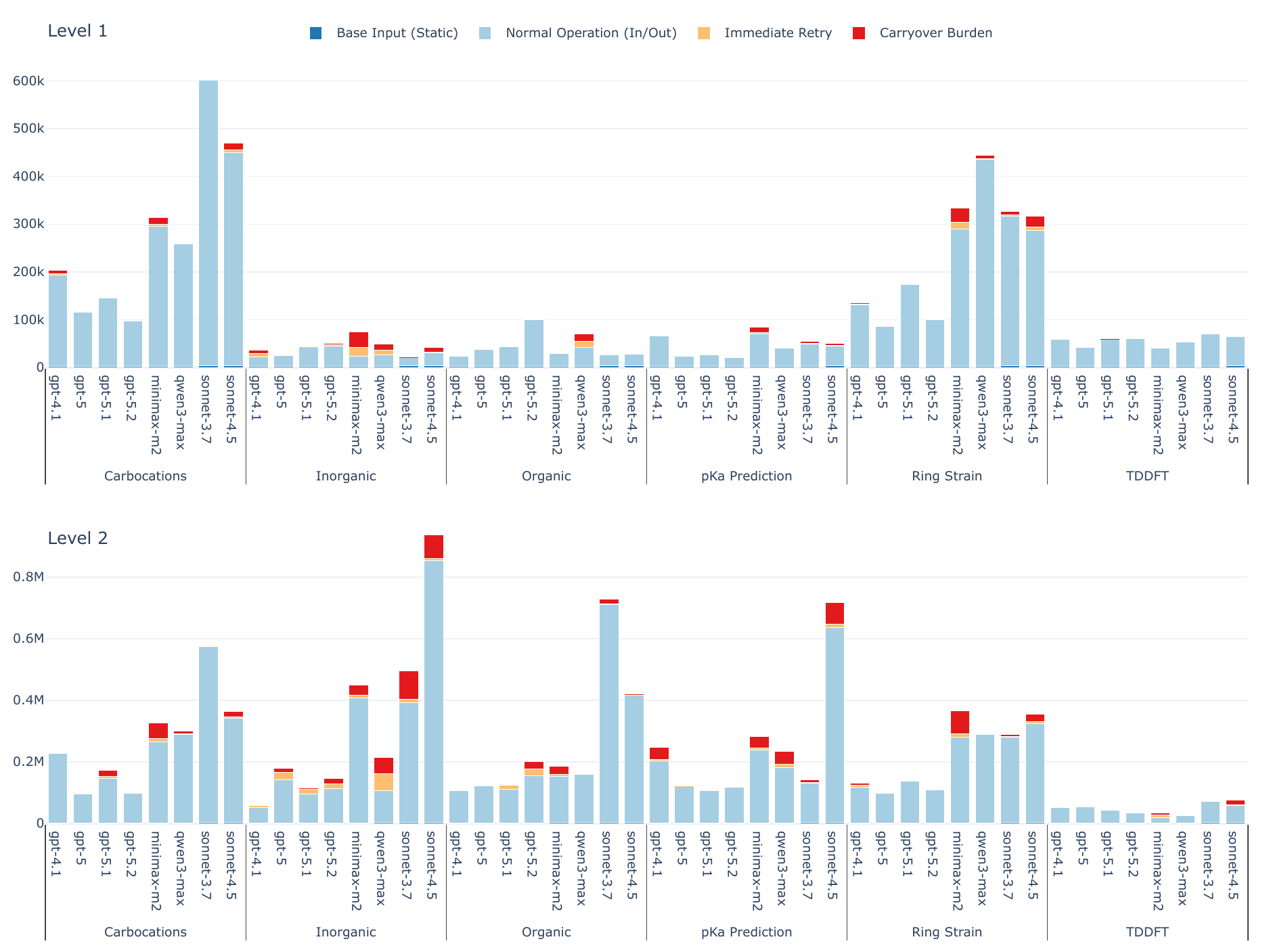}
    \caption{Mean token composition by role across tasks, models, and levels. Stacked bars partition total tokens into base input, normal operation (in/out), immediate retry, and carryover burden across agentic traces; rows separate Level 1 and Level 2.}
    \label{fig:si_trace_token_composition}
\end{figure}

\begin{figure}[!ht]
    \centering
    \includegraphics[width=\linewidth]{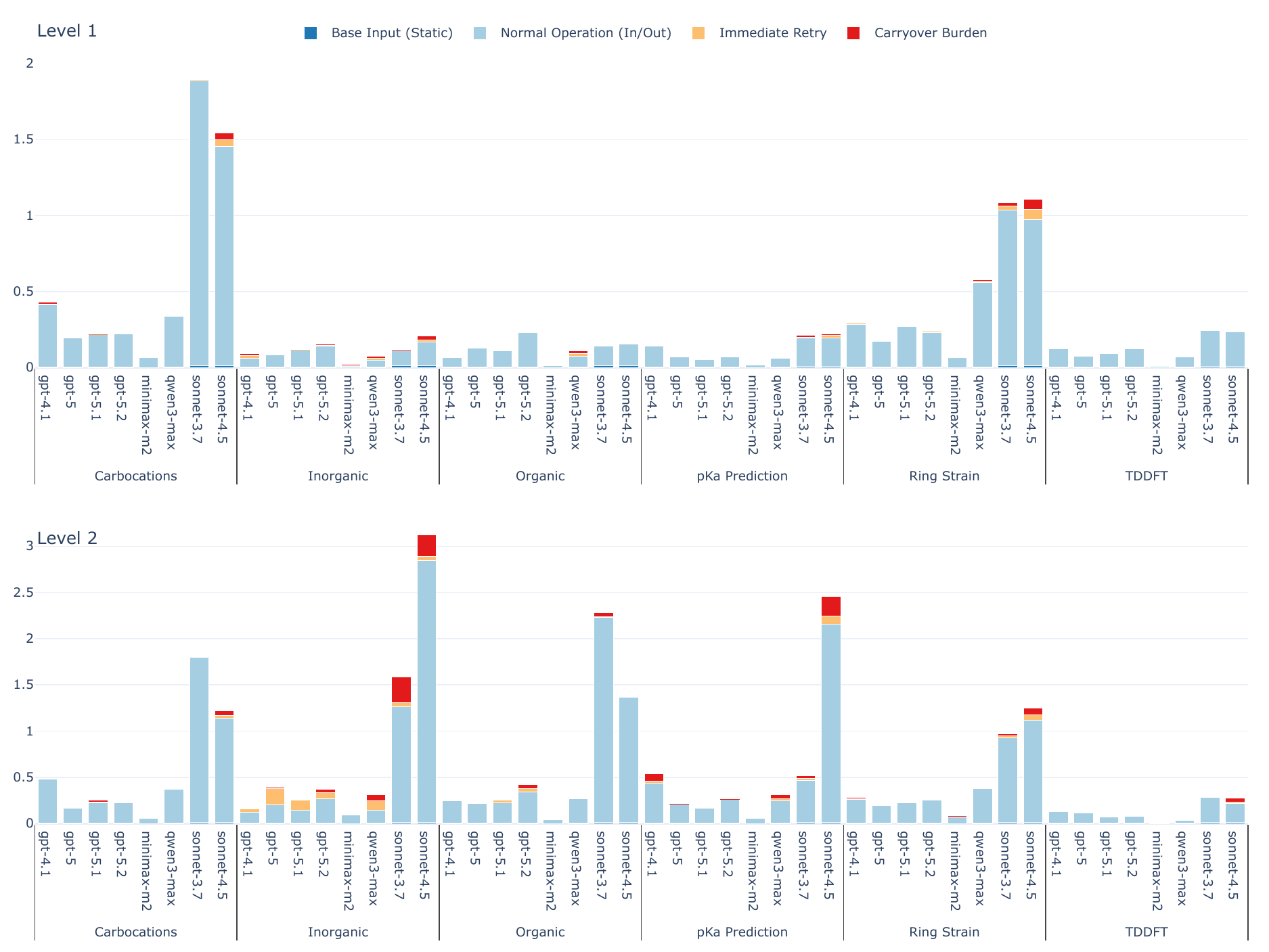}
    \caption{Same split as Fig.~\ref{fig:si_trace_token_composition}, but measured in USD cost calculated from pricing provided by LLM providers.}
    \label{fig:si_trace_usd_composition}
\end{figure}

\FloatBarrier

%% file: includes/si_tex/si-bare-agent-summary.tex
Informed by prior work on agentic systems that integrate code execution and web search, where \ac{llm} agents dynamically construct quantum simulation scripts using these tools, we examine whether a similar approach generalizes to computational chemistry benchmarks.
We set up a lightweight \ac{llm} agent with a web search tool and code execution capabilities and tested it on two representative El Agente tasks: inorganic compounds level 1 and pKa prediction level 2. The first exercise examines the \ac{llm}'s ability to write scripts to optimize the geometry, remove imaginary frequencies, obtain the point group symmetry, and obtain the IAO populations, none of which are readily accessible from PySCF with one-liners. The latter exercise examines the \ac{llm}'s ability to extract the Gibbs free energy and fit the results against the provided experimental values. 

For the inorganic compounds exercise, the agent sets up the molecules from SMILES using the OpenBabel package, performs geometry optimization with PySCF, followed by a tighter convergence single-point calculation. However, it was not able to extract all properties correctly. In particular, it made mistakes in: (1) extracting the point-group symmetry, (2) running Lowdin analysis correctly, (3) omitting frequency analysis and imaginary frequency removal, and (4) creating the correct geometry of chlorine trifluoride (producing a trigonal planar structure instead of the correct T-shaped structure). The full transcript can be found in~Sec.~\ref{si:chat_bare_agent_inorganic}.

For the pKa prediction exercise level 2, the agent was also able to run chloro-, fluoro-, and chlorofluoro-acetic acid, as well as acetic acid, successfully. Molecular geometries were first generated from SMILES and then optimized with PySCF, followed by frequency analysis to obtain the Gibbs free energy. However, to estimate the free energy of solvation, the \ac{llm} obtained energies using the COSMO implicit solvent model and subtracted them from the Gibbs free energy of the gas-phase molecules. With this inaccurate protocol, it obtained a pKa of -5.0, well outside the range considered correct in our rubric (-2.70 to 1.50). The \ac{llm} also omitted checking for imaginary frequencies after geometry optimization. The full transcript can be found in~Sec.~\ref{si:chat_bare_agent_pka}. 

These exploratory experiments used \texttt{gpt-5} and comprised one bare-agent trace per task. The inorganic-compounds and p$K_\mathrm{a}$ runs required approximately 40~min and 650k trace tokens, and 16~min and 450k trace tokens, respectively. With the same model, the corresponding \grafico{} runs required approximately 3~min and 25k tokens, and 5~min and 122k tokens. Because the bare-agent runs were not repeated, these comparisons characterize the two recorded traces rather than expected performance distributions and were excluded from the repeated benchmark statistics. The traces nevertheless illustrate how an unconstrained coding agent constructed procedures: it searched the web for the p$K_\mathrm{a}$ task and inspected the PySCF source code for the inorganic-compounds task. The resulting functions suggest a route for bootstrapping new tools, but the scientific omissions described above show why generated procedures require validation before reuse.

System prompt provided to this lightweight agent is (no additional context or instructions provided):

\begin{lstlisting}[
basicstyle=\ttfamily\normalsize,
frame=single,
breaklines=true,
breakindent=0pt,
columns=fullflexible,
xleftmargin=0.5em,
xrightmargin=0.5em
]
You're a computational chemistry agent that interprets user intent and executes quantum chemistry calculations by writing your own python functions using the following python packages: gpu4pyscf, rdkit, numpy, and openbabel. You can execute your Python code via the `Python_REPL` tool. You can use DuckDuckGo to search for code examples of gpu4pyscf on https://github.com/pyscf/gpu4pyscf/tree/master/examples. You have access to GPU, so ALWAYS use imports from gpu4pyscf to accelerate the calculation, or use `to_gpu()` method from pyscf to make sure it's accelerated on GPU. When running tools in parallel, fix error for any of the failed tools before reply back.
\end{lstlisting}

%% file: includes/si_tex/si-crest-pyscf-summary.tex
Chemical processes rarely occur on the basis of a single, static molecular structure.
Instead, a thermodynamic ensemble, e.g., one mole of molecules, comprises a large number of distinct conformations, often numbering in the millions or more~\cite{desouzaGOATGlobalOptimization2025}.
This configurational diversity goes beyond finite-temperature effects arising from the occupation of vibrational energy levels, which are typically accounted for via thermostatistical corrections~\cite{grimmeEfficientQuantumChemical2021}.
Accurately predicting spectroscopic properties in solution, therefore, requires explicit sampling of relevant molecular conformations and Boltzmann-weighted averaging over the resulting ensemble.
Beyond brute-force sampling via long and/or high-temperature \ac{md} simulations, a range of efficient algorithms for conformer search and ensemble generation has emerged in recent years~\cite{desouzaGOATGlobalOptimization2025,prachtCRESTProgramExploration2024,rdkit}.

To assess solvent effects on electronic absorption spectra under both implicit and explicit solvation models, we executed two user queries that required conformer sampling, solvation modelling, geometry refinement using DFT, and spectral analysis.
Across both use cases, the agent orchestrated specialized tools, including CREST~\cite{Steinbach2025crestGPU} for conformer discovery, \ac{qcg} for explicit solvent cluster construction~\cite{Spicher2022QCG}, and PySCF for electronic structure calculations, such as geometry optimizations, vibrational analysis, and \ac{tddft}, while managing parallel execution and enforcing methodological constraints.
Intermediate molecular structures were exchanged using knowledge-graph-backed \texttt{ConceptualAtoms} identifiers, enabling reproducible tool-to-tool handoff without repeated serialization of large coordinate payloads through the context window of \acp{llm}.

\paragraph{\textbf{Implicit-solvent ensemble}}
For a merocyanine compound, \grafico{} launched independent CREST conformer searches at the GFN2-xTB/ALPB level~\cite{bannwarthGFN2xTBAccurateBroadly2019,ehlertRobustEfficientImplicit2021} in water and \textit{n}-heptane. Five conformers from each solvent were passed by \ac{iri} to parallel $\omega$B97X-D4~(SMD)/def2-SVP geometry-refinement, frequency and \ac{tddft} workflows~\cite{mardirossianOb97XV10parameterRangeseparated2014,caldeweyherGenerallyApplicableAtomiccharge2019,weigendBalancedBasisSets2005,marenichUniversalSolvationModel2009}. Frequency analysis verified the retained minima; one \textit{n}-heptane conformer entered the graph's imaginary-mode removal and reoptimization path before the excited-state calculation.

After refinement, the agent wrote and executed Python code to deduplicate the conformers, recompute their Boltzmann populations and aggregate the \ac{tddft} outputs into solvent-specific spectra. Deduplication applied an energy-first criterion ($\Delta E \leq 0.05$~kcal~mol$^{-1}$), followed by heavy-atom \ac{rmsd} validation ($\leq 0.15$~\AA). The closest pairs remained distinct minima (water: $\Delta E=0.368$~kcal~mol$^{-1}$, \ac{rmsd}$=0.041$~\AA; \textit{n}-heptane: $\Delta E=0.447$~kcal~mol$^{-1}$, \ac{rmsd}$=0.047$~\AA). The water ensemble was distributed over four conformers with weights of 43.3, 29.9, 18.9 and 7.9\%, whereas one conformer accounted for 95.2\% of the \textit{n}-heptane ensemble. Gaussian broadening gave ensemble absorption maxima of 412.8~nm in water and 429.9~nm in \textit{n}-heptane; the corresponding dominant-conformer maxima were 412.6 and 429.6~nm. Thus, conformational averaging only weakly modified the calculated 17~nm solvent-dependent shift. The complete query ran for 46~min, used 1.3 million tokens and cost \$0.95 with \texttt{gpt-5.4} at medium reasoning effort.

\paragraph{\textbf{Explicit-solvent ensemble}}
For 2,3-epoxybutanol~\cite{beckEvaluatingImportanceConformers2024}, a gas-phase CREST search ran concurrently with construction of a 15-water solvation shell by QCG~\cite{Spicher2022QCG} and subsequent sampling in CREST's non-covalent-interaction mode. Conformers were selected to exceed 95\% of the initial Boltzmann population: four gas-phase conformers provided 96.8\% coverage and eight microsolvated conformers provided 95.6\% coverage. The gas-phase structures were refined at the $\omega$B97X-D4/def2-SVP level, whereas the microsolvated structures additionally used SMD(water), before parallel \ac{tddft} calculations.

The agent again wrote Python code to recompute populations from the refined energies and construct the ensemble spectra. The gas-phase ensemble was dominated by one conformer with 97.9\% weight, whereas two microsolvated structures contributed 63.4 and 36.6\%. Explicit water together with SMD shifted the lowest bright absorption from 9.47~eV (131.0~nm) to 9.21~eV (134.7~nm), a change of 0.26~eV (3.7~nm). In the Gaussian-broadened spectra, the ensemble maximum shifted from 121.8 to 130.9~nm and the relative intensities of the contributing transitions changed. These results are consistent with perturbation of the excited-state electronic structure by direct solute--solvent interactions. The complete query ran for 31~min, used 350,000 tokens and cost \$0.47 with \texttt{gpt-5.4} at medium reasoning effort.

%% file: includes/si_tex/si-mof-workflow.tex
This section describes the internal data flow within the \ac{mof} execution graph, detailing how structures are decomposed, represented, translated between software libraries, and passed through each computational stage. Throughout, the execution graph operates over typed state objects and optionally persists intermediate and final artifacts to the knowledge graph via a custom \ac{ogm} layer (based on \texttt{twa}~\cite{Bai2025twa}; see the infrastructure description in the main text).

\textbf{High-level node chains.}
Based on the connections (type hints) of each node, we expect the \ac{mof} execution graph to support three main entry routes. It should be noted, however, that the routing between these nodes is controlled by an \ac{llm} \ac{api} call that decides the next node based on both the current state of the execution graph and the summarized user intent. It is therefore possible that unexpected behaviour may emerge from an agent operating over the \ac{mof} execution graph, \textit{e.g.}, as observed in Supplementary~Sec.~\ref{si:mof_interactive_exploration}.
\begin{itemize}
    \item \textbf{CIF route (local or downloaded):} \texttt{StartWorkflow} $\rightarrow$ \texttt{(DownloadFromCSD)} $\rightarrow$ \texttt{ProcessCIF} $\rightarrow$ \texttt{BuildMOFBasic} $\rightarrow$ \texttt{MLFFGeomOpt} $\rightarrow$ \texttt{ZeoppAnalysis} $\rightarrow$ \texttt{MOFGraphEnd}. [\noindent\emph{Note: \texttt{DownloadFromCSD} is used only when a CSD refcode is provided; when a CIF path is provided, the execution graph enters directly at \texttt{ProcessCIF}.}]
    \item \textbf{\ac{kg}-driven route:} \texttt{StartWorkflow} $\rightarrow$ \texttt{CombinatorialSearch} $\rightarrow$ \texttt{BuildMOFBasic} $\rightarrow$ \texttt{MLFFGeomOpt} $\rightarrow$ \texttt{ZeoppAnalysis} $\rightarrow$ \texttt{MOFGraphEnd}. [\textit{Note: it has been observed that sometimes the main agent only wants to inspect the possible new \acp{mof} without constructing them, therefore it instructs the routing agent to exit after the combinatorial search.}]
    \item \textbf{PORMAKE-component route (direct building blocks):} \texttt{StartWorkflow} $\rightarrow$ \texttt{BuildMOFBasic} $\rightarrow$ \texttt{MLFFGeomOpt} $\rightarrow$ \texttt{ZeoppAnalysis} $\rightarrow$ \texttt{MOFGraphEnd}.
\end{itemize}

\subsubsection{CIF decomposition to \ac{ogm} building blocks}

When an experimental CIF enters the execution graph via the \textbf{ProcessCIF} node, it undergoes a three-step decomposition pipeline implemented in \texttt{ontomofs\_from\_cif.py}.

\textbf{Step 1: Topology classification.}
The CIF is passed to CrystalNets.jl~\cite{CrystalNets} (via \texttt{juliacall}) to determine its underlying topological net. The classification returns a topology label (\textit{e.g.}, \texttt{pcu}, \texttt{mtn}), a dimensionality, and a catenation number. The node currently restricts its scope to catenation $= 1$ with a single topology label; CIFs yielding multiple labels are rejected.

\textbf{Step 2: Structure decomposition.}
PORMAKE's~\cite{Lee2021pormake} experimental \texttt{MOFDecomposer} is invoked on the CIF to extract provisional building blocks as ASE~\cite{larsen2017atomic} \texttt{Atoms} objects. These raw fragments include both real atoms and PORMAKE ``docking points'' (pseudo-atoms with atomic number $Z = 0$) that mark connection sites.

\textbf{Step 3: Building block deduplication.}
Because the decomposer produces one fragment per crystallographic site (which may include symmetry-equivalent copies), a two-stage deduplication strategy reduces the set to unique components:
\begin{itemize}
    \item \textbf{Stage 1 (fast grouping):} apply PORMAKE's \texttt{hash\_atoms} (complexity parameter \texttt{7}) to structurally hash and group fragments.
    \item \textbf{Stage 2 (structural verification):} only when multiple candidate nodes/linkers remain after hashing, perform pairwise structural comparison using \texttt{mofid}'s~\cite{Bucior2019mofid1} interface to pymatgen's \texttt{StructureMatcher}~\cite{ong2013python} with tolerances \texttt{ltol=0.3}, \texttt{stol=2.0}, and \texttt{angle\_tol=5.0}.
\end{itemize}
The pipeline then enforces the current scope constraint: exactly one unique metal node and one unique organic linker per CIF.

\textbf{\ac{ogm} instantiation and in-process canonicalization.}
Each deduplicated building block is instantiated as an \ac{ogm} class: \texttt{MetalNode} or \texttt{OrganicLinker}, both subclasses of \texttt{BuildingBlock}. These \ac{ogm} classes wrap the underlying PORMAKE \texttt{BuildingBlock} objects and enforce semantic constraints: \texttt{MetalNode} validates \texttt{has\_metal=True} (via PORMAKE metal detection), while \texttt{OrganicLinker} validates \texttt{has\_metal=False}. The topology is similarly wrapped as a \texttt{Topology} \ac{ogm} instance.

To prevent duplicate Python objects when parallel execution graph invocations process the same component concurrently, \texttt{Topology}, \texttt{MetalNode}, and \texttt{OrganicLinker} instances are registered in an in-process canonicalization registry guarded by per-key \texttt{threading.RLock} instances. This provides \emph{within-process} identity stability under concurrency. When \ac{kg} integration is enabled, a second deduplication layer queries the external graph for pre-existing equivalents. These functions implement a two-stage strategy: a SPARQL query first filters candidates by atom count (for building blocks) or identifier (for topologies), then performs structural comparison using pymatgen's \texttt{StructureMatcher}. If an equivalent is found in the \ac{kg}, the existing instance is reused rather than creating a duplicate.

\subsubsection{\ac{ogm} instances and ConceptualAtoms}

The ontomofs \ac{ogm} classes store atomic geometry through \texttt{ConceptualAtoms}, a Pydantic model that serves as the framework's canonical validated geometry container. When a PORMAKE building block is wrapped, its ASE \texttt{Atoms} object is split into two parts:
\begin{itemize}
    \item \textbf{\texttt{BuildingBlock.atoms} (\texttt{ConceptualAtoms}):} contains only real atoms ($Z \neq 0$), with positions in \AA{}, the unit cell (if periodic), periodic boundary conditions, and electronic state (charge, spin multiplicity). Atomic numbers and positions are stored in an \texttt{XYZ} sub-model as validated NumPydantic arrays.
    \item \textbf{\texttt{BuildingBlock.docking\_points} (\texttt{XYZ}):} contains the pseudo-atoms ($Z = 0$) that mark PORMAKE connection sites.
\end{itemize}

\textbf{Round-tripping across libraries.}
\texttt{ConceptualAtoms} provides bidirectional conversion utilities:
\begin{itemize}
    \item \texttt{from\_ase\_atoms(atoms)} / \texttt{to\_ase\_atoms(...)}: round-trip to and from ASE \texttt{Atoms}, preserving charge and spin multiplicity via \texttt{atoms.info}.
    \item \texttt{from\_qcelemental\_molecule(mol)} / \texttt{to\_qcelemental\_molecule()}: round-trip to and from QCElemental \texttt{Molecule} instances.
\end{itemize}

\textbf{Identity across in-memory state and \ac{kg} persistence.}
\texttt{ConceptualAtoms} is integrated with the \ac{ogm} layer (via \texttt{GraphBaseModel}), meaning each instance carries an \texttt{instance\_iri} that uniquely identifies it across in-memory Python state and the external knowledge graph. When desired, this \ac{iri} can be preserved through ASE round-trips by embedding it in \texttt{atoms.info["instance\_iri"]} using \texttt{to\_ase\_atoms(preserve\_iri=True)}. When \texttt{from\_ase\_atoms} encounters an \ac{iri} in the info dict, it can retrieve the existing \texttt{ConceptualAtoms} from the \ac{ogm} lookup table (when available) rather than creating a new instance, ensuring object identity is maintained across serialization boundaries.

\subsubsection{{\ac{mof} combinatorial search algorithms}}
\input{includes/si_tex/si-sparql-queries}

\subsubsection{Connecting combinatorial search to PORMAKE construction}

The \textbf{CombinatorialSearch} node queries the knowledge graph via SPARQL to enumerate feasible (topology, metal node, organic linker) combinations. The SPARQL result rows contain \acp{iri} for each component. These \acp{iri} serve as stable identifiers that are resolved by the \ac{ogm} into concrete Python objects, which are then queued for PORMAKE construction.

\begin{itemize}
    \item Each \ac{iri} is resolved to a fully-hydrated \ac{ogm} instance using \texttt{pull\_from\_kg(..., recursive\_depth=-1)}, which recursively materializes \texttt{Topology}, \texttt{MetalNode}, and \texttt{OrganicLinker} objects (including any available nested geometry and motifs stored in the \ac{kg}).
    \item Retrieved components are passed through the in-process canonicalization registry (\texttt{get\_or\_register\_topology}, \texttt{get\_or\_register\_building\_block}) so repeated \acp{iri} across result rows map to the same Python objects within a run.
\end{itemize}

Finally, the canonicalized triples are enqueued on \texttt{state.build\_queue}, a lightweight queue containing topologies, node building blocks, edge building blocks, and names for construction by the downstream \textbf{BuildMOFBasic} node.

\subsubsection{Direct construction from PORMAKE building-block Identifiers}
\label{sec:si-mof-pormake-entry}

In addition to CIF and \ac{kg}-driven entry routes, the execution graph supports direct \ac{mof} construction from user-specified PORMAKE components. In this route, the user provides identifiers for a topology and building blocks (\textit{e.g.}, \texttt{bcu} with metal node \texttt{N625} and linkers \texttt{E14}/\texttt{E32}/\texttt{E34}). The execution graph bypasses CIF decomposition and proceeds directly to the construction stage.

\textbf{Identifier resolution and queueing.}
Identifiers are resolved to concrete atomistic objects by retrieving (or instantiating) the corresponding \ac{ogm} wrappers:
\begin{itemize}
    \item \textbf{Topology:} the topology identifier is mapped to a \texttt{Topology} \ac{ogm} instance (wrapping the underlying PORMAKE topology definition, including its node/edge types and local motifs).
    \item \textbf{Building blocks:} each PORMAKE building-block identifier is mapped to an \ac{ogm} \texttt{MetalNode} or \texttt{OrganicLinker}. These objects encapsulate the underlying PORMAKE \texttt{BuildingBlock} and store validated geometry in \texttt{ConceptualAtoms} (real atoms) plus a separate \texttt{XYZ} object for PORMAKE docking points ($Z=0$).
\end{itemize}

After resolution, the resulting (\texttt{Topology}, \texttt{MetalNode}, \texttt{OrganicLinker}) triples are passed through the in-process canonicalization registry and enqueued onto \texttt{state.build\_queue}, exactly as in the \ac{kg}-driven route. This design ensures that all downstream stages (\texttt{BuildMOFBasic}, \texttt{MLFFGeomOpt}, \texttt{ZeoppAnalysis}) operate on a uniform representation regardless of the entry point. Routing between these nodes is controlled by the \ac{llm}-based routing agent by checking the existing state of the execution graph and the target user intent.

\subsubsection{\ac{mof} construction via PORMAKE}

The \textbf{BuildMOFBasic} node iterates over the build queue and, for each (topology, node, linker) triple, performs:

\begin{enumerate}
    \item \textbf{Type assignment.} Unique node-type tags are read from \texttt{topology.unique\_node\_types} (integer identifiers from the CGD net). The single \texttt{MetalNode} is assigned to all node types. Similarly, the single \texttt{OrganicLinker} is assigned to all edge types from \texttt{topology.unique\_edge\_types} (pairs of node-type tags).

    \item \textbf{RMSD validation.} For each node-type assignment, \texttt{MOFBuilder.rmsd\_for\_node\_type} computes the RMSD between the building block's connection-point geometry and the topology's local coordination motif (a PORMAKE \texttt{LocalStructure}). Assignments exceeding a configurable threshold (PORMAKE default $0.3$) raise an error.

    \item \textbf{Framework assembly.} \texttt{MOFBuilder.build\_by\_type} delegates to PORMAKE's \texttt{Builder.build\_by\_type}, which handles locating building blocks onto topology sites, scaling the unit cell, and constructing the periodic framework. PORMAKE returns a \texttt{Framework} object whose \texttt{atoms} attribute is an ASE \texttt{Atoms} object with the fully assembled periodic structure.

    \item \textbf{ConstructedMOF creation.} The PORMAKE \texttt{Framework} is wrapped as a \texttt{ConstructedMOF} \ac{ogm} instance. This step converts the framework's ASE \texttt{Atoms} to \texttt{ConceptualAtoms} (via \texttt{ConceptualAtoms.from\_ase\_atoms}), generates CIF text (via a PORMAKE write-to-tempfile path), and records provenance (source topology, building blocks used, bond connectivity). The \texttt{ConstructedMOF} carries its own \texttt{instance\_iri} and stores references to \texttt{source\_topology} and \texttt{building\_blocks\_used} as \ac{ogm} object properties, preserving construction provenance for \ac{kg} persistence.
\end{enumerate}

\subsubsection{\ac{mlip} geometry optimization}

The \textbf{MLFFGeomOpt} node relaxes constructed \ac{mof} geometries using machine-learned interatomic potentials (MLIPs). The data flow proceeds as:

\begin{enumerate}
    \item \textbf{ASE Atoms extraction.} For each \texttt{ConstructedMOF}, the geometry is extracted via \texttt{cmof.atoms.to\_ase\_atoms()}, converting \texttt{ConceptualAtoms} positions (in \AA{}) and unit cell back to an ASE \texttt{Atoms} object with periodic boundary conditions.

    \item \textbf{Calculator attachment.} The \texttt{get\_mlff\_calculator} factory instantiates an ASE-compatible calculator for the selected MLIP model (MACE-MOF~\cite{batatia2022mace}, Orb~\cite{rhodes2025orb}, MatterSim~\cite{yang2024mattersim}, or MACE-OMOL). For MACE-MOF, model weights are lazily downloaded and cached. Calculator instances are cached per-thread to reduce repeated initialization and avoid GPU contention across parallel execution graph invocations.

    \item \textbf{Structure relaxation.} The \texttt{relax\_structure} utility wraps ASE optimizers (default: FIRE). When \texttt{relax\_cell=True}, a \texttt{FrechetCellFilter} is applied to allow simultaneous optimization of atomic positions and lattice parameters. A \texttt{TrajectoryObserver} records energies, forces, stresses, and atomic positions at each step. An optional \texttt{on\_update} callback streams intermediate CIF snapshots to the \texttt{GraphChat} frontend for live visualization.

    \item \textbf{Post-relaxation update (identity-preserving).} After convergence, a single-point calculation is performed on the final structure to obtain the relaxed energy. The optimized ASE \texttt{Atoms} object is written back to the \texttt{ConstructedMOF} by updating the \emph{existing} \texttt{ConceptualAtoms} instance in place, preserving the original \texttt{instance\_iri}. Updated CIF text is regenerated from the optimized geometry and stored in \texttt{cmof.cif\_text}.
\end{enumerate}

\subsubsection{Zeo++ porosity analysis}

The \textbf{ZeoppAnalysis} node computes geometric porosity descriptors on the structures included in the execution graph state. For each \texttt{ConstructedMOF}, the CIF text (stored as \texttt{cmof.cif\_text}) is written to a temporary file and passed to Zeo++~\cite{Willems2012zeopp} via the \texttt{CoREMOF}~\cite{Zhao2025coremof} Python wrapper. Five analysis types are available, each represented by an \ac{ogm} input/result pair:

\begin{itemize}
    \item \textbf{Pore diameter} (\texttt{PoreDiameterInput} $\to$ \texttt{PoreDiameterResult}): largest cavity diameter (LCD), pore limiting diameter (PLD), and largest free pore diameter (LFPD).
    \item \textbf{Surface area} (\texttt{SurfaceAreaInput} $\to$ \texttt{SurfaceAreaResult}): accessible and non-accessible surface area (m$^2$/g, m$^2$/cm$^3$).
    \item \textbf{Pore volume} (\texttt{PoreVolumeInput} $\to$ \texttt{PoreVolumeResult}): pore volume (cm$^3$/g, \AA$^3$) and void fraction.
    \item \textbf{Channel dimensionality} (\texttt{ChannelDimensionInput} $\to$ \texttt{ChannelDimensionResult}): counts of 1D, 2D, and 3D channels.
    \item \textbf{Framework dimensionality} (\texttt{FrameworkDimensionInput} $\to$ \texttt{FrameworkDimensionResult}): the dimensionality of the framework itself.
\end{itemize}

Each result is stored as an \ac{ogm} object property on the \texttt{ConstructedMOF} instance (\textit{e.g.}, \texttt{cmof.pore\_diameter\_analysis}). On failure (\textit{e.g.}, problematic CIF geometry), a \texttt{ZeoppAnalysisError} object is returned instead, capturing Zeo++ stdout/stderr, return code, and the original configuration for diagnostics by the agent.

\subsubsection{Knowledge graph persistence}

At execution graph completion, the \textbf{MOFGraphEnd} node pushes all \texttt{ConstructedMOF} instances to the knowledge graph via the \ac{ogm}'s \texttt{push\_to\_kg} method. Because components (topology, building blocks, \texttt{ConceptualAtoms} geometry, local structures, and Zeo++ results) are represented as interconnected \ac{ogm} objects with stable \acp{iri}, the full provenance chain can be serialized as RDF triples. Subsequent user sessions can query these persisted structures via SPARQL, enabling \ac{kg}-driven combinatorial search over previously processed components and frameworks.

\subsubsection{Summary of representation translations}

Table~\ref{tab:mof-translations} summarizes key representation translations that occur as data flows through the \ac{mof} execution graph.

\begin{table}[H]
\centering
\caption{Representation translations across the \ac{mof} execution graph. Each row describes a stage in the pipeline, the input and output data representations, and the mechanism used for translation.}
\label{tab:mof-translations}
\small
\setlength{\tabcolsep}{6pt}
\renewcommand{\arraystretch}{1.75}
\begin{tabular}{P{3.0cm} P{3.0cm} P{3.5cm} P{4.5cm}}
\hline
\textbf{Stage} & \textbf{Input Representation} & \textbf{Output Representation} & \textbf{Translation Mechanism} \\
\hline
CIF \pipe Topology &
CIF file path &
\texttt{Topology} (\ac{ogm}) &
CrystalNets.jl classification \pipe PORMAKE topology \pipe \texttt{Topology} \ac{ogm} wrapper \\

CIF \pipe Building Blocks &
CIF file path &
\texttt{MetalNode} / \texttt{OrganicLinker} (\ac{ogm}) &
PORMAKE \texttt{MOFDecomposer} \pipe ASE \texttt{Atoms} fragments \pipe
\texttt{deduplicate\_building\_blocks} \pipe \texttt{create\_ontomofs\_building\_block}
(splits real atoms vs docking points into \texttt{ConceptualAtoms} + \texttt{XYZ}) \\

\ac{kg} \pipe Python Objects &
SPARQL result \acp{iri} &
\texttt{Topology}, \texttt{MetalNode}, \texttt{OrganicLinker} &
\texttt{pull\_from\_kg} with recursive hydration (\texttt{recursive\_depth=-1}) \\

PORMAKE IDs \pipe \ac{ogm} components &
String identifiers (topology / node / linker) &
\texttt{Topology}, \texttt{MetalNode}, \texttt{OrganicLinker} &
Resolve identifiers to PORMAKE definitions \pipe wrap as \ac{ogm} objects
(\texttt{ConceptualAtoms} + docking points) \pipe canonicalize and enqueue in
\texttt{state.build\_queue} \\

\ac{ogm} \pipe PORMAKE &
\texttt{BuildingBlock} (\ac{ogm}) &
PORMAKE \texttt{BuildingBlock} &
Lazy property access (\textit{e.g.}, cached PORMAKE \texttt{BuildingBlock} construction inside \ac{ogm} wrapper) \\

PORMAKE \pipe ConstructedMOF &
PORMAKE \texttt{Framework} &
\texttt{ConstructedMOF} (\ac{ogm}) &
\texttt{from\_pormake\_framework} \pipe \texttt{ConceptualAtoms.from\_ase\_atoms} \pipe provenance wiring \\

\ac{ogm} \pipe ASE (for MLIP) &
\texttt{ConceptualAtoms} &
ASE \texttt{Atoms} &
\texttt{to\_ase\_atoms()}
(optionally \texttt{preserve\_iri=True} to embed \ac{iri} in \texttt{atoms.info} for later geometry updates) \\

ASE \pipe \ac{ogm} (post-MLIP) &
ASE \texttt{Atoms} &
\texttt{ConceptualAtoms} (updated) &
In-place update of existing \texttt{ConceptualAtoms} from relaxed ASE geometry
(preserves original \texttt{instance\_iri}) \\

\ac{ogm} \pipe CIF (for Zeo++) &
\texttt{ConstructedMOF.cif\_text} &
Temporary CIF file &
Write CIF text to \texttt{tempfile} for Zeo++ execution \\

\ac{ogm} \pipe \ac{kg} &
All \ac{ogm} instances &
RDF triples &
\texttt{push\_to\_kg} via \ac{ogm} serialization \\
\hline
\end{tabular}
\end{table}

%% file: includes/si_tex/si-sparql-queries.tex
Query~\ref{lst:si:sparql:alg1} presents a SPARQL query to propose candidate \acp{mof} from historical synthesis data, inspired by the similar algorithm implemented for metal-organic polyhedra~\cite{kondinski2022automated,Bai2025twa}. The underlying hypothesis is that chemical building blocks (metal nodes and organic linkers) that have successfully crystallized into a specific topological net in prior experiments possess inherent chemical or steric compatibilities with that topology. For a target topology $T$, the algorithm retrieves the set of metal nodes $M_{\text{proven}} = \{ m \mid \exists (m, T) \in \mathcal{K}_{\text{obs}}\}$ and organic linker $L_{\text{proven}} = \{ l \mid \exists (l, T) \in \mathcal{K}_{\text{obs}}\}$, where $\mathcal{K}_{\text{obs}}$ represents the set of experimentally observed \acp{mof} instantiated in the graph. A Cartesian product $P=M_{\text{proven}} \times L_{\text{proven}}$ is computed and subsequently filtered for pairs $(m,l)$ where the metal node and linker represent different local structural roles to satisfy topological constraints. The possible combinations are then filtered by negation to remove any pair $(m,l,T)$ that already exists as a \texttt{mofs:ConstructedMOF} in the graph. The query can be configured with \texttt{VALUES} clause to restrict the proposed candidate to only include certain topologies or building blocks.

\begin{lstlisting}[language=SPARQL, caption={Intra topology search}, label={lst:si:sparql:alg1}]
# Algorithm 1: find new MOF combinations using topologies that have already
# succeeded with each component (metal node and organic linker) individually.
#
# Placeholder markers (`# {{VALUES_*}}`) are replaced programmatically by
# `sparql_alg_with_values` in `sparql_utilities.py` to inject optional VALUES clauses.

PREFIX mofs: <https://elagente.ca/ontomof/>
PREFIX rdf: <http://www.w3.org/1999/02/22-rdf-syntax-ns#>
PREFIX rdfs: <http://www.w3.org/2000/01/rdf-schema#>
PREFIX grafico: <https://elagente.ca/grafico/>

SELECT DISTINCT ?predicted_mof_name ?topology ?metal_node ?organic_linker
WHERE {
  # ---------------------------------------------------------------
  # 1. Find metal nodes that have already succeeded on a topology
  # --------------------------------------------------------------
  {
      SELECT DISTINCT ?metal_node ?topology ?metal_local_structure
      WHERE {
        # {{VALUES_TOPOLOGY}}
        # {{VALUES_METAL}}
        ?metal_node a mofs:MetalNode ;
            mofs:functions_as ?metal_local_structure.

        # This metal node was used in an existing MOF on a topology
        ?metal_node ^mofs:building_blocks_used ?existing_mof.
        ?existing_mof mofs:source_topology ?topology.

        ?metal_local_structure ^mofs:local_structures ?topology.
      }
  }

  # ------------------------------------------------------------------------
  # 2. Find organic linkers that have already succeeded on the same topology
  # ------------------------------------------------------------------------
  {
      SELECT DISTINCT ?organic_linker ?topology ?linker_local_structure
      WHERE {
        # {{VALUES_TOPOLOGY}}
        # {{VALUES_LINKER}}
        ?organic_linker a mofs:OrganicLinker ;
                        mofs:functions_as ?linker_local_structure.

        # This linker was used in an existing MOF with the same topology
        ?organic_linker ^mofs:building_blocks_used ?existing_mof.
        ?existing_mof mofs:source_topology ?topology.

        ?linker_local_structure ^mofs:local_structures ?topology.
     }
  }

  # ------------------------------------------------------------------
  # 3. Ensure they fill different local-structure roles (node vs edge)
  # ------------------------------------------------------------------
  FILTER (?metal_local_structure != ?linker_local_structure)

  # ---------------------------------------------------------------
  # 4. Ensure this specific pair hasn't been combined yet
  # ---------------------------------------------------------------
  FILTER NOT EXISTS {
    ?_mof a mofs:ConstructedMOF ;
          mofs:source_topology ?topology ;
          mofs:building_blocks_used ?metal_node, ?organic_linker .
  }

  # ---------------------------------------------------------------
  # 5. Human-friendly naming with graceful fallbacks
  # ---------------------------------------------------------------

  # Prefer chemical formulas
  OPTIONAL { ?metal_node mofs:atoms ?metal_atoms .
             ?metal_atoms grafico:chemical_formula ?node_formula . }
  OPTIONAL { ?organic_linker mofs:atoms ?linker_atoms .
             ?linker_atoms grafico:chemical_formula ?linker_formula . }
  # Fallback to names
  OPTIONAL { ?topology mofs:name ?topo_name . }
  OPTIONAL { ?metal_node mofs:name ?node_name . }
  OPTIONAL { ?organic_linker mofs:name ?linker_name . }
  # Fallback to last iri fragment
  BIND (COALESCE(?topo_name, REPLACE(STR(?topology), "^.*/", "")) AS ?topo_label)
  BIND (COALESCE(?node_formula, ?node_name, REPLACE(STR(?metal_node), "^.*/", "")) AS ?node_label)
  BIND (COALESCE(?linker_formula, ?linker_name, REPLACE(STR(?organic_linker), "^.*/", "")) AS ?linker_label)

  # Final predicted name uses formulas when available
  BIND (CONCAT(?topo_label, "_", ?node_label, "_", ?linker_label) AS ?predicted_mof_name)
}
ORDER BY ?topology ?predicted_mof_name
\end{lstlisting}

Query~\ref{lst:si:sparql:alg2} proposes \acp{mof} candidates by matching building blocks to the geometric roles (\textit{local structures} as defined in Ref.~\cite{Lee2021pormake}) of a target topology $T$, inspired by the similar algorithm implemented for metal-organic polyhedra~\cite{kondinski2022automated,Bai2025twa}. The SPARQL property path expression \texttt{(\^{}mofs:functions\_as/mofs:functions\_as)*} finds geometrically isomorphic (from the local structure perspective) building blocks using transitive closure. This enables the discovery of novel candidates solely based on geometric fit rather than observed compatibility, even those building block-topology pairs that have not previously been annotated in the graph as \texttt{compatible\_with}. Same as Query~\ref{lst:si:sparql:alg1}, candidates are filtered for distinctness and can be constrained via a \texttt{VALUES} clause.

\begin{lstlisting}[language=SPARQL, caption={Cross topology search}, label={lst:si:sparql:alg2}]
# Algorithm 2: Propose (metal, linker) pairs for a target topology by matching required LocalStructure roles,
# without requiring prior builds on that topology.
# Interpretation of "occurs on a topology": a role may match either the exact LocalStructure
#       listed by the topology or any role equivalent to it via the closure
#       (^mofs:functions_as / mofs:functions_as)*.
# Method: Choose a MetalNode and an OrganicLinker such that each can function_as a role
#         (or an equivalent role via the closure) that the topology requires; require the two
#         roles to be different (node vs edge); exclude pairs already built on that topology.
#
# Placeholder markers (`# {{VALUES_*}}`) are replaced programmatically by
# `sparql_alg_with_values` in `sparql_utilities.py` to inject optional VALUES clauses.

PREFIX mofs: <https://elagente.ca/ontomof/>
PREFIX rdf: <http://www.w3.org/1999/02/22-rdf-syntax-ns#>
PREFIX rdfs: <http://www.w3.org/2000/01/rdf-schema#>
PREFIX grafico: <https://elagente.ca/grafico/>

SELECT DISTINCT ?predicted_mof_name ?topology ?metal_node ?organic_linker
WHERE {
  # ---------------------------------------------------------------
  # 1. Metals: gather (metal_node, topology, metal_local_structure)
  # ---------------------------------------------------------------
  {
      SELECT DISTINCT ?metal_node ?topology ?metal_local_structure
      WHERE {
        # {{VALUES_TOPOLOGY}}
        # {{VALUES_METAL}}
        ?metal_node a mofs:MetalNode ;
            mofs:functions_as ?_metal_local_structure.
        # Walk the equivalence/closure of local-structure roles:
        # (^functions_as / functions_as)* allows zero-or-more repetitions of:
        #   LocalStructure <-functions_as- BuildingBlock -functions_as-> LocalStructure
        # which groups "equivalent" or "compatible" local-structure roles for substitution.
        ?_metal_local_structure (^mofs:functions_as/mofs:functions_as)* ?metal_local_structure.

        ?metal_local_structure ^mofs:local_structures ?topology.
      }
  }

  # ------------------------------------------------------------------------
  # 2. Linkers: gather (organic_linker, topology, linker_local_structure)
  # ------------------------------------------------------------------------
  {
      SELECT DISTINCT ?organic_linker ?topology ?linker_local_structure
      WHERE {
        # {{VALUES_TOPOLOGY}}
        # {{VALUES_LINKER}}
        ?organic_linker a mofs:OrganicLinker ;
            mofs:functions_as ?_linker_local_structure.
        # Same closure over local-structure roles as above (see note there)
        ?_linker_local_structure (^mofs:functions_as/mofs:functions_as)* ?linker_local_structure.

        ?linker_local_structure ^mofs:local_structures ?topology.
     }
  }
  # ------------------------------------------------------------------
  # 3. Ensure they fill different local-structure roles on that topology
  #     (prevents node-node or edge-edge collisions in the same slot)
  # ------------------------------------------------------------------
  FILTER (?metal_local_structure != ?linker_local_structure)

  # ---------------------------------------------------------------
  # 4. Ensure this specific (topology, metal, linker) combo is new
  # ---------------------------------------------------------------
  FILTER NOT EXISTS {
    ?_mof a mofs:ConstructedMOF ;
          mofs:source_topology ?topology ;
          mofs:building_blocks_used ?metal_node, ?organic_linker .
  }

  # ---------------------------------------------------------------
  # 5. Human-friendly naming with graceful fallbacks
  # ---------------------------------------------------------------

  # Prefer chemical formulas
  OPTIONAL { ?metal_node mofs:atoms ?metal_atoms .
             ?metal_atoms grafico:chemical_formula ?node_formula . }
  OPTIONAL { ?organic_linker mofs:atoms ?linker_atoms .
             ?linker_atoms grafico:chemical_formula ?linker_formula . }
  # Fallback to names
  OPTIONAL { ?topology mofs:name ?topo_name . }
  OPTIONAL { ?metal_node mofs:name ?node_name . }
  OPTIONAL { ?organic_linker mofs:name ?linker_name . }
  # Fallback to last iri fragment
  BIND (COALESCE(?topo_name, REPLACE(STR(?topology), "^.*/", "")) AS ?topo_label)
  BIND (COALESCE(?node_formula, ?node_name, REPLACE(STR(?metal_node), "^.*/", "")) AS ?node_label)
  BIND (COALESCE(?linker_formula, ?linker_name, REPLACE(STR(?organic_linker), "^.*/", "")) AS ?linker_label)

  # Final predicted name uses formulas when available
  BIND (CONCAT(?topo_label, "_", ?node_label, "_", ?linker_label) AS ?predicted_mof_name)
}
ORDER BY ?topology ?predicted_mof_name
\end{lstlisting}

%% file: includes/si_tex/si-mof-summary.tex
When a network error interrupted one workflow, the agent recovered by isolating and re-running the failed job, ensuring the full pipeline completed without data loss.

%% file: includes/si_tex/si-redox-short.tex
\prompt{
The cathodic and anodic peaks are very different in a disulfide system. Please simulate diphenyl disulfide in DMF and explain the peak separation, focusing on a redox cycle where the molecule is reduced to a −1 state and then the reduced state re-oxidized back to the neutral state. Consider all possible electrochemical and chemical reactions. Report the energy and the standard potential in V vs. Fc/Fc(\textasciicircum{}+). Use 5 eV as the absolute electrode potential of Fc. Use a cheaper functional/basis-set for fast results!
}

\graficoreslong{

\begin{tcolorbox}[fontupper=\footnotesize, colback=green!3!white, colframe=blue!5!white, boxrule=0.5pt]
\faLightbulb\ \underline{\textbf{Thoughts}}

\textbf{Considering formatting options}

I’m thinking about how to denote PhS− and PhS• in my work. A potential approach could be to create a table that includes G\_tot (Eh), reaction ΔG (in eV or kJ/mol), and E° vs. Fc. Organizing the information this way might make it more visually appealing and easier to understand. I wonder if there are any other formats I should consider to effectively present these details? It feels like there’s much to think about!
\end{tcolorbox}

\begin{tcolorbox}[fontupper=\footnotesize, colback=green!3!white, colframe=blue!5!white, boxrule=0.5pt]
\faLightbulb\ \underline{\textbf{Thoughts}}

\textbf{Considering computational methods}

I'm thinking I should clarify the differences between CE and ECE regarding fast and slow scans. In fast scans, some A− survives, leading to a weak return wave near E1, while the slower scan shows cleavage and reduction, with the return wave dominated by thiolate oxidation and dimerization. It's important to state that formal potentials derive from reaction free energies, not peak positions, and that actual peaks depend on various factors. I'll mention the S–S bond changes, like the neutral bond length and radical anion length, since it's linked to the added electron. Lastly, I’ll detail the computational method used: B3LYP/def2-SVP with SMD(DMF).
\end{tcolorbox}

\begin{tcolorbox}[fontupper=\footnotesize, colback=green!3!white, colframe=blue!5!white, boxrule=0.5pt]
\faLightbulb\ \underline{\textbf{Thoughts}}

\textbf{Evaluating electrochemical steps}

I think I should mention the "standard potential in V vs Fc/Fc+" for each electrochemical step. It might also make sense to report both the absolute values and the values vs Fc. Since the user specifically asked for the 5 eV absolute electrode potential of Fc, I’ll include that as well. This way, I’m covering all the bases and providing complete information!
\end{tcolorbox}

\begin{tcolorbox}[fontupper=\footnotesize, colback=green!3!white, colframe=blue!5!white, boxrule=0.5pt]
\faLightbulb\ \underline{\textbf{Thoughts}}

\textbf{Evaluating redox potentials}

I’m calculating the reduction potentials for some compounds. I have the values for PhSSPh and PhS, along with uncertainties to consider—about ±0.2–0.4 V. The user wants quick results, so I plan to present my findings concisely in a table format. I’ll note that all optimized structures had 0 imaginary frequencies and interpret results carefully. For the oxidation of PhS−, the equivalent reduction potential is -0.87 V vs Fc. I’ll address the formal-potential separation too.
\end{tcolorbox}

\begin{tcolorbox}[fontupper=\footnotesize, colback=green!3!white, colframe=blue!5!white, boxrule=0.5pt]
\faLightbulb\ \underline{\textbf{Thoughts}}

\textbf{Detailing reduction mechanism}

I’m considering including the S–S distance change upon reduction, going from 2.12 Å to 2.91 Å. It's essential to mention that the thiyl radical formed after cleavage is easier to reduce than the parent disulfide. This suggests the ECE route to generate two PhS− species is feasible, with an average 2e potential around -1.42 V. I’ll summarize the mechanism with clear steps for both the cathodic and anodic processes, emphasizing the oxidation of thiolates rather than a straightforward reverse reduction.
\end{tcolorbox}

\begin{tcolorbox}[fontupper=\footnotesize, colback=green!3!white, colframe=blue!5!white, boxrule=0.5pt]
\faLightbulb\ \underline{\textbf{Thoughts}}

\textbf{Evaluating chemical reactions}

I'm observing a smaller return wave near E1, noting that if it's slower, the chemical follow-up may dominate, causing a shift in the return wave. There are potential caveats to consider, like the use of a continuum solvent only, no explicit supporting electrolyte, and no transition state for cleavage, plus it's a cheaper level of analysis. I recognize that formal potentials aren't necessarily exact peak positions. Alright, let's move forward with the final production.
\end{tcolorbox}

I ran a fast DFT screen for diphenyl disulfide in DMF using:

\providecommand{\tightlist}{%
  \setlength{\itemsep}{0pt}\setlength{\parskip}{0pt}}

\begin{itemize}
\tightlist
\item B3LYP/def2-SVP
\item SMD implicit solvent = DMF
\item geometry optimization + frequency for each species
\item unrestricted open-shell treatment where needed
\end{itemize}

All optimized structures had 0 imaginary frequencies.

\paragraph{\textbf{Key computed species free energies}}
Gibbs free energies at this level:

\begin{itemize}
\tightlist
\item \textbf{PhSSPh}: \textbf{-1259.000553 Eh}
\item \textbf{[PhSSPh]•−}: \textbf{-1259.118593 Eh}
\item \textbf{PhS−}: \textbf{-629.631695 Eh}
\item \textbf{PhS•}: \textbf{-629.480054 Eh}
\end{itemize}

A useful structural indicator:
\begin{itemize}
\tightlist
\item \textbf{S–S bond in neutral PhSSPh:} \textasciitilde{}\textbf{2.12 Å}
\item \textbf{S–S bond in [PhSSPh]•−:} \textasciitilde{}\textbf{2.91 Å}
\end{itemize}

So the one-electron-reduced state is extremely S–S weakened.

---

\paragraph{\textbf{Redox thermochemistry and standard potentials}}
Using
\[
E^\circ_{\mathrm{abs}} = -\Delta G / e
\]
with \(\Delta G\) in eV for the reduction reaction, and then
\[
E^\circ(\mathrm{vs\ Fc/Fc^+}) = E^\circ_{\mathrm{abs}} - 5.00\ \mathrm{V}
\]

\paragraph{\textbf{1) First reduction}}
\[
\mathrm{PhSSPh + e^- \rightarrow [PhSSPh]^{\bullet -}}
\]

\begin{itemize}
\tightlist
\item \(\Delta G\textasciicircum{}\circ = -3.212\) eV
\item \textbf{\(E\textasciicircum{}\circ = -1.788\) V vs Fc/Fc\(\textasciicircum{}+\)}
\end{itemize}

This is the cathodic electron-transfer step.

\paragraph{\textbf{2) Radical-anion cleavage}}
\[
\mathrm{[PhSSPh]^{\bullet -} \rightarrow PhS^- + PhS^{\bullet}}
\]

\begin{itemize}
\tightlist
\item \(\Delta G\textasciicircum{}\circ = +0.186\) eV = \textbf{+18.0 kJ/mol}
\end{itemize}

So at this cheap level, full cleavage is \textbf{slightly uphill}, but only mildly so. Combined with the huge S–S elongation, the reduced state is clearly very close to dissociation.

\paragraph{\textbf{3) Thiyl/thiolate redox couple}}
\[
\mathrm{PhS^{\bullet} + e^- \rightarrow PhS^-}
\]

\begin{itemize}
\tightlist
\item \(\Delta G\textasciicircum{}\circ = -4.126\) eV
\item \textbf{\(E\textasciicircum{}\circ = -0.874\) V vs Fc/Fc\(\textasciicircum{}+\)}
\end{itemize}

Equivalently, on the reverse scan the oxidation
\[
\mathrm{PhS^- \rightarrow PhS^{\bullet} + e^-}
\]
is the anodic process associated with this couple.

\paragraph{\textbf{4) Radical dimerization}}
\[
\mathrm{2\,PhS^{\bullet} \rightarrow PhSSPh}
\]

\begin{itemize}
\tightlist
\item \(\Delta G\textasciicircum{}\circ = -1.101\) eV = \textbf{-106.2 kJ/mol}
\end{itemize}

This is strongly favorable.

---

\paragraph{\textbf{Why the cathodic and anodic peaks separate so much}}
The important point is that the return wave is \textbf{not} forced to be the exact reverse of the first electron transfer.

\paragraph{\textbf{If the process were simple and reversible}}
You would expect:
\[
\mathrm{PhSSPh + e^- \rightleftharpoons [PhSSPh]^{\bullet -}}
\]
with the anodic return near the same formal potential as the cathodic step.

\paragraph{\textbf{But the reduced state reorganizes dramatically}}
The computed S–S bond goes from \textbf{2.12 Å} to \textbf{2.91 Å} after reduction. That means the added electron strongly occupies an antibonding orbital and makes the radical anion chemically labile.

\paragraph{\textbf{This opens an EC/ECE mechanism}}
A plausible network is:

1. \textbf{Electrochemical reduction}
   \[
   \mathrm{PhSSPh + e^- \rightarrow [PhSSPh]^{\bullet -}}
   \]

2. \textbf{Chemical S–S cleavage}
   \[
   \mathrm{[PhSSPh]^{\bullet -} \rightleftharpoons PhS^- + PhS^{\bullet}}
   \]

3. At the negative cathodic potential, the formed \textbf{PhS•} is actually \textbf{easier to reduce} than the parent disulfide:
   \[
   \mathrm{PhS^{\bullet} + e^- \rightarrow PhS^-}
   \quad E^\circ = -0.874\ \mathrm{V\ vs\ Fc}
   \]
   So if cleavage occurs, an \textbf{additional electron transfer} is thermodynamically accessible.

4. On the reverse sweep, the system may oxidize \textbf{thiolate}, not the original intact radical anion:
   \[
   \mathrm{PhS^- \rightarrow PhS^{\bullet} + e^-}
   \]

5. Then the radicals rapidly recombine:
   \[
   \mathrm{2\,PhS^{\bullet} \rightarrow PhSSPh}
   \]

---

\paragraph{\textbf{Bottom-line interpretation}}
The large peak separation is explained by \textbf{follow-up chemistry after the first reduction}:

\begin{itemize}
\tightlist
\item The first cathodic event is reduction of \textbf{PhSSPh} at about \textbf{-1.79 V vs Fc/Fc\(\textasciicircum{}+\)}.
\item The reduced state is highly distorted and close to dissociation.
\item Once cleavage occurs, the reverse anodic wave is dominated by \textbf{oxidation of thiolate-derived species}, associated with the \textbf{PhS•/PhS−} couple near \textbf{-0.87 V vs Fc/Fc\(\textasciicircum{}+\)}.
\item That gives a predicted formal-potential difference of about:
\end{itemize}

\[
|-0.874 - (-1.788)| \approx 0.91\ \mathrm{V}
\]

So the anodic peak can be far from the cathodic one because the system is no longer oxidizing the same species it reduced.

---

\paragraph{\textbf{Recommended redox-cycle picture}}
For the “reduce to −1, then re-oxidize to neutral” scenario, the most relevant cycle is:

\paragraph{\textbf{Direct reversible path}}
\[
\mathrm{PhSSPh + e^- \rightleftharpoons [PhSSPh]^{\bullet -}}
\]
\begin{itemize}
\tightlist
\item formal potential: \textbf{-1.788 V vs Fc/Fc\(\textasciicircum{}+\)}
\end{itemize}

\paragraph{\textbf{Competing chemically coupled path}}
\[
\mathrm{[PhSSPh]^{\bullet -} \rightleftharpoons PhS^- + PhS^{\bullet}}
\]
followed on the reverse scan by
\[
\mathrm{PhS^- \rightarrow PhS^{\bullet} + e^-}
\]
and
\[
\mathrm{2\,PhS^{\bullet} \rightarrow PhSSPh}
\]

This second route is the best explanation for very different cathodic and anodic peaks.

---

\paragraph{\textbf{Caveat}}
Because you asked for a fast calculation, this is a \textbf{cheap screening model}. Expect absolute potentials to carry maybe a few tenths of a volt uncertainty. Also, explicit electrolyte/counterion effects in DMF can further stabilize thiolate and make cleavage more favorable than this continuum-only model suggests.

If you want, I can next:
1. summarize this as a compact reaction scheme for a paper/report, or  
2. rerun it at a slightly better level for more reliable potentials.
}

\begin{figure}[H]
    \centering
    \includegraphics[width=0.75\linewidth]{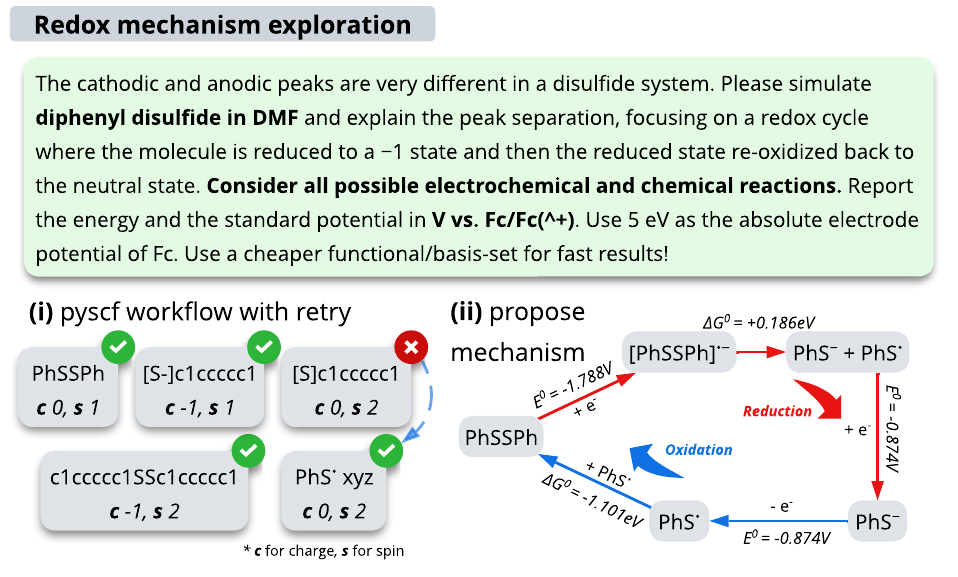}
    \caption{Redox mechanism exploration for diphenyl disulfide (PhSSPh) in DMF. (i) \grafico{} starts PySCF workflows to evaluate relevant charge and spin states for PhSSPh, its radical anion, thiolate, and thiyl radical species. The failed calculation for the \(\mathrm{PhS^{\bullet}}\) was retried with xyz coordinates generated directly by \grafico{}. (ii) Proposed electrochemical redox network explaining the large separation between cathodic and anodic peaks. Computed free energies and standard potentials versus Fc/Fc+ are shown for the key elementary steps.}
    \label{si_fig:redox}
\end{figure}

%% file: includes/si_tex/si-luca-short.tex
\prompt{
I want to test Luca Turins theory of smell! Please tell me why thiol doesn't smell like ethanol use vibrational analysis.
}

\providecommand{\tightlist}{%
  \setlength{\itemsep}{0pt}\setlength{\parskip}{0pt}}

\graficoreslong{
\begin{tcolorbox}[fontupper=\footnotesize, colback=green!3!white, colframe=blue!5!white, boxrule=0.5pt]
\faLightbulb\ \underline{\textbf{Thoughts}}

\textbf{Analyzing thiol and ethanol}

I see the user is asking why thiol doesn't smell like ethanol and wants a discussion based on vibrational analysis. I think it's important to mention Turin theory and how it compares frequencies, though it's controversial. I'll clarify that "thiol" could refer to ethanethiol, which is what I’m interpreting. I’ve already computed vibrational frequencies, noting key differences, such as the significant SH stretch at 2764 cm\textasciicircum{}-1 for ethanethiol, compared to the OH stretch at 3879 cm\textasciicircum{}-1 for ethanol.
\end{tcolorbox}

\begin{tcolorbox}[fontupper=\footnotesize, colback=green!3!white, colframe=blue!5!white, boxrule=0.5pt]
\faLightbulb\ \underline{\textbf{Thoughts}}

\textbf{Exploring vibrational theory and odor}

I'm examining Turin's vibrational theory, where smell receptors respond to inelastic electron tunneling frequencies. This means that sulfur-containing molecules have unique vibrational bands, like S-H and C-S stretches, causing them to have distinct odors compared to alcohols. For instance, ethanol has a mild alcoholic smell, while ethanethiol's odor is strong and skunky due to sulfur modes. I'll note that both optimized minima showed no imaginary frequencies and mention how reduced mass and force constants contribute to these lower frequencies. Although calculating reduced masses could be interesting, it may not be necessary right now.
\end{tcolorbox}

\begin{tcolorbox}[fontupper=\footnotesize, colback=green!3!white, colframe=blue!5!white, boxrule=0.5pt]
\faLightbulb\ \underline{\textbf{Thoughts}}

\textbf{Explaining vibrational analysis in odor differences}

I need to frame this with respect to Turin's theory, where receptors "hear" different vibrational bands. However, it's also critical to consider conventional receptor shapes and chemistry, especially since sulfur compounds interact uniquely due to their high polarizability. To directly address why thiols smell different from ethanol, I could say it's because replacing the O in ethanol with S significantly alters the vibrational fingerprint. I plan to mention the gas phase DFT wB97X-D4/def2-TZVP calculations, noting the frequencies are approximate and unscaled, which is an essential caveat.
\end{tcolorbox}

I interpreted “thiol” here as \textbf{ethanethiol} (ethyl mercaptan), the sulfur analogue of \textbf{ethanol}.

Using DFT vibrational analysis on both molecules  
(\textbf{wB97X-D4/def2-TZVP}, optimized structures, both with \textbf{0 imaginary frequencies}), the short answer is:

\textbf{Under Turin’s vibrational theory, ethanethiol should not smell like ethanol because their vibrational fingerprints are very different, especially in the modes involving O vs S.}

\textbf{Key computed differences}

\begin{tabular}{lll}
Molecule & Distinctive high-frequency stretch & Other characteristic modes \\
\hline
\textbf{Ethanol} & \textbf{O–H stretch: 3879 cm⁻¹} & C–O / C–C–O region around \textbf{1078, 1114, 1156 cm⁻¹} \\
\textbf{Ethanethiol} & \textbf{S–H stretch: 2764 cm⁻¹} & C–S / C–C–S related modes much lower, e.g. \textbf{714, 807, 889 cm⁻¹} \\
\end{tabular}

\paragraph{\textbf{Why this matters in Turin’s framework}}
Turin’s idea is that odor receptors may respond not only to molecular shape, but also to \textbf{vibrational energies} via inelastic electron tunneling. If so, replacing O by S changes the “notes” the molecule can play:

\begin{itemize}
\tightlist
\item \textbf{O–H stretch} is very high frequency: \textasciitilde{}\textbf{3880 cm⁻¹}
\item \textbf{S–H stretch} is much lower: \textasciitilde{}\textbf{2760 cm⁻¹}
\item \textbf{C–O modes} sit around \textbf{\textasciitilde{}1100 cm⁻¹}
\item \textbf{C–S modes} shift downward into the \textbf{\textasciitilde{}700–900 cm⁻¹} region
\end{itemize}

So ethanol and ethanethiol do \textbf{not} present the same vibrational spectrum to a receptor.

\paragraph{\textbf{Physical reason for the shift}}
Vibrational frequency roughly follows:

\[
\nu \propto \sqrt{\frac{k}{\mu}}
\]

where:
\begin{itemize}
\tightlist
\item \(k\) = bond force constant
\item \(\mu\) = reduced mass
\end{itemize}

Sulfur is much heavier than oxygen, and S–H / C–S bonds are generally different in stiffness from O–H / C–O bonds. That drives the sulfur-containing vibrations to \textbf{lower frequencies}.

\textbf{What this means for smell} If Turin were right in a strong sense, then:

\begin{itemize}
\tightlist
\item \textbf{ethanol} would activate receptors consistent with an alcohol-like vibrational signature
\item \textbf{ethanethiol} would activate receptors consistent with a sulfur-compound signature
\end{itemize}

That gives a plausible vibrational explanation for why:
\begin{itemize}
\tightlist
\item \textbf{ethanol} smells mild/alcoholic
\item \textbf{ethanethiol} smells intensely sulfurous/skunky
\end{itemize}

\textbf{Important caveat} This is a good \textbf{Turin-style argument}, but it is \textbf{not definitive proof} of the vibrational theory. Mainstream olfaction models still emphasize:
\begin{itemize}
\tightlist
\item receptor shape/fit
\item electrostatics
\item binding interactions
\item receptor-specific chemistry
\end{itemize}

So the safest statement is:

\textbf{Vibrational analysis shows ethanol and ethanethiol have markedly different spectra, especially in O–H vs S–H and C–O vs C–S modes. That supports why Turin’s theory would predict different smells, even though the theory itself remains controversial.}

If you want, I can also turn these frequencies into a simple \textbf{side-by-side IR-style comparison} and highlight which bands Turin would likely consider most odor-relevant.
}

%% file: includes/si_tex/si-skills-mof-mining.tex
\providecommand{\tightlist}{%
  \setlength{\itemsep}{0pt}\setlength{\parskip}{0pt}}

\subsubsection{Task specification and trace scope}

The initial \grafico{} prompt was:

\prompt{
I will provide five open-access MOF synthesis papers in the example folder, and your task is to build a generalisable execution graph that automatically mines each paper for the synthesized MOF, key synthesis conditions, reported BET surface area, and associated CCDC/CSD structure information; retrieves and prepares the corresponding structure files, cleaning solvents and converting them to P1 if needed; calculates surface area, pore volume, and channel dimensionality using Zeo++ (use bash commend directly); and compares the calculated values with the text-mined experimental BET values. The graph should separate the workflow into clear generic steps such as paper parsing, information extraction, structure retrieval/preparation, Zeo++ calculation, validation, and result visualization. The final output should include the execution graph script, a structured results table, a comparison plot, and a brief note on missing or mismatched values between calculations and mined data, explaining discrepancies and listing your assumptions. The example papers are in the working directory. The built graph should be generalisable to other papers, please run your graph on the test folder after finishing building the graph on the five example papers, but do not look at the papers in the test folder before you finished development as those papers should be used as a held-out set.
}

Claude Code received the same scientific requirements. Both prompts subsequently required an execution graph with the same stages and outputs. Both coding agents loaded \texttt{pydantic-graph-builder-scientific-workflow} and \texttt{coremof-curation}. The former specified typed dataflow, model-mediated semantic nodes, deterministic scientific nodes, fan-out and joins, terminal failure handling, tests and file-backed provenance. The latter required preservation of the input CIF, generation and checking of free-solvent-removed (FSR) and all-solvent-removed (ASR) candidates, comparison with the reported activation conditions, and selection of the least aggressive chemically valid candidate~\cite{Zhao2025coremof}.

Five open-access articles~\cite{Evans2022,Xia2024,Li2023,Gu2023,Booth2025} were supplied for development and three articles~\cite{Wang2018,Lenzen2019,Cho2020} were reserved from development. The corresponding deposited structures were downloaded and provided locally as \texttt{<ccdc\_number>.cif} without changes; neither trace performed a live CCDC retrieval. The first paper-content access to the reserved folder occurred after each agent had produced its development-stage graph and tests. The study comprises one authoring trace for each coding-agent stack.

\begin{table}[!ht]
    \centering
    \footnotesize
    \caption{\textbf{Chronology of development, held-out execution and revision.} The final row distinguishes results obtained with the development-stage graph from revisions made after the reserved-paper outcomes had been observed.}
    \label{tab:si_mof_trace_sequence}
    \begin{tabularx}{\textwidth}{@{}>{\raggedright\arraybackslash}p{0.17\textwidth}>{\raggedright\arraybackslash}X>{\raggedright\arraybackslash}X@{}}
        \toprule
        Phase & \grafico{} & Claude Code \\
        \midrule
        Development & Detected the reversed UNSW-100/UNSW-101 CIF contents and encoded composition-, name- and space-group-based identity checks. The completed graph blocked both mismatched pairs; three tests passed before reserved-paper access. & Authored the graph and completed 23 tests without an article--CIF identity check. Its final development run calculated both mismatched UNSW pairs; extraction varied across model calls. \\
        \addlinespace
        First reserved-paper pass & Applied unchanged, the development-stage gate calculated CAU-23 and one KMF-1 deposition, blocked the reversed Zr pair, and blocked Zr-bptc for insufficient formula evidence; the run returned \texttt{partial}. & Processed the KMF-1 and Zr-framework papers without detecting the Zr reversal. CAU-23 entered a \texttt{paper\_failed} carrier after its extraction call reached the 180-s timeout. \\
        \addlinespace
        Revision after observation & Explained the conflicts and, after user authorization, added atom-site composition fallback and provenance-backed local-CIF overrides for both reversed pairs. Six tests passed, and both folders were rerun. & Increased the extraction timeout to 300~s, added one retry and reran the reserved folder. It did not add a structure-identity check or revise the local-CIF associations. \\
        \bottomrule
    \end{tabularx}
\end{table}

\paragraph{\textbf{Trace-level model usage and cost}}

Both trajectory totals were calculated at public API prices and are reported as API-equivalent model costs. For \grafico{}, token counts from the aggregated root spans matched the sums over their child model requests, and the public-price cost was obtained by summing the per-request \texttt{operation.cost} values recorded by Logfire. For Claude Code, repeated assistant records were deduplicated by message identifier and model fallbacks were retained as separate billable iterations; the resulting token categories were then priced at standard first-party API rates current on 28 July 2026.

\begin{table}[!ht]
    \centering
    \footnotesize
    \caption{\textbf{Trace-level model usage and API cost.} Counts are summed over requests and therefore represent accounting events rather than unique text. For the Claude Code input-side total, base input, one-hour cache writes and cache reads are separate billable categories. Reasoning tokens are a subset of the \grafico{} output count.}
    \label{tab:si_mof_trace_cost}
    \begin{tabularx}{\textwidth}{@{}>{\raggedright\arraybackslash}p{0.19\textwidth}>{\raggedright\arraybackslash}X>{\raggedright\arraybackslash}X@{}}
        \toprule
        Measure & \grafico{} & Claude Code \\
        \midrule
        Scope & Three consecutive traces covering initial authoring and reserved-paper execution, explanation of the blocked identities and the user-authorized repair. & Session start through the workflow completion. \\
        \addlinespace
        Model requests & 42 calls to \texttt{gpt-5.6-sol}. & 154 billable model iterations: 27 on Claude Fable~5 and 127 on Claude Opus~4.8, including fallback iterations. \\
        \addlinespace
        Input-side accounting & 3,234,165 input tokens, including 3,111,841 cache-read tokens (96.2\%) and 122,324 uncached tokens. & 25,623,337 token events: 9,675 base-input tokens, 748,189 one-hour cache-write tokens and 24,865,473 cache-read tokens (97.0\%). \\
        \addlinespace
        Output tokens & 32,141, including 7,985 reasoning tokens. & 139,253. \\
        \addlinespace
        API-equivalent model cost & US\$3.13 (unrounded: US\$3.131771), summed from public-price cost fields in the trace telemetry. & US\$28.10 (unrounded: US\$28.104450), reconstructed from recorded usage and public API rates. \\
        \bottomrule
    \end{tabularx}
\end{table}

At these stated scopes, the Claude Code API-equivalent cost was 9.0-fold the \grafico{} API-equivalent cost. The figures exclude scientific-compute, tool and infrastructure costs, and the amount billed for either trajectory may differ under subscription allowances.

\subsubsection{Generated execution graphs}

Both generated programs used the current \texttt{pydantic-graph} GraphBuilder API. A typed Pydantic AI node extracted synthesized frameworks, synthesis conditions, experimental surface areas and deposition identifiers from each paper. Both generated extraction nodes defaulted to \texttt{openai-responses:gpt-5.6-luna}; this workflow model is distinct from the coding models that authored the programs. Structure preparation, $P1$ conversion, Zeo++ execution and numerical comparison were deterministic. Both workflows wrote tables, plots, discrepancy notes and provenance manifests.

\begin{table}[!ht]
    \centering
    \footnotesize
    \caption{\textbf{Implementation choices in the two generated MOF-mining workflows.} Test counts refer to the final state of each single authoring trace and are not a standardized software benchmark.}
    \label{tab:si_mof_implementation}
    \begin{tabularx}{\textwidth}{@{}>{\raggedright\arraybackslash}p{0.15\textwidth}>{\raggedright\arraybackslash}X>{\raggedright\arraybackslash}X@{}}
        \toprule
        Dimension & \grafico{} & Claude Code \\
        \midrule
        Graph topology & Separate paper-parsing and paper-extraction maps, each followed by a join; a subsequent per-record identity, preparation and calculation map terminated in a result join. & A paper-level map combined parsing and extraction before a paper join; a second map processed each (MOF, CCDC) pair before a result join. \\
        \addlinespace
        Failure representation & Invalid identities, preparation failures and calculation failures produced result rows. Extraction failure was retained in run state and provenance but did not produce a paper-level result carrier. & Explicit carriers represented analysed, paper-failed, no-CCDC and CIF-not-found lanes; preparation and calculation exceptions were also retained. \\
        \addlinespace
        Structure identity & Compared deposition number, formula-derived diagnostic elements, recognizable block name and space group. A later revision added atom-site composition fallback and explicit local-CIF overrides that were revalidated. & Resolved each mined identifier to the matching local filename. It did not compare the paper-derived framework identity with CIF formula, name or space group. \\
        \addlinespace
        Solvent removal & Selected FSR when CoRE-MOF outputs and checks passed; otherwise retained the raw CIF with a warning. ASR was not selected automatically. & Selected ASR whenever available, then FSR, and finally the raw CIF if cleaning failed. \\
        \addlinespace
        Zeo++ settings & High-accuracy mode; $1.86$~\AA{} probe radius; 4,000 surface-area and pore-volume samples per atom. & High-accuracy mode; $1.86$~\AA{} probe radius; 2,000 surface-area samples and 50,000 pore-volume samples per atom. \\
        \addlinespace
        Tests & Three tests before the first reserved-paper run, a fourth dry-run test after that pass, and six after the identity-repair revision. & Twenty-three unit and integration tests before the final reserved-paper rerun. \\
        \addlinespace
        Provenance & Hashed run artefacts and retained source input, commands, output paths, failures and final reports. & Registered paper, extraction, structure and report artefacts and retained per-command arguments, stdout, stderr, environment and final reports. \\
        \bottomrule
    \end{tabularx}
\end{table}

The graph-construction skill provides a common execution boundary without prescribing a fixed topology. Claude Code implemented more granular terminal carriers and broader test coverage. \grafico{} implemented the additional scientific identity gate. These differences are directly observable in the generated code and run artefacts.

\subsubsection{Development-set outputs}

The two extraction nodes did not always define the same set of framework records. Table~\ref{tab:si_mof_development_results} reports the target frameworks from the last complete development-stage results before either agent accessed the reserved paper contents. Claude Code additionally emitted three rows without a local deposited structure (FJU-40-H, MOF-74-Ni and UiO-66-(OH)$_2$), which remained explicit \texttt{no\_calculated\_value} outcomes.

\begin{table}[!ht]
    \centering
    \footnotesize
    \caption{\textbf{Development-stage surface-area outputs.} Values are reported BET and calculated Zeo++ accessible surface area in \(\mathrm{m^2\,g^{-1}}\). These runs preceded access to the reserved paper contents.}
    \label{tab:si_mof_development_results}
    \begin{tabularx}{\textwidth}{@{}>{\raggedright\arraybackslash}p{0.16\textwidth}>{\raggedright\arraybackslash}p{0.15\textwidth}>{\raggedright\arraybackslash}X>{\raggedright\arraybackslash}X@{}}
        \toprule
        Framework & Reported BET & \grafico{} & Claude Code \\
        \midrule
        PRI-1 & 317.3 (CO$_2$) & 0; CCDC 641709, FSR & 0; CCDC 641709, ASR \\
        ALF & 588 & 0; CCDC 2098663, FSR & 0; CCDC 2098663, ASR \\
        ZJU-300 & 796.2 & 0; CCDC 2218592, FSR & 0 for CCDC 2218592--2218594, ASR; the final extraction omitted the BET value \\
        Ca-tcpb & 260.65 & 107.59; CCDC 2277051, FSR & 102.17; CCDC 2277051, ASR \\
        UNSW-100 & 1,490 & Blocked at article-reported CCDC 2448651 by identity checks & 0; local 2448651.cif, raw; its contents mismatch article-defined UNSW-100 \\
        UNSW-101 & Not numerically reported & Blocked at article-reported CCDC 2448652 by identity checks & 933.82; local 2448652.cif, ASR; its contents mismatch article-defined UNSW-101 \\
        \bottomrule
    \end{tabularx}
\end{table}

\subsubsection{Held-out paper results and interventions}

Table~\ref{tab:si_mof_reserved_results} reports the final numerical outputs together with the revisions that preceded them. The \grafico{} values are from the user approved post-diagnostic rerun. The Claude Code values are from its rerun after the generic extraction retry was added. They are therefore revision results on papers that were originally held out, not a frozen-graph estimate of generalization.

\begin{table}[!ht]
    \centering
    \footnotesize
    \caption{\textbf{Results for the three held-out papers.} Surface areas are in \(\mathrm{m^2\,g^{-1}}\). \emph{Raw} denotes a deposited structure for which CoRE-MOF curation failed. Values calculated from local CIF contents that conflict with the article-reported assignments are retained to show the downstream consequence but are not treated as aligned cross-agent comparisons.}
    \label{tab:si_mof_reserved_results}
    \begin{tabularx}{\textwidth}{@{}>{\raggedright\arraybackslash}p{0.15\textwidth}>{\raggedleft\arraybackslash}p{0.10\textwidth}>{\raggedright\arraybackslash}p{0.20\textwidth}>{\raggedright\arraybackslash}p{0.20\textwidth}>{\raggedright\arraybackslash}X@{}}
        \toprule
        Framework & Reported BET & \grafico{} post-diagnostic & Claude Code final & Qualification \\
        \midrule
        CAU-23 & 1,250 & 1,367.69; CCDC 1878820, FSR & 1,396.96; CCDC 1878820, ASR & Aligned identity; \(+9.4\%\) and \(+11.8\%\) relative to the reported BET. Claude's value followed the timeout/retry revision. \\
        \addlinespace
        KMF-1 & 1,130 & 1,033.80; CCDC 1984701, FSR & 1,040.98 and 1,014.92; CCDC 1984701 and 1984702, ASR & Aligned identities; \(-8.5\%\), \(-7.9\%\) and \(-10.2\%\), respectively. \\
        \addlinespace
        Zr-bptc & 1,030 & 0; CCDC 1557153, raw & 0; CCDC 1557153, raw & Both curation paths failed and retained the solvated deposited structure; zero accessibility is not a comparison with the activated experimental material. \\
        \addlinespace
        Zr-abtc & 1,318 & 1,191.97; article CCDC 1557154, local 1557155.cif, FSR & 2,907.24; local 1557154.cif, ASR & The supplied 1557154 CIF matches article-defined Zr-tptc-(Me)$_2$ rather than Zr-abtc; the Claude Code value therefore does not evaluate the reported Zr-abtc material. \\
        \addlinespace
        Zr-tptc-(Me)$_2$ & Not reported & 2,566.82; article CCDC 1557155, local 1557154.cif, FSR & 1,496.95; local 1557155.cif, ASR & The supplied 1557155 CIF matches article-defined Zr-abtc rather than Zr-tptc-(Me)$_2$; no experimental BET was mined for the latter. \\
        \bottomrule
    \end{tabularx}
\end{table}

For the two aligned frameworks, CAU-23 and KMF-1 deposition 1984701, the independently generated calculations differed from one another by 2.1\% and 0.7\%, respectively. This numerical proximity applies only after matching the same deposited structure. Zeo++ accessible surface area is a static geometric quantity, whereas BET surface area is fitted from an adsorption isotherm. Activation state, residual or coordinated solvent, framework flexibility, pore gating, defects and the adsorptive can therefore produce differences even when the paper--structure association is correct.

\subsubsection{CIF-content mismatches as a scientific-validity gate}

The development article describing UNSW-100 and UNSW-101 assigns CSD 2448651 to UNSW-100 and 2448652 to UNSW-101. The internal contents of the supplied files instead indicate the opposite material identities:

\begin{enumerate}
\tightlist
\item \texttt{2448651.cif} declares the block \texttt{UNSW101}, contains bromine in \ce{C20H22BrN8O6Zn2} and has space group \(C2\), consistent with the article's description of brominated UNSW-101;
\item \texttt{2448652.cif} declares \texttt{UNSW100}, has the bromine-free formula \ce{C20H24N8O6Zn2} and has space group \(P2_1/n\), consistent with UNSW-100.
\end{enumerate}

This mismatch was identified while \grafico{} developed the graph on the five example papers. In response, the agent encoded a general identity-validation gate that compared the article-derived assignment with CIF deposition number, formula-derived diagnostic elements, recognizable block name and space group. The completed development graph blocked both mismatched paper--CIF lanes and calculated neither value. Claude Code instead resolved each article-reported identifier to the same-numbered local file and lacked an equivalent content check, so it assigned the supplied brominated structure to UNSW-100 and the unbrominated structure to UNSW-101.

Both completed development graphs were then run unchanged on the held-out papers. The pre-existing \grafico{} gate detected an analogous mismatch between the article-defined Zr materials and the downloaded CIF contents. The article associated Zr-abtc with 1557154 and Zr-tptc-(Me)$_2$ with 1557155. In the supplied files, however, 1557154 has no nitrogen and space group \(Imma\), matching Zr-tptc-(Me)$_2$, whereas 1557155 contains nitrogen and has space group \(C2/m\), matching the abtc linker and Zr-abtc. The gate therefore blocked both calculations. Claude Code had not implemented this check during development and continued to resolve each article-reported identifier to the same-numbered local file, so it did not detect the reversal and assigned the resulting values to chemically mismatched structures.

After the first held-out pass, the user requested an explanation of the blocked identities. \grafico{} explained why proceeding would invalidate the material-specific comparisons, and the user then authorized the proposed repair. The agent added an override file that selected local 2448652 for article-defined UNSW-100, local 2448651 for UNSW-101, local 1557155 for Zr-abtc and local 1557154 for Zr-tptc-(Me)$_2$. Each override preserved the article-reported material identity and accession identifier in provenance, recorded the replacement local CIF, required at least three independent evidence attributes and revalidated the replacement structure. For Zr-bptc, the supplied 1557153 file lacked a usable formula tag but contained atom-site element information; the same revision added this fallback without changing the identifier. Thus, the article-defined material identity remained the reference, and the overrides changed only which supplied local file represented that material in the calculation.

For this evaluation, the material identities and accession assignments reported in the articles were treated as the reference. The \grafico{} identity gate cross-checked article-derived framework names, accession identifiers and chemical descriptions against CIF-internal block names, formulae, atom-site composition and space groups. Each CIF was readable and could enter structure preparation and Zeo++ calculation. If its content did not match the article-defined material, however, it was an invalid structural input for that material: a successful calculation would still yield a scientifically erroneous material-specific answer. The gate therefore required article--CIF consistency before interpreting or reporting any downstream value.

\subsubsection{Curation, extraction variability and comparison scope}

The CoRE-MOF skill states that ASR may remove coordinated framework species and must not be selected without checking metal coordination and framework connectivity. \grafico{} encoded an FSR preference, while Claude Code encoded an ASR preference on the premise that experimental BET values refer to activated samples. Both workflows continued with the raw deposited structure when CoRE-MOF failed, although the skill instructed the workflow to stop on missing or unreadable curation outputs. The Zr-bptc result illustrates why the retained preparation state matters: both workflows calculated zero accessible area only after CoRE-MOF failed and the raw, guest-containing CIF was converted to $P1$.

Both extraction nodes were \ac{llm}-mediated and varied between runs. One \grafico{} repair rerun mentioned PRI-1 in its warnings but returned no typed record, so its combined development plot reused the preceding validated row. One Claude Code development rerun omitted a previously extracted ZJU-300 BET value. These single traces therefore do not estimate model-level extraction success. Quantifying extraction variability and its impact on downstream calculations would require repeated authoring runs with frozen extraction outputs and an audited paper--structure reference set.